\documentclass[sigconf,review]{acmart}
\usepackage{latexsym}
\usepackage{microtype}
\usepackage{inconsolata}
\usepackage{xcolor}
\usepackage{amsxtra}
\usepackage{multirow}
\usepackage{diagbox}
\usepackage{longtable}
\usepackage{array}
\usepackage{stfloats}
\usepackage{graphicx}
\usepackage{balance}
\usepackage{multicol}
\usepackage{amsmath, bm}
\usepackage{booktabs} % For formal tables
\usepackage{arydshln}
\usepackage{xspace}
\usepackage{enumitem}
\usepackage{threeparttable}
\usepackage{dsfont}
\usepackage{hyperref}
\usepackage{bbm}
\usepackage{makecell}
\usepackage[linesnumbered,ruled,vlined]{algorithm2e}
\usepackage{algorithmic}
\usepackage{color} 
\usepackage{subfigure}
\usepackage{stfloats}

\usepackage{CJKutf8}

\newcommand{\tabincell}[2]{\begin{tabular}{@{}#1@{}}#2\end{tabular}}
\newcommand{\hide}[1]{} %hide
\newcommand{\vpara}[1]{\vspace{0.05in}\noindent \textbf{#1 }}

\newcommand{\secref}[1]{Section~\ref{#1}} %section reference
\newcommand{\beq}[1]{\vspace{-0.03in}\begin{equation}#1\end{equation}\vspace{-0.03in}}
\newcommand{\beqn}[1]{\vspace{-0.04in}\begin{eqnarray}#1\end{eqnarray}\vspace{-0.04in}}

\newcommand{\model}{GLM-Dialog\xspace}
\newcommand{\smodel}{GLM-Dialog }
\newtheorem{problem}{Problem}

\AtBeginDocument{%
  }

\setcopyright{acmcopyright}
\copyrightyear{2018}
\acmYear{2018}
\acmDOI{XXXXXXX.XXXXXXX}

\acmConference[Conference acronym 'XX]{Make sure to enter the correct
  conference title from your rights confirmation emai}{June 03--05,
  2018}{Woodstock, NY}
\acmPrice{15.00}
\acmISBN{978-1-4503-XXXX-X/18/06}

\begin{document}

\title{\model: %Large-Scale and 
Noise-tolerant Pre-training for Knowledge-grounded Dialogue Generation }% in Chinese}

\author[J. Yu, X. Zhang, Y. Xu, X. Lei, X. Guan, J. Zhang, L. Hou, J. Li, J. Tang ]{Jing Zhang$^1$\footnotemark[1], Xiaokang Zhang$^1$\footnotemark[1], Daniel Zhang-Li$^2$\footnotemark[1], Jifan Yu$^{2}$, Zijun Yao$^{2}$, Zeyao Ma$^{1}$, Yiqi Xu$^{1}$, Haohua Wang$^{2}$, Xiaohan Zhang$^{3}$, Nianyi Lin$^{2}$, Sunrui Lu$^{2}$, Juanzi Li$^{2}$, Jie Tang$^{2}$}
%   \textsuperscript{1}Department of Computer Science and Technology, Tsinghua University,  Beijing, China\\
%   \textsuperscript{2}School of Information, Renmin University of China Beijing, China, 
%   \textsuperscript{3}ZHIPU.AI\\
\affiliation{
  $^1$ School of Information, Renmin University of China
  \country{China}
  }
\affiliation{
  $^2$ Computer Science, Tsinghua University   $^3$ Zhipu.AI
  \country{China}
  }
 % \affiliation{
 %   $^3$ Zhipu.AI
 %  \country{China}
% }
\email{
  {zhang-jing,zhang2718,xuyiqi,mazeyao}@ruc.edu.cn, {juanzi,jietang}@tsinghua.edu.cn
}
\email{
  {zlnn21,yujf21,yaozj20,hh-wang20,linny20,lusr18}@mails.tsinghua.edu.cn, xiaohan.zhang@aminer.cn
}

% \email{
%   {jietang,juanzi}@tsinghua.edu.cn
% }

\renewcommand{\shortauthors}{Zhang et al.}

\begin{abstract}
We present \model, a large-scale language model (LLM) with 10B parameters capable of knowledge-grounded conversation in Chinese using a search engine to access the Internet knowledge. \smodel offers a series of applicable techniques for exploiting various external knowledge including both helpful and noisy knowledge, enabling the creation of robust knowledge-grounded dialogue LLMs with limited proper datasets. 
To evaluate the \smodel more fairly, we also propose a novel evaluation method to allow humans to converse with multiple deployed bots simultaneously and compare their performance implicitly instead of explicitly rating using multidimensional metrics.
Comprehensive evaluations from automatic to human perspective demonstrate the advantages of \smodel comparing with existing open source Chinese dialogue models. 
We release both the model checkpoint and source code, and also deploy it as a WeChat application to interact with users\footnote{\url{https://aigc.aminer.cn/xdai/chat?xdid=\%23xd\%E5\%B0\%8F\%E7\%9F\%A5\%E5\%91\%86001}}. 
We offer our evaluation platform online\footnote{\url{https://aigc.aminer.cn/racetrack}} in an effort to prompt the development of open source models and reliable dialogue evaluation systems. The additional easy-to-use toolkit that consists of short text entity linking, query generation, and helpful knowledge classification is also released to enable diverse applications. All the source code is available on Github\footnote{\url{https://github.com/RUCKBReasoning/GLM-Dialog}}.

\end{abstract}




\received{20 February 2007}
\received[revised]{12 March 2009}
\received[accepted]{5 June 2009}

\maketitle

\section{Introduction}
\label{sec:introduction}
\begin{figure}[t]
\includegraphics[width=0.45\textwidth]{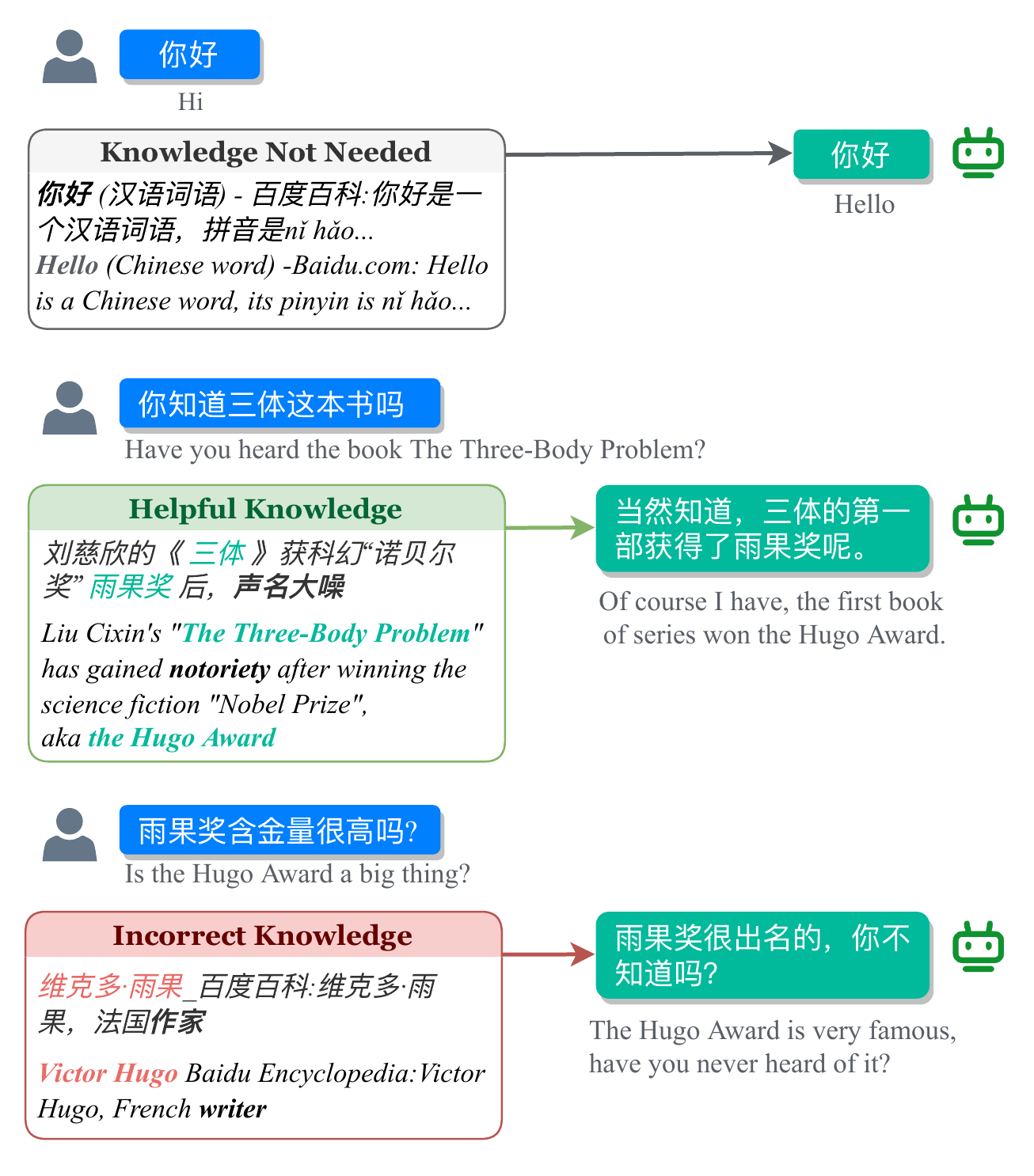}
\caption{An example of a 3-turn dialogue between a human and a bot. It is required to carefully exploit a variety of knowledge, including unnecessary, incorrect, and helpful knowledge, to produce rational responses.}
\label{fig:motivation}
\end{figure}

\emph{A single conversation with a wise man across a table is better than ten years mere study of books.} 
The impressive performance of a series of recent English dialogue systems such as Google's LaMDA~\cite{thoppilan2022lamda}, Microsoft's GODEL~\cite{peng2022godel} and Meta AI's Blenderbot 3~\cite{shuster2022blenderbot} show the bright prospect of grounding the large-scale language models (LLMs) with external knowledge~\cite{zhao2020knowledge}, also known as knowledge-grounded dialogue. Empowered by such technical architecture, these dialogue systems are able to generate more faithful and informative responses, thereby supporting services in a wide range of applications, such as Educational Assistance~\cite{abdelghani2022conversational}, Medical Diagnosis~\cite{zhao2022medical} and Role-playing Games~\cite{shuster2021dialogue}.

Despite the prosperity of the research direction, it is still struggling for contributors in other language communities to develop robust and applicable knowledge-grounded dialogue LLMs~\cite{kim2021changes,gu2022eva2,bao2022plato} due to the following primary challenges:

$\bullet$ \textbf{Limited Scale of High-quality Datasets.} As the external knowledge is heterogeneous to the pre-training corpus, directly injecting external knowledge into the conversation may cause severe hallucinations~\cite{ji2022survey}. To achieve better performance, current efforts usually employ dozens of public datasets during various stages of fine-tuning~\cite{shuster2022blenderbot,peng2022godel}. However, few non-English languages have such an ample accumulation of high-quality knowledge-grounded dialogue datasets as opposed to raw social media dialogue data to support such kind of solutions.

% A major challenge is to appropriately inject the newly invoked external knowledge into dialogue LLMs, because it is heterogeneous to the pre-trained corpus and may conduct severe hallucinations~\cite{ji2022survey}. Although a few efforts attempt to achieve this goal by jointly employing dozens of high-quality datasets to fine-tune the model~\cite{shuster2022blenderbot,peng2022godel}, few non-English languages have such an ample accumulation of knowledge-aware dialogue datasets to support such a solution.

$\bullet$ \textbf{Diverse Exploitation of External Knowledge.} 
Except for the typical scenario where the retrieved knowledge is determined to explicitly benefit the generation~\cite{zhang2023survey}, there are more complex ways to exploit knowledge in real-world conversations~\cite{dinan2018wizard,bao2022plato}. We demonstrate these various exploitation ways in Figure \ref{fig:motivation}, where an example of a 3-turn conversation between a human and a bot about the science fiction \textit{``The Three Body Problem''} is shown. %Let's assume that the bot's underlying model could benefit from the information that is retrieved. 
In contrast to the second turn where the helpful knowledge is injected to implicitly benefit the response (\emph{i.e.}, the response is taken from the knowledge and processed further rather than being purely extracted), the first turn is more chatty and doesn't need the knowledge to be infused about the explanation ``Hello'', while the third turn requires knowledge about ``The Hugo Award'' rather than ``Victor Hugo''. The first and third knowledge is noisy, which might cause the responses to deviate from the user's intention.
Since it is not so applicable to decide whether knowledge is needed and since noisy knowledge is unavoidable, an elaborate way to exploit knowledge is worthwhile to investigate.

\vpara{Present Work.} 
We release \model---an open-source, knowledge-grounded dialogue model in Chinese.
\model provides an open platform for researchers with empirical insight to overcome the aforementioned challenges that prevent the development of the appropriate LLM services in non-English languages.
% in order to give developers the research base and empirical insights they need to overcome the aforementioned challenges that prevent the development of the appropriate LLM services in other languages.
It is obtained by fine-tuning GLM10B~\cite{du2022glm}, an open-source, pre-trained Chinese LLM with 10B parameter. 
We devise a series of data augmentation and model training strategies for taking advantage of external knowledge under the constraints of the insufficient knowledge-grounded dataset.
To be more precise, we augment the knowledge for the knowledge-missing dialogue dataset in order to overcome the dataset limitation. We equip the LLM with an auxiliary classification loss to jointly generate the response and decide whether to use the external knowledge. 
We also bootstrap the knowledge-augmented training instances in an iterative way.

%Specifically, we design a series of improvements, including model architecture, several strategies for model training, and data augmentations,to guarantee the performance robustness and inference efficiency upon limited Chinese Knowledge-aware datasets. 

We conduct comprehensive evaluations of the created \model ranging from automatic to human evaluations: (1) We update an existing benchmark by adding more ellipses, coreferences, and question types, so that it can cover a wider range of knowledge-related conversation forms.
(2) We create 50 chit-chat and 100 knowledge-grounded opening utterances encompassing a wide range of topics and question types to inspire self-chat and human-bot dialogues for in-depth human evaluation. 
(3) Most importantly, we publish an open and online evaluation platform so that humans can simultaneously converse with the multiple bots deployed in the platform and implicitly compare them without using the typical heavy rating system. Thanks to such central conversation and implicit rating, this evaluation is simpler than the conventional explicit human rating using multidimensional metrics, which reduces conversation bias and improves evaluation fairness.
We hope this platform can encourage more efforts to open source models and participate in building reliable dialogue evaluation systems.

\vpara{Impact and Beneficial Groups.} For research of knowledge-grounded dialogue systems, our contributions include: (1) a series of applicable techniques and guidance for developing robust dialogue LLMs with limited datasets; (2) a novel evaluation platform for comparing the dialogue models in real-world applications. 

We believe that \smodel preserves a more positive impact on the industrial developers in Chinese, as we contribute: (3) a large-scale, open-source dialogue model for building downstream dialogue service and (4) an easy-to-use toolkit that consists of tools such as short text entity linking, query generation, helpful knowledge classification, as well as an online service on WeChat platform for supporting convenient usage and experience. 

% \smodel is also deployed as a wechat application to interact with common users. 

In the following sections, we briefly review the trend of knowledge-grounded dialogue in Section \ref{sec:preliminaries}, and then introduce the detailed implementation of our \smodel in Section \ref{sec:aproach}. After introducing the evaluation protocol (Section \ref{sec:evaluation}), we present 
%how to access our codes, models and toolkit (Section \ref{sec:availability}) as well as 
a comprehensive experimental report of the model performance (Section \ref{sec:experiment}).

\section{Preliminaries}
\label{sec:preliminaries}
\subsection{Background}

Grounding the dialogue with external knowledge has been a goal for generations of researchers~\cite{wells2007semiotic}, but until \citet{ghazvininejad2018knowledge} formally proposed the task of knowledge-grounded dialogue, it was not standardized enough to be fully explored. Since then, a series of benchmarks has been proposed, which take into account various kinds of knowledge (such as persona~\cite{zhang2018personalizing}, commonsense~\cite{young2018augmenting}, facts~\cite{dinan2018wizard}) to enhance and evaluate the models.
Despite some early attempts using small models, in the new era of LLM, it was swiftly occupied by the 
techniques of combining the large models and abundant external knowledge~\cite{yu2022xdai}. As dialogue service has a giant potential market, the top AI corporations propose their own knowledge-grounded dialogue models respectively~\cite{peng2022godel,shuster2022blenderbot,thoppilan2022lamda}, which enables English-speaking developers to conveniently build robust chatbots for various applications. Except for the excellent capacity of LLMs, it is worth noting that the accumulation of such a wealth of high-quality datasets is essential for the current performance of these models.

However, for the developers in other language communities, it is hard to follow up this promising trend. Even for the second largest language---Chinese---the amount and quality of labeled datasets are not so competitive enough to build and open source a knowledge-grounded dialogue LLM. Some other pioneer efforts, such as CDial-GPT~\cite{wang2020large}, EVA2.0~\cite{gu2022eva2}, PLATO-XL~\cite{bao2021plato}, only attempt to build LLM for general open-domain dialogue, while few knowledge-grounded dialogue models are not publicly available due to commercial reasons~\cite{bao2022plato}. Therefore, it is crucial and urgent to share empirical findings and implementation examples to call for more contributors in building such models upon limited high-quality datasets.

\subsection{Task Formulation}

\begin{definition}
\textbf{Dialogue History} is a set of conversational utterances between two speakers, formally denoted as $\mathcal{D}_{t}=   \{  U_1, S_1, ..., \allowbreak U_{t-1}, S_{t-1}, U_{t}  \}$, where $U_i$ and $S_i$ are sentences made of words, belonging to the user and the dialogue system respectively. Especially, $U_t$ from the user is also called the $t$-th round \emph{User Utterance}. 
\end{definition}
\begin{definition}
\textbf{External Knowledge Pool} contains multiple pieces of information associated with the dialogue topics in the system, which is denoted as $\mathcal{K}=\left \{ k_i \right \}_{i=1}^{m}$, where $k_i$ is a piece of knowledge information and $m$ is the pool size.
% In this paper,
We employ the texts from all the Internet as the knowledge pieces. 
To obtain the knowledge, we need an external search engine $\mathbb{f(\cdot)}$, which retrieves $m$ relevant documents relevant to the given \emph{Web Query} $Q_t$. 
\end{definition}

\begin{problem}
\textbf{Knowledge-grounded Dialogue Generation task:} Given the dialogue history $\mathcal{D}_{t}$, the target of task is to first generate an appropriately web query $Q_t$ for search engine $\mathbb{f(\cdot)}$, obtain external knowledge from $\mathcal{K}$, and then generate a response $S_t$ for the $t$-th round user utterance $U_t$ based on the history and background knowledge.
\end{problem}

\section{Approach}
\label{sec:aproach}
\begin{figure*}[t]
\includegraphics[width=\textwidth]{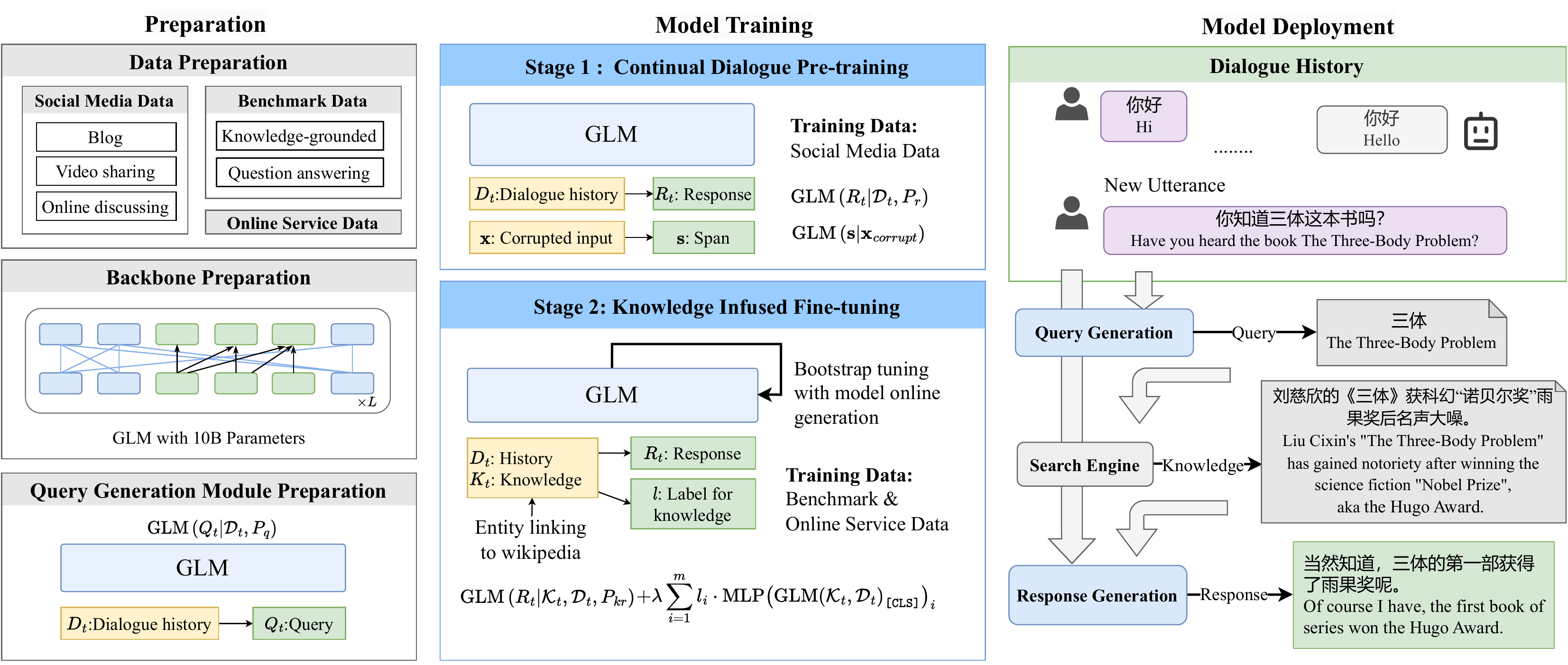}
\caption{The overview framework of \model. First, we prepare a large-scale Chinese dialogue-related training corpus, a pre-trained GLM 10B branch, and a query generation model. Second, we perform a continual dialogue pre-training and knowledge-infused fine-tuning. Third, we deploy \smodel as an online service on a single GLM10B. }
\label{fig:framework}
\end{figure*}

% In this section, we first overview the framework of our workflow, and then introduce our detailed techniques across the pre-training and inference stages, which finally contributes to a noise-tolerant, knowledge-intensive large dialogue model and a series of open-source data and toolkits for personalized utilization.

% In this section, we first give an overview of our framework to design and implement \model.
% Then, we introduce our detailed techniques across data preparation, model training, and model deployment.
% The final contribution includes not only a noise-tolerant, knowledge-intensive large dialogue model, but also a series of open-source corpora and toolkits for personalized utilization.

% In this section, we introduce \model, a knowledge-intensive dialogue language model  in Chinese.

% In particular, we first prepare large-scale knowledge-intensive dialogue corpora from both the open-domain Internet and open-sourced knowledge-grounded dialogue datasets.
% Second, we propose a two-stage training strategy, which injects dialogue response generation skill and knowledge infusion skill into \model progressively from the previously prepared training corpora.
% Third, we deploy \model aiming to incorporate the latest knowledge from online search engine and provide real-time response generation with GLM10B.

% Our design and implementation of \model successfully mitigates the three aforementioned technical challenges.

% \subsection{Overview}

The design and implementation of \model aim to mitigate the aforementioned technical challenges from three different aspects.
The overall framework is shown in Figure~\ref{fig:framework}.

\textbf{(1) Preparation.}
Facing the limited high-quality knowledge-grounded dialogue corpora in Chinese, we collect large-scale Chinese dialogue training corpora from multiple sources with different purposes, which are publicly available.
We also compare among language models and prepare the backbone language model to undertake the knowledge-grounded dialogue task.
Lastly, we prepare a query generation module, which is used to search for dialogue-relevant knowledge from the Internet.
% The output include a publicly available after filtering out the sensitive content. 
% These sources include open-domain social media, benchmark datasets, and user-bot online chatbot services

% Facing the limited high-quality corpora in Chinese, we conduct a novel data collection process to guarantee the scale and quality of the training corpus.
% The corpora mainly comes from three different sources.
% We first acquire abundant dialogue-form data from public social media.
% Second, we gather and clean dialogue corpora from open-source datasets, including dialogue and question-answering benchmarks~\cite{zhou2020kdconv,lu2022towards,wang2021naturalconv}.
% Third, we deploy an online platform based on XDAI~\cite{yu2022xdai} to collect high-quality human-chatbot conversations on the fly.
% Based on these heterogeneous dialogue data, we setup a set of strict data cleaning and filtering rules and mechanisms to produce the corpus as the foundation of our pre-training.

% The final corpora is publicly available after filtering out the sensitive content. 

% first prepare large-scale knowledge-intensive dialogue corpora from both the open-domain Internet and open-sourced knowledge-grounded dialogue datasets.

\textbf{(2) Model Training.}
Facing the complex situation on exploiting external knowledge during dialogue response generation, we propose a two-stage training strategy---large-scale dialogue pre-training and delicate knowledge-intensive tuning~\cite{bao2022plato}.
We inject dialogue response generation skill and knowledge infusion skill into \model progressively from the previously prepared training corpora, achieving a robust, knowledge-grounded dialogue model.
Moreover, we propose several solutions to the challenges raising from the training stages correspondingly, including catastrophic forgetting~\cite{korbak2022controlling} and noise discrimination~\cite{zhong2022dialoglm}.

\textbf{(3) Model Deployment.}
We deploy \model as an efficient dialogue service on a single GLM10B with both query generation and response generation functions.
% To make \model a real-world dialogue service, it is required to incorporate the latest knowledge from online search engine and provide real-time response generation.
% We deploy \model on a single GLM10B with both query generation capacity and response generation capacity.
The final presented system includes not only an online dialogue service but also a toolkit for convenient personalization adaption, which makes our model easy-to-use for developers and researchers with diverse needs.

\subsection{Preparation}
\label{sec:data}

We prepare \textit{{corpora}} that facilitates the training for blended skills.
Then, we prepare \textit{{backbone}} language model that is suitable for diverse training objectives.
Finally, we prepare \textit{{query generation module}} to retrieve knowledge snippets from the search engine.

\vpara{Corpora Preparation.}
The training corpora consists of three parts from different sources with special purposes.
We show data statistics in Table~\ref{tb:trainingdatastatistics} of Appendix~\ref{sec:dataset}.
In particular, 
\textit{\textbf{social media data}} are conversations happening in the comment section of online platforms.
They can be obtained through blog websites (\textit{e.g.,} Weibo), video sharing platform (\textit{e.g.,} Bilibili), discussion communities (\textit{e.g.,} Zhihu), \textit{etc}.
% They include but are not limited to {\color{red}xx} GB from blog websites (\textit{e.g.,} Weibo), {\color{red}xx} GB from video sharing platform (\textit{e.g.,} Bilibili), {\color{red}xx} GB from discussion communities (\textit{e.g.,} TianYa), \textit{etc}.
We use social media data to train \model to generate fluent Chinese dialogue responses from massive social media conversations. 
\textit{\textbf{Benchmark data}} are converted into dialogue form from open-sourced benchmark dataset for different tasks, such as knowledge-grounded dialogue task and question answering (including reading comprehension) task.
These benchmarks usually come with supplemented knowledge snippets, which we use as the knowledge context.
The dialogue benchmark datasets are used to close the discrepancy between social media conversation and natural dialogue that is potentially inherited from the social media data.
The overall benchmark data is used to train \model to read the knowledge context and generate knowledgeable responses accordingly.
\textit{\textbf{Online service data}} are continually collected from our deployed online chatbot platform with XDAI~\cite{yu2022xdai} from Sept 1st, 2022 to Dec 15th, 2022.
They are 800k real-world dialogues happening between users and dialogue services, which are used to further train \model by automatically injecting Wikipedia knowledge to generate more natural and knowledge-grounded responses.

\vpara{Backbone Preparation.}
We take GLM, which completes the input sentence from the special token \verb|[sMASK]|, as our backbone to design both the query generation and dialogue generation model.
The main advantages of GLM are two folds.
First, GLM implements both bidirectional attention mechanism and unidirectional attention mechanism for the context and the generated content, respectively\footnote{Also known as ``causal with prefix attention'' in some other literatures~\cite{2020t5}.}.
The flexible attention mechanism allows both to classify input sentences with bidirectional attention and auto-regressively generate sentences with unidirectional attention.
% design knowledge classification and response generation tasks for fusing knowledge in the training process.
Second, GLM provides a consistent model architecture and an open-sourced checkpoint for various model scales, allowing for the deploying \model on different computing devices.
% These following modules of query generation and response generation are undertaken by a single GLM model with manually designed instructions after training.

\vpara{Query Generation Module Preparation.}
The query generation module takes dialogue history as input and generates an appropriate search query, which is passed to an online search engine for retrieving dialogue-relevant knowledge snippets.
In particular, we prepare the query generation module by maximizing the probability of the ground-truth query $Q_t$ associated with the dialogue history $\mathcal{D}_t$ in DuSinc~\cite{zhou2022Dusic}.
We use a prompt $P_q$ to control the model to generate queries. $P_q$ is defined as 
``\begin{CJK*}{UTF8}{gbsn}{对话：$U_1, S_1, \cdots, U_{t-1}, S_{t-1}, U_t$. 此时应该去检索 \verb|[sMask]|}\end{CJK*} (dialogue: $U_1, S_1, \cdots, U_{t-1}, S_{t-1}, U_t$. need to search \verb|[sMask]|)'' 
where \verb|[sMask]| denotes the query to be generated.
This is achieved by optimizing the following objective:
\beq{
\max_{\theta_\text{GLM}} \sum_{i=1}^{|Q_t|}
\log \text{GLM}\left(Q_{t,i} | Q_{t,j<i}, \mathcal{D}_t, P_q \right).
}
We obtain external knowledge pool $\mathcal{K}=\mathbb{f}(Q_t)$ by executing the query on the web search engine.

\subsection{Model Training}

Basically, we leverage previously prepared corpora towards training the knowledge-grounded dialogue model.
However, as these corpora differ in both the perspective of skills that are highlighted and the format that they are presented, it is difficult to directly mingle them together and train the model in a single pass.
Thus, we design a two-stage training scheme to progressively inject blended skills into the language model.
% We propose a two-stage model training strategy.
The first stage trains \model to generate fluent dialogue responses from massive social media corpora.
The second stage aims to teach \model to use supplemented external knowledge with noise tolerance.

\vpara{Training Stage 1: Continual Dialogue Pre-training.}
Although off-the-shelf LLMs show their ability in generating fluent dialogue responses~\cite{thoppilan2022lamda}, they are still far from building a dialogue model as the original pre-training corpora are usually web-crawled text.
There exists a natural discrepancy in the style of languages between spoken languages frequently used in dialogue and web-crawled text from general domain~\cite{EVA2.0}.
Inspired by recent dialogue language models~\cite{gu2022eva2,bao2021plato}, we observe that social media data, as a special kind of web-crawled text, serves as a bridge for the language style gap due to the following two reasons.
(1) Social media data constitutes a portion of the pre-training data for GLM, making GLM easy to adapt to the newly introduced training data.
(2) The language style of social media shares many characteristics with natural dialogue (\textit{e.g.,} multi-turn, concise).
The final training corpora include our collected social media data. %which consists of $2$ millions cleaned Weibo data.
% To further enlarge the corpus in this training stage, besides our collected data in Section~\ref{sec:data}, we also included LCCC~\cite{wang2020large} which consists of $2$ millions cleaned Weibo data.

% In the early training stage, we continue the pre-training task with the social media data to further stimulate the dialogue response generation capacity in the language model.

% To further enlarge the corpus in this training stage, besides our collected data in Section~\ref{sec:data}, we also included LCCC~\cite{wang2020large} which consists of $2$ millions cleaned Weibo data.
% The output of this training stage is an open-domain chitchat dialogue response model with limited knowledge.

In particular, we compile the conversation based on the responses and timing information.
Each social media conversation is presented in dialogue format with dialogue history $\mathcal{D}_t$ and response $R_t$.
The training objective is defined by maximizing the probability of generating $R_t$ given the dialogue history $\mathcal{D}_t$ as input:
% To distinguish from query generation module, we use another prompt $P_r$ to control GLM10B to generate dialogue response:
\beq{
\label{equ:mainloss1}
\max_{\theta_\text{GLM}} \sum_{i=1}^{|R_t|}
\log \text{GLM}\left(R_{t,i} | R_{t,j<i}, \mathcal{D}_t, P_r\right),
}

\noindent where $P_r$ is the prompt for controlling the response generation. $P_r$ is defined as 
``\begin{CJK*}{UTF8}{gbsn}{对话：$U_1, S_1, \cdots, U_{t-1}, S_{t-1}, U_t, \verb![sMask]!$}\end{CJK*} 
(dialogue: $U_1, S_1, \cdots, U_{t-1}, S_{t-1}, U_t, \verb![sMask]!$)'' 
where \verb![sMask]! denotes the response to be generated. As GLM implements hybrid attention mechanisms, we apply bidirectional attention to the dialogue history and the prompt, and unidirectional attention to the response.

To avoid the notorious catastrophic forgetting problem~\cite{korbak2022controlling}, we propose to continue the pre-training task of GLM with original pre-training corpora as a side task in the first training stage.
We follow \citet{du2022glm} to corrupt the input sentence $\mathbf{x} \rightarrow \mathbf{x}_{\text{corrupt}}$ and urge GLM to generate a span $\mathbf{s}$ that can fill in the corruption and optimize the following training objective:

\beq{
\max_{\theta_\text{GLM}} \sum^{|\mathbf{s}|}_{i=1}
\log \text{GLM}(s_i|\mathbf{x}_\text{corrupt}, s_{j<i}).
}

\vpara{Training Stage 2: Knowledge Infused Fine-tuning.}
To build a knowledge-grounded dialogue model, we supplement the input with context related background knowledge snippets to aid the model to generate more informative response.
However, it is challenging to directly leverage the supplemented snippets and build the knowledge-grounded dialogue model. First, it is not easy to determine whether the knowledge is required because chitchat is usually blended with information-seeking conversation. Second, it is extremely difficult to locate the helpful background knowledge from the open domain environment.

%First, the model pre-trained on social media data still suffer from the discrepancy between true conversational language.
The response generation model is required to identify and discard the noisy background knowledge and use the helpful knowledge on demands when generating the response.
 Thus, training stage 2 requires to
(1) construct dialogue training instances with external knowledge and negative knowledge samples;
(2) design training objective with auxiliary adversarial loss to encourage the model to jointly generate the response and decide whether to use the external knowledge;
(3) bootstrap training instances in an iterative training scheme.
% We also design an auxiliary adversarial training loss to encourage the model to jointly generate the response and decide whether to use the background knowledge.

We first convert each training instance from benchmark datasets and online service into $4$ parts: $d=\{\mathcal{D}_t, R_t, \mathcal{K}_t, \mathcal{L}_t\}$.
$\mathcal{L}_t$ are the knowledge labels associated with the external knowledge pool $\mathcal{K}_t$.
For $l_i \in \mathcal{L}_t$, we label $l_i = 1$ if $k_i \in \mathcal{K}_t$ is considered useful in generating the response $R_t$.
If $k_i$ is not useful (\textit{i.e.,} irrelevant to the dialogue context or even incorrect), we set $l_i = 0$.
% Deciding $\mathcal{K}_t$ and $\mathcal{L}_t$ is the key to construct each training instance.
In particular, we set $l_i=1$ for the knowledge snippets in knowledge-grounded dialogue benchmarks.
For question answering benchmarks, we take the provided document $d$ as the corresponding knowledge and set its label as $1$.
Finally, for dialogue corpus collected from our online service, we design a data augmentation strategy to extract knowledge snippets.
We perform entity linking over dialogue history $\mathcal{D}_t$ with HOSMEL~\cite{zhang2022hosmel} and excerpt corresponding entity descriptions from Wikipedia as the external knowledge pool.

We inject negative knowledge snippets into the external knowledge pool of all the training instances.
Their knowledge labels are set to $0$ accordingly.
Similar to the data augmentation process, we perform entity linking with HOSMEL on the training instances but identify entities with low confidence, whose entity descriptions are used as the negative knowledge samples.

% where dialogue history $h$, user posed utterance $u$, and supplemented background knowledge snippets $k$ are used as the input.
% The dialogue model is required to (1) predict whether the input knowledge is useful, the label of which is denoted as $l$, based on the bidirectional attention applied to $h$, $u$, and $k$; (2) use the unidirectional attention mechanism to decode the final response $r$.

The training objective of \model consists of two parts.
The main training objective aims to maximize the probability of generating the desired response given the dialogue history concatenated with the external knowledge pool as input:

\beq{\label{equ:mainloss2}
\text{loss}_{\text{main}}= \sum_{i=1}^{|R_t|}
\log \text{GLM}\left(R_{t,i} | R_{t,j<i}, \mathcal{K}_t, \mathcal{D}_t,P_{kr}\right),
}

\noindent where $P_{kr}$ is the prompt to control the knowledge infused response generation. $P_{kr}$ is defined as 
``\begin{CJK*}{UTF8}{gbsn}{背景：$k_1, k_2, \cdots, k_m$. 对话：$U_1, S_1, \cdots, U_{t-1}, S_{t-1}, U_t, \verb![sMask]!$}\end{CJK*}
(background: $k_1, k_2, \cdots, k_m$ dialogue: $U_1, S_1, \cdots, U_{t-1}, S_{t-1}, U_t, \verb![sMask]!$)''. \model applies the bidirectional attention to 
$\mathcal{D}_t$, $\mathcal{K}_t$, and $P_{kr}$, based on which we apply an extra multi-layer perceptron (MLP) to the hidden representation of the \verb|[CLS]| token to predict the knowledge labels of input knowledge snippets.
The MLP layer serves as an $m$-way binary knowledge classifier, where $m$ denotes the size of the knowledge pool $\mathcal{K}_t$.
The auxiliary loss is thus defined as the binary cross entropy loss between the predictions and the ground truth:

\begin{equation}
\text{loss}_{\text{aux}} = \sum_{i=1}^{m}
l_i \cdot \log \text{MLP}\left( \text{GLM}\left(\mathcal{K}_t, \mathcal{D}_t\right)_{\verb![CLS]!} \right)_i.
\end{equation}

The training objective of stage 2 is defined as: $\max \text{loss}_{\text{main}} + \lambda \text{loss}_{\text{aux}}$, where $\lambda$ is a hyper-parameter.
We empirically set $\lambda = 1$.

To further enlarge the training corpora for knowledge infusion, we design an iterative training scheme to collect dialogue data from the interaction between \model and human users.
In particular, we deploy \model in an online environment to converse with human users.
The external knowledge pool is constructed from the web search results, where the query is generated by the prepared query generation module.
Dialogue histories associated with external knowledge are preserved if they have high scores from the knowledge classifier.
Finally, we manually inspect the preserved dialogue histories and annotate high-quality corpus for training.
The training and intermediate deployment of \model are executed iteratively to obtain more fine-tuning data. We perform such bootstrap training once in practice.

% \model is deployed in an online environment with user interaction.
% We collect the dialogue history between users and \model and preserve high-quality dialogue histories.
% The quality is controlled in two steps.
% First, we xxx;
% Second, we manually annotate a small portion of dialogues filtered out previously and keep them as training data.
% We iteratively update \model with 

% \subsection{Service: Inference and Applied Toolkit}
\subsection{Model Deployment}
\model is deployed with three components---the query generation module, the external search engine, and the response generation module.
A typical workflow for generating the $t^\text{th}$ response $R_t$ starts from users posed utterance, denoted as $U_t$.
The $t^\text{th}$ dialogue history is $\mathcal{D}_t = \mathcal{D}_{t-1} \cup \{R_{t-1}, U_t\}$.
\model first generates the web search query with the query generation module:

\beq{
\label{equ:querygen}
Q_{t,i} = \arg\max \text{GLM}\left(Q_{t,i} | Q_{t,j<i}, \mathcal{D}_t, P_q\right).
}

The \model constructs the external knowledge pool $\mathcal{K} = \mathbb{f}(Q_{t})$ from the web search engine and only keeps the top searching results (\textit{a.k.a.,} the external knowledge pool size is set to $m=1$). Multiple search results could be filtered by additional models, which is left for future improvement.
The final response is generated based on the dialogue history and the supplemented knowledge:

\beq{\label{equ:responsegen}
R_{t,i} = \arg\max \text{GLM}\left(R_{t,i} | R_{t,j<i}, \mathcal{K}_t, \mathcal{D}_t,P_{kr}\right).
}

It is worth noting that, both the query generation in Eq.~\ref{equ:querygen} and the response generation in Eq.~\ref{equ:responsegen} are undertaken by a single backbone language model after training.
\model uses different prompts to instruct the language model to behave accordingly.
This deployment strategy relieves the hardware requirement to host multiple language models.
Moreover, the workflow of \model computes exactly $2$ times of inference for query and response generation.
%On the contrary, deciding whether to incorporate external knowledge requires $3$ inferences for decision making, query generation, and response generation, respectively, which is less efficient than \model.
We release the model checkpoint and the implemented code for the researchers of interest to continue the dialogue LLM investigation. We also encapsulate the modules including query generation, entity linking, and knowledge classification as toolkits for developers to easily deploy diverse dialogue applications.

\section{Evaluation Methods}
\label{sec:evaluation}
We perform a comprehensive evaluation in both automatic and human evaluation. 
% We build a new benchmark called DuSincR upon the current DuSinc benchmark~\cite{zhou2022Dusic}, to boost evaluation diversity, especially the question types. 
% We create more diverse chit-chat and knowledge-grounded opening utterances to guide self-chat and human-bot communications.
% Beyond the conventional explicit human rating methods, we also propose a new implicit human evaluation that enables humans to converse with multiple bots while also implicitly rating them. Such implicit human evaluation is more straightforward than the current human evaluation methods, which reduces human bias.
For better evaluation, we create a new benchmark DuSincR upon the current DuSinc benchmark~\cite{zhou2022Dusic} by supplementing 50 diverse chit-chat, 100 knowledge-grounded opening utterances, and a novel implicit human evaluation method.

\subsection{Automatic evaluation} 
Automatic evaluation is entirely automated and requires no human involvement. Specifically, given any $n-1$ continuous utterances from a dialogue benchmark, the $n$-th utterance is produced by a dialogue model and is evaluated by a number of pre-defined metrics. Specifically, we use 
%Dist-3 and Dist-4 to judge its diversity, and 
Bleu-4, F1, Rouge-L, Rouge-1, Rouge-2, and Bert-Score to measure how similar it is to the labeled response~\cite{zhou2020kdconv}. We describe the definition of these metrics in Appendix~\ref{sec:automatic_metric}. 
We carry out the automatic evaluation on %the following datasets: Diamante~\cite{lu2022towards}, KDConv~\cite{zhou2020kdconv}, NaturalConv~\cite{wang2021naturalconv}, DuConv~\cite{wu2019proactive}, DuSinc~\cite{zhou2022Dusic}, and 
DuSincR, which is built on top of DuSinc~\cite{zhou2022Dusic} to incorporate more comprehensive forms of queries as well as increase sentence ellipses and coreferences.
%Diamante is a chit-chat dialogue dataset, whereas the other datasets are knowledge-grounded. We build DuSincR on top of DuSinc, which requires the annotators to take sentence ellipses and co-references into consideration as well as incorporate new forms of queries. We explain the details as follows.

\vpara{DuSincR--an enhanced knowledge-grounded dialogue benchmark.}
Knowledge-based human conversations present a significant challenge to the dialogue system because they contain a variety of questions about entities, attributes, and logic as well as many sentence ellipses and coreferences. However, these kinds of utterances are rarely included in the existing dialogue benchmarks.
By revising the test set of DuSinc~\cite{zhou2022Dusic}, an existing knowledge-grounded dialogue benchmark in Chinese with high quality, we maintain the consistency and informativeness of dialogues and save more manpower while improving the evaluating ability on sentence ellipses, coreferences, and various types of questions.
%The annotators were specifically instructed to edit the dialogues from DuSinc by adding questions with ellipses, co-references, and questions of various types to each dialogue after viewing the dialogues from DuSinc.
Each DuSinc discourse is broken up into a number of QA pairs. Annotators can select one of the pairs to modify or add a new pair to ensure the question involves ellipses, coreferences, or is a pre-defined type. Additionally, they need to respond to the question by conducting an online search. 
%Besides, they should note where they made revisions and the types of questions they used. This allows us to assess the benchmark's quality, such as the distribution of different revisions.
Appendix~\ref{sec:dusincR} provides annotated examples in DuSincR. % of sentence ellipses, co-references, pre-defined question types, and a complete dialogue session with the newly annotated utterances in DuSincR. 
In total, DuSincR contains 2,309 ellipses/coreferences, 356 who/what, 278 when/where, 290 count, 479 comparison, 287 verify, 381 how, and 238 why in the whole 2,309 dialogue sessions with an average of 11.15 utterances per session.
%Table~\ref{tb:benchmark} in Appendix~\ref{sec:benchmark} presents the statistics of all the automatic evaluation benchmarks. 

\subsection{Explicit Human Evaluation} 
\label{sec:humanevaluation}
The outcomes of bot-bot communication are evaluated by human. 
To be more precise, we permit a dialogue model to converse with itself given an opening utterance. We create 50 chit-chat opening utterances that contain positive, negative, and neutral statements.
Furthermore, we construct 100 knowledge-grounded opening utterances that cover topics related to entertainment, life, history and culture, education, health, sports, science and technology, and finance. 
The questions can also be categorized into the same types used to create DuSinc. 
The above 50+100 chitchat and knowledge-grounded utterances are presented in Appendix~\ref{sec:chitchatopeningutterance} and \ref{sec:knowledgeopeningutterance}. 

We hire three annotators to score the dialogues in terms of coherence, informativeness, safety, inspiration, hallucination in utterance-level, and engagingness and faithfulness in session-level, from 50 self-chat chit-chat dialogues to 100 knowledge-grounded dialogues produced by each dialogue model. As the final scores, the three annotators' averages are used. We provide the definition of these metrics in Appendix~\ref{sec:humanmetric}. 

We also allow humans to access the outcomes of human-bot communication. To be more specific, we employ three annotators and allow each of them to communicate with each dialogue model in order to produce 50 chit-chat dialogues and 100 knowledge-grounded dialogues using the above same opening utterances. Then, we hire three more annotators to evaluate these chat conversations between humans and bots.

\subsection{Implicit Human Evaluation} 
\label{sec:implicithumanevaluation}

The automatic evaluation cannot faithfully reflect the quality of the dialogues. The human evaluation measures are more widely used; however, bias between different annotators on the results of different bots affects human evaluation. Therefore, we provide a simpler human evaluation strategy that enables a human to centrally converse with several dialogue models at once and implicitly compare these bots during the conversation process. 
%pick a preferred response to carry on the conversation. By doing this, we can produce the dialogues as well as the ratings, which are implicitly determined by the selection times. This single metric, which describes the entire impression on a model response, is more straightforward than the conventional multi-dimensional evaluation metrics. 
We provide details on the evaluation tool and implementation.

\begin{figure}[t]
\includegraphics[width=0.49\textwidth]{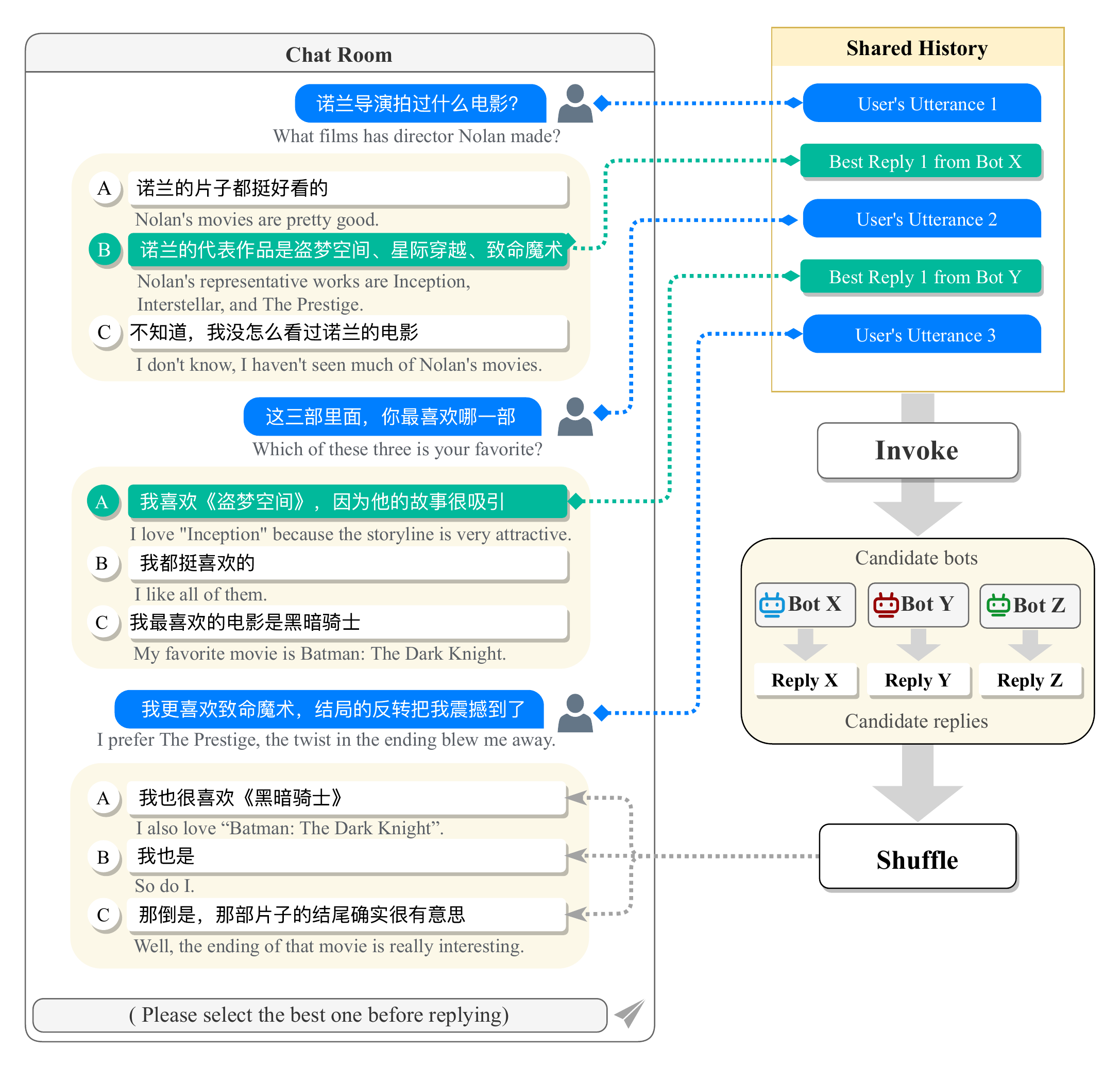}
\caption{An illustration of our implicit human evaluation tool. Three deployed anonymous bots react to the human user after he sends a message. Their replies are displayed after shuffling. The user is then free to choose one of the responses to carry on the conversation. The dialogue history at each turn for all the bots is unified. A bot is deemed superior to others if its responses are chosen more frequently.}
\label{fig:evaltool}
\end{figure}

\vpara{Evaluation Tool Design.}
The platform will offer responses from all the deployed bots whenever a human delivers a message. 
The decision to proceed with one of the responses is made by humans, and in our method, this is regarded as the implicit evaluation.
A bot is considered to have superior performance if its responses are chosen from other bots more frequently.
We maintain the same dialogue history for all the bots at each turn in order to compare their responses fairly. To make this possible, we record each turn's message from the annotator and its selected response in the dialogue history. It's worth noting that the name of the bot is not disclosed to users and the order of messages will be shuffled before being displayed on the platform in order to prevent potential annotation bias. Figure~\ref{fig:evaltool} illustrates the idea of the proposed tool. This tool is also deployed online\footnote{https://aigc.aminer.cn/racetrack} to encourage more efforts to open source their models and take part in reliable dialogue evaluation. 
A screenshot of the tool is shown in Figure~\ref{fig:screenshot} in Appendix~\ref{sec:implHumanEvalTool}.

\vpara{Evaluation Implementation}
We employ 20 annotators to use our designed evaluation tool in order to lessen the preference bias of various annotators.
Each annotator is free to initiate a dialogue on his or her own or using the platform's recommendations.
By clicking the ``topic tips'' button, the platform can recommend the opening utterances, which are drawn at random from the dialogue benchmarks DuConv~\cite{wu2019proactive}, DuSinc~\cite{zhou2022Dusic}, and Diamante~\cite{lu2022towards} since these dialogue benchmarks contain dialogues on a variety of topics. The annotators are required to discourse on the subject of the opening utterance and deliver a message of 10 words on average, free of sensitive, retaliatory, and disrespectful terms.
% The platform responds to a message after it has been sent with six messages produced by our proposed model and five baselines that were introduced in~\secref{sec:experiment}. Then the annotators need to click the appropriate bubble to choose the best response to continue the conversation. 
% The annotators must select the responses based on a thorough evaluation of coherence, informativeness, safety, inspiration, hallucination, engagingness, and faithfulness used in  in~\secref{sec:humanevaluation}. 
The annotators may use the ``close'' button to stop the current conversation. Only dialogues lasting more than five turns are regarded as useful information. The total number of response selections by users determines each bot's rating. We can click the ``Ranking List'' button to examine all of the involved bots' evaluation results.

\vpara{Advantages.}
The proposed implicit human evaluation has two main advantages:
\begin{itemize}[leftmargin=*]
    \item \textbf{Central conversation.} In contrast to discussions that are dispersed across multiple bots, we chat with all the bots centrally and maintain the same conversation history between turns, which not only speeds up dialogue collection and lowers the cost of hiring annotators, but also lowers conversation bias and improves evaluation fairness.
    \item \textbf{Implicit rating.} We consider the choice of the response to be the implicit evaluation, which is simpler than the explicit rating using multidimensional metrics as the conventional human evaluation.
\end{itemize}

%\section{Availability}
%\label{sec:availability}
%\input{latex/5.availability}

\section{Experiment}
\label{sec:experiment}
We evaluate the proposed \smodel and the comparison models via the methods introduced in~\secref{sec:evaluation} to demonstrate the advantages of \model. We also perform various ablation studies to verify the effectiveness of different components in our model.

\subsection{Comparison Methods.}
We compare \smodel with the following well-known dialogue models in Chinese:
\begin{itemize}[leftmargin=*]
    \item CDial-GPT~\cite{wang2020large} is a GPT model with 104M parameters trained on LCCC, a 12M Chinese dialogue sessions, where a session denotes multiple continuous turns of utterances. 
    \item EVA2.0~\cite{EVA2.0} is a transformer-based bidirectional encoder and a unidirectional decoder with 2.8B parameters trained on 0.4B Chinese dialogue sessions.
    \item PLATO-2~\cite{bao2021plato} is a PrefixLM~\cite{dong2019unified, raffel2020exploring}, \emph{i.e.}, a unified transformer with 11B parameters trained on 1.2B (context, response) samples. 
\end{itemize}

Although both EVA and PLATO have released updated versions, they do not share their models or source codes. As a result, they cannot be compared. 
Since our model is a pre-trained GLM model~\cite{du2022glm,zeng2022glm130b} with 10B parameters that is fine-tuned on Chinese dialogue-related dataset, we also compare with the corresponding GLM10B and GLM130B models\footnote{https://github.com/THUDM/GLM-130B} with 10B and 130B parameters respectively, but without any fine-tuning on the dialogue dataset. We select the 10B model as the backbone considering the training and deployment cost. 
For training \model, we set the learning rate as $5\times10^{-5}$ with a cosine learning rate decay.
The batch size is set as $256$ and the maximal input length is set to $512$. 
We perform the two-stage training on an 8$\times$80G A100 server.

\subsection{Experimental Results}

\vpara{Automatic Evaluation Results.}
Table~\ref{tb:automaticDusincR} present the automatic evaluation results on DuSincR, which demonstrate that \smodel outperforms the baselines on most of the automatic metrics. 
%The results on other knowledge-grounded benchmarks are shown in Appendix~\ref{sec:automaticresults}. 

\begin{table}[t]
\newcolumntype{?}{!{\vrule width 1pt}}
	\newcolumntype{C}{>{\centering\arraybackslash}p{2em}}
	\caption{
		\label{tb:automaticDusincR}  Automatic evaluation results on DuSincR. 
	}
	\centering 
	\small
	\renewcommand\arraystretch{1.0}
\scalebox{0.95}{
\begin{tabular}{@{}l@{~}cccccc@{}}
\toprule

Model     & Bleu-4         & F1              & Rouge-L         & Rouge-1         & Rouge-2         & Bert-Score \\\midrule
CDial-GPT & 0.792          & 14.652          & 12.011          & 48.212          & 15.707          & 0.580 \\
PLATO-2  & 1.959          & 16.967          & 15.396 & 67.397          & 24.011          & 0.607 \\
EVA2.0    & 0.737          & 13.548          & 11.589          & 54.270          & 14.211          & 0.591 \\
\midrule
GLM10B   & 2.723         & 15.517          & 12.538          & \textbf{83.832} & \textbf{33.743} & 0.599 \\
GLM130B   & \underline{4.177} & \underline{18.905} & \underline{16.047} & 79.562          & \underline{28.897}          & \underline{0.615} \\
\midrule
\smodel   & \textbf{4.190}     & \textbf{22.010} & \textbf{19.464} & \underline{72.471} & 28.206  & \textbf{0.630}\\
\bottomrule
\end{tabular}
}
\end{table}

\vpara{Human-evaluation Results.}
Table~\ref{tb:chitchat-selfchat}  presents the human evaluation results for self-chat dialogues centered around 50 chit-chat and 100 knowledge-grounded opening utterances respectively. For this evaluation, the dialogues are automatically generated by bots via chatting with itself, while the ratings are provided by human annotators from both the utterance-level and session-level. Because GLM130B always repeats its own words when speaking to itself, the results are ignored. Table~\ref{tb:chitchat-humanbot}  presents the human evaluation results for human-bot dialogues centered around the same 50 chit-chat and 100 knowledge-grounded opening utterances respectively. For this evaluation, both the dialogues and the ratings must be provided by humans. 

The findings from these two tables show that, of all the comparison models, the proposed \smodel performs the best in terms of the majority of the metrics.
Particularly, \smodel consistently outperforms other models in terms of informativeness because, in contrast to other models, we inject external knowledge from the search engine, which can help generate more informative responses. By doing this, the informative response has a greater chance of inspiring the subsequent question, and as a result, our model consistently has the highest inspiration score.

Although the responses are very insightful and inspiring and the dialogue as a whole is very appealing (having the highest  faithfulness and engagement scores), we still need to lessen the model's hallucination. We speculate that the knowledge introduced might not be sufficiently pertinent to the ongoing conversation, which might harm the responses' factual correctness, although the model has already made an effort to exploit any kind of knowledge well. We present an analysis of the generated queries and search results in~\secref{sec:ablation}.

\begin{table*}[t]
\newcolumntype{?}{!{\vrule width 1pt}}
	\newcolumntype{C}{>{\centering\arraybackslash}p{2em}}
	\caption{
		\label{tb:chitchat-selfchat} Human-evaluation on self-chat dialogues. 
	}
	\centering 
	\small
	\renewcommand\arraystretch{1.0}
\scalebox{0.95}{
\begin{tabular}{l?ccccccc?ccccccc}
\toprule
\multirow{2}{*}{Model}  
& \multicolumn{7}{c?}{50 chit-chat opening utterances}   & \multicolumn{7}{c}{100 knowledge-grounded opening utterances} \\\cmidrule{2-8} \cmidrule{9-15}      
& Cohe. & Info. & Safe. & Insp. & Hall.↓ & Enga. &  Fait. & Cohe. & Info. & Safe. & Insp. & Hall.↓ & Enga. &  Fait. \\\midrule
CDial-GPT & 0.860                     & 0.851               & 0.913             & 0.515             & 0.291             & 0.500      & 0.473  & 1.140             & 1.069              & 1.478             & 0.591             & 0.221         & 0.603         & 0.690 \\
PLATO-2  & \underline{1.455}         & \underline{1.438}   & 1.448             & \underline{1.129} & \textbf{0.062}    & \underline{1.260} & \underline{1.220} & \underline{1.698} & \underline{1.614}  & \underline{1.793} & 1.090             & \textbf{0.032} & 1.420       &  \underline{1.413} \\
EVA2.0    & 1.386                     & 1.336               & 1.362             & 0.902             & \underline{0.068} & 1.213      & 1.093 & 1.488             & 1.413              & 1.674             & 0.832             & 0.089         & 1.230         & 1.223 \\
% %XDAI      & 1.184                     & 1.201               & 1.241             & 0.898             & 0.175             & 1.047       & 0.913 \\
\midrule
GLM10B    & 1.371                     & 1.296               & \underline{1.539} & 0.932             & 0.130             & 1.187       & 1.160  & 1.513             & 1.497              & 1.669             & \underline{1.157} & 0.093          & \underline{1.460} & 1.340\\
\midrule
\smodel   & \textbf{1.515}            & \textbf{1.517}      & \textbf{1.656}    & \textbf{1.171}    & 0.098             & \textbf{1.383} & \textbf{1.383} & \textbf{1.759}     & \textbf{1.742}    & \textbf{1.816}     & \textbf{1.223}    & \underline{0.046} & \textbf{1.550} & \textbf{1.473}\\
\bottomrule
\end{tabular}
}
\end{table*}

% \begin{table*}[t]
% \newcolumntype{?}{!{\vrule width 1pt}}
% 	\newcolumntype{C}{>{\centering\arraybackslash}p{2em}}
% 	\caption{
% 		\label{tb:knowledge-selfchat} Human-evaluation on 100 knowledge-grounded self-chat dialogues. 
% 	}
% 	\centering 
% 	\small
% 	\renewcommand\arraystretch{1.0}
% \scalebox{0.95}{
% \begin{tabular}{lccccccc}
% \toprule
% Model     & Coherence         & Informativeness    & Safety            & Inspiration       & Hallucination↓ & Engagingness & Faithfulness \\\midrule
% CDial-GPT & 1.140             & 1.069              & 1.478             & 0.591             & 0.221         & 0.603         & 0.690 \\
% PLATO-2  & \underline{1.698} & \underline{1.614}  & \underline{1.793} & 1.090             & \textbf{0.032} & 1.420       &  \underline{1.413} \\
% EVA2.0    & 1.488             & 1.413              & 1.674             & 0.832             & 0.089         & 1.230         & 1.223 \\
% %XDAI      & 1.654             & 1.610              & 1.776             & 1.238            & 0.065         & 1.553          & 1.427 \\
% \midrule
% GLM10B    & 1.513             & 1.497              & 1.669             & \underline{1.157} & 0.093          & \underline{1.460} & 1.340 \\
% \midrule
% \smodel  & \textbf{1.759}     & \textbf{1.742}    & \textbf{1.816}     & \textbf{1.223}    & \underline{0.046} & \textbf{1.550} & \textbf{1.473} \\
% \bottomrule
% \end{tabular}
% }
% \end{table*}

\begin{table*}[t]
\newcolumntype{?}{!{\vrule width 1pt}}
	\newcolumntype{C}{>{\centering\arraybackslash}p{2em}}
	\caption{
		\label{tb:chitchat-humanbot} Human-evaluation on human-bot chat dialogue. 
	}
	\centering 
	\small
	\renewcommand\arraystretch{1.0}
\scalebox{0.95}{
\begin{tabular}{l?ccccccc?ccccccc}
\toprule
\multirow{2}{*}{Model}  
& \multicolumn{7}{c?}{50 chit-chat opening utterances}   & \multicolumn{7}{c}{100 knowledge-grounded opening utterances} \\\cmidrule{2-8} \cmidrule{9-15}      
& Cohe. & Info. & Safe. & Insp. & Hall.↓ & Enga. &  Fait. & Cohe. & Info. & Safe. & Insp. & Hall.↓ & Enga. &  Fait. \\\midrule
CDial-GPT & 1.138             & 0.984                & 1.310             & 0.690              & 0.272             & 0.696             & 0.660 & 0.956             & 0.777             & 1.194            & 0.543            & 0.363            & 0.562            & 0.542\\
PLATO-2  & \textbf{1.725}    & \underline{1.610}    & \underline{1.741} & 1.239              & \textbf{0.068}    & \underline{1.392} & \underline{1.316} & \underline{1.585} & 1.387             & \underline{1.650} & 1.086           & \textbf{0.129}   & 1.244            & 1.128\\
EVA2.0    & \underline{1.690} & 1.494                & \textbf{1.743}    & 1.107              & \underline{0.077} & 1.312             & 1.292 & 1.524             & 1.275             & 1.616            & 0.961            & 0.151            & 1.150            & 1.096\\
%XDAI & 1.639 & 1.585 & 1.729 & 1.263 & 0.102 & 1.440 & 1.344 \\
\midrule
GLM10B    & 1.439             & 1.436                & 1.513             & \underline{1.249}  & 0.164             & 1.236             & 1.208 & 1.543             & \underline{1.528} & 1.570            & \underline{1.329} & 0.174           & \underline{1.324} & \underline{1.282}\\
GLM130B   & 1.232             & 1.179                & 1.378             & 1.000              & 0.257             & 0.816             & 0.784 & 1.177             & 1.128             & 1.315            & 0.954             & 0.303            & 0.852            & 0.832\\
\midrule
\smodel   & 1.660             & \textbf{1.641}       & 1.688             & \textbf{1.376}     & 0.127             & \textbf{1.440}    & \textbf{1.460} & \textbf{1.668}    & \textbf{1.624}    & \textbf{1.688}   & \textbf{1.393}    & \underline{0.134} & \textbf{1.412}  & \textbf{1.368} \\
\bottomrule
\end{tabular}
}
\end{table*}

% \begin{table*}[t]
% \newcolumntype{?}{!{\vrule width 1pt}}
% 	\newcolumntype{C}{>{\centering\arraybackslash}p{2em}}
% 	\caption{
% 		\label{tb:knowledge-humanbot} Human-evaluation on 100 knowledge-grounded human-bot chat dialogue. 
% 	}
% 	\centering 
% 	\small
% 	\renewcommand\arraystretch{1.0}
% \scalebox{0.95}{
% \begin{tabular}{lccccccc}
% \toprule
% Model       & Coherence         & Informativeness   & Safety           & Inspiration      & Hallucination↓   & Engagingness & Faithfulness \\\midrule
% CDial-GPT   & 0.956             & 0.777             & 1.194            & 0.543            & 0.363            & 0.562            & 0.542 \\
% PLATO-2    & \underline{1.585} & 1.387             & \underline{1.650} & 1.086           & \textbf{0.129}   & 1.244            & 1.128 \\
% EVA2.0      & 1.524             & 1.275             & 1.616            & 0.961            & 0.151            & 1.150            & 1.096 \\
% %XDAI & 1.625 & 1.536 & 1.706 & 1.218 & 0.126 & 1.410 & 1.296 \\
% \midrule
% GLM10B      & 1.543             & \underline{1.528} & 1.570            & \underline{1.329} & 0.174           & \underline{1.324} & \underline{1.282} \\
% GLM130B     & 1.177             & 1.128             & 1.315            & 0.954             & 0.303            & 0.852            & 0.832 \\
% \midrule
% \smodel     & \textbf{1.668}    & \textbf{1.624}    & \textbf{1.688}   & \textbf{1.393}    & \underline{0.134} & \textbf{1.412}  & \textbf{1.368} \\
% \bottomrule
% \end{tabular}
% }
% \end{table*}

\begin{figure}[htbp]
	\centering
	\subfigure[Implicit human evaluation]{\label{subfig:ranking}
		\includegraphics[width=0.23\textwidth]{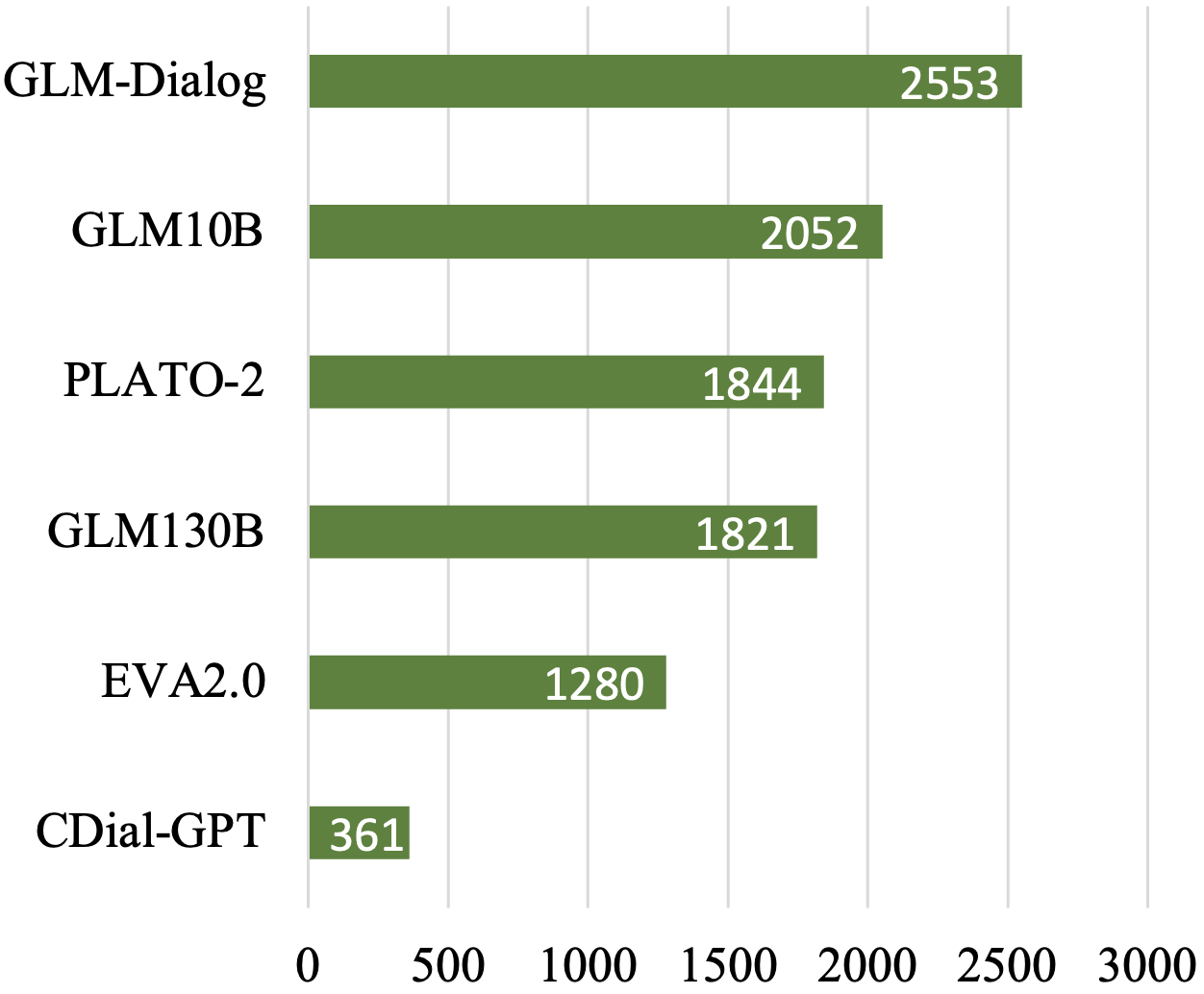}
	}
	\hspace{-0.10in}
	\subfigure[Query and search similarity]{\label{subfig:query_eval}
		\includegraphics[width=0.23\textwidth]{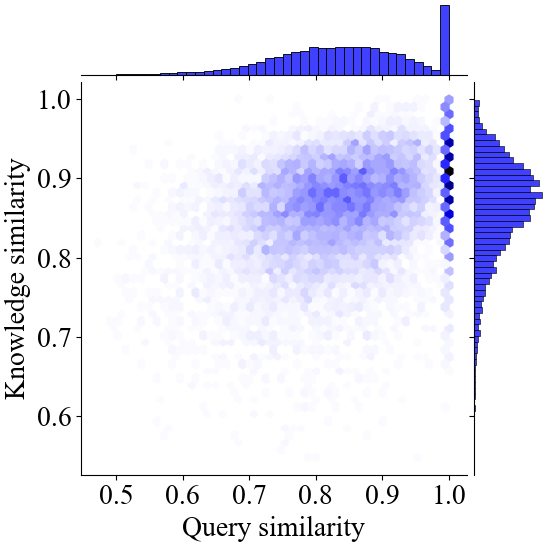}
	}		
	\caption{\label{fig:exp} (a) The evaluation of implicit human evaluation; (b) Frequency histogram and scatter plot on query/knowledge similarity between \model's produced queries/knowledge and the ground truth on DuSinc test set.}
\end{figure}

\vpara{Implicit Human Evaluation Results.}
Figure~\ref{subfig:ranking} presents the results gathered by the proposed implicit human evaluation method in~\secref{sec:implicithumanevaluation}. In total, 10,000 selections are produced by the 20 hired annotators. The annotators need to choose a response from the six deployed models to continue the conversation. Each time a response is chosen, a score is accumulated for the model which produces the response. We rank the models according to their results. The highest score is achieved by our model, which suggests that it can produce more appealing responses than the comparison models. 
The annotation bias can be effectively reduced by this evaluation method since it collects ratings implicitly through a selecting action, which is easier than explicit rating using multidimensional metrics. 

% \begin{figure}[t]
% \includegraphics[width=0.47\textwidth]{figure/ranking-hist.png}
% \caption{The selection scores of implicit human evaluation.}
% \label{fig:ranking}
% \end{figure}

% \begin{figure}[h]
% \includegraphics[width=0.38\textwidth]{figure/combine_hist.png}
% \caption{Frequency histogram and scatter plot on query/knowledge similarity between the \model's created queries/knowledge and the ground truth ones on Dusinc test set.}
% \label{fig:query_eval}
% \end{figure}

\subsection{Ablation Studies of Response Generation}

We conduct ablation tests on response generation to confirm the impact of injected external knowledge and knowledge classification, where the four major model variants include:

\begin{itemize}[leftmargin=*]
    \item \textbf{w/o stage-2 training.} We only keep training stage 1 on social media data, but delete training stage 2. The knowledge injection is also excluded during inference.
    \item \textbf{w/o knowledge injection.} Based on ``w/o stage-2 training'', we add training stage 2, but do not inject any knowledge in the online service data. 
    \item \textbf{w/o knowledge classification.} Based on ``w/o knowledge injection'', we add knowledge to the online service data, but do not classify knowledge. The knowledge is injected during inference.
    \item \textbf{w/o iterative knowledge injection.} Based on ``w/o knowledge classification'', we add the knowledge classification but remove the iterative knowledge injection. 
\end{itemize}

We conduct the human evaluation on 100 randomly selected conversations from DuSincR. To be more specific, we use each model variant to re-generate the last utterance based on the dialogue history and evaluate the generation by the same utterance-level metrics. We add knowledgeability~\cite{bao2022plato}, an additional utterance-level metric, to evaluate whether the utterance contains factual information that can be verified by the injected knowledge. The definition is given in Table~\ref{tb:humanmetric}.

Table~\ref{tb:ablationstudy} shows the effects of different components for knowledge injection, which reveals that
(1) without the 2nd training on the knowledge-grounded conversations, the model is unable to combine the injected background knowledge with the dialogue history, resulting in significant drops in all the metrics;
(2) The amount of the knowledge-grounded benchmarks is extremely limited as compared with the online gathered dialogue.
Thus, without injecting knowledge into the online large-scale service data, the knowledge integration ability mainly relies on the knowledge-grounded benchmarks, which affect the final performance;
(3) Even if we introduce knowledge into online service data, there is much noisy knowledge that is irrelevant to the response, which could have an adverse influence on response production. The performance declines as a result of the knowledge classification being removed;
(4) Without using the classifier to repeatedly sort helpful knowledge, the performance is also worse than \model.

We also compare with the following two model variants to confirm the advantage of the proposed knowledge integration way.

\begin{itemize}[leftmargin=*]
    \item \textbf{GLM10B with knowledge prompting.} We inject the same external knowledge as the proposed \smodel as the prompts on GLM10B without any fine-tuning on dialogue datasets. 
    \item \textbf{Pre-classifier.} We maintain the same training stages 1 and 2 and add external knowledge to the online service data. We train a classifier based on the human-annotated knowledge snippet of each dialogue in DuSinc and then use it to determine whether the knowledge is needed or not before injecting knowledge. The query and search processes are the same with \model.
\end{itemize}

The comparison with different knowledge integration ways yields the results in Table~\ref{tb:ablationstudy}, which reveal that 
(1) Even though the same knowledge is injected into GLM10B as prompts, the performance is poorer than the proposed \model, which demonstrates the advantages of fine-tuning;
(2) Pre-classifier decreases the performance compared with the proposed \model. The pre-classifier discards some knowledge-seeking before injection. On the contrary, \smodel injects knowledge into any dialogue. Its capacity to classify knowledge at the moment of response generation enables such complete injection, which is more suited to the real-world situation when chit-chat conversation and knowledge-grounded conversation are frequently blended.

\begin{table}[t]
\newcolumntype{?}{!{\vrule width 1pt}}
	\newcolumntype{C}{>{\centering\arraybackslash}p{2em}}
	\caption{
		\label{tb:ablationstudy} Ablation studies by human evaluation on 100 randomly selected knowledge-grounded dialogues on DuSincR. 
	}
	\centering 
	\small
	\renewcommand\arraystretch{1.0}
\scalebox{0.95}{
\begin{tabular}{@{}l?cccccc@{}}
\toprule
Model & Cohe. & Info. & Hall.↓ & Know. & Safe. & Insp.\\
\midrule
\model & \textbf{1.820} & \textbf{1.840} & \textbf{0.107} & \textbf{0.727} & \textbf{1.840} & \textbf{1.400} \\
\midrule
\multicolumn{6}{c}{Effects of different components for knowledge injection} \\
\midrule
w/o stage-2 training & 1.437 & 1.413 & 0.293 & 0.447 & 1.603 & 1.127\\
w/o know. injection & 1.527 & 1.503 & 0.223 & 0.483 & 1.687 & 1.173\\
w/o know. class. & 1.730 & 1.633 & 0.167 & 0.633 & 1.743 & 1.303\\
w/o iter. know.  & 1.757 & 1.770 & 0.137 & 0.660 & 1.810 & 1.313\\
\midrule
\multicolumn{6}{c}{Comparing with different knowledge integration ways} \\
\midrule
GLM10B w. know. & 1.563 & 1.523 & 0.227 & 0.500 & 1.623 & 1.150\\
Pre-classifier & 1.593 & 1.567 & 0.217 & 0.490 & 1.697 & 1.167\\
\midrule
\multicolumn{6}{c}{Effect of Query Generation} \\
\midrule		
w/o query generation & 1.637 & 1.593 & 0.190 & 0.523 & 1.737 & 1.187\\
\bottomrule
\end{tabular}
}
\end{table}

\subsection{Ablation Studies of Query Generation}
\label{sec:ablation}
%We conduct ablation tests on query generation to confirm the effect of the generated queries and the search results. 

% \begin{figure}[h]
% % \includegraphics[width=0.5\textwidth]{figure/new_evaluation_hist_2.jpg}
% \includegraphics[width=0.5\textwidth]{figure/query_fig.png}
% \caption{Frequency histogram of semantic similarities between the generated search queries and the ground truth queries on DuSinc test set.}
% \label{fig:query_eval}
% \end{figure}

We first create a model variant ``\textbf{w/o query generation}'' by removing the query generation step but directly using the user-posted utterance to search information snippets from the Internet. The human evaluation results of the variant are shown in Table~\ref{tb:ablationstudy}. The results demonstrate that without the generated query, the performance drops significantly, because the ellipses, coreferences, and long utterances cannot serve as a good query for search engines.

To directly demonstrate the usefulness of the generated queries, we compute the similarities between the created queries and the human-annotated actual queries on DuSinc by the cosine similarity of their embeddings produced by sentence-BERT~\cite{DBLP:journals/corr/abs-2004-13922}. 
The frequency histogram of the query similarity scores on 9,353 test cases of DuSinc is displayed in the upper part of Figure~\ref{subfig:query_eval}. The created queries are of good quality, as shown by the mean score of 0.85.% and the median score of 0.85 as well. 
%12.7\% of the generated queries have a score larger than $0.99$, resembling the generated query identical to the given ground truth query. %For the 5 score sections, 30.3\% of the generated queries fall in the score range of 0.9 to 1.0, 37.6\% in the range 0.8 to 0.9, 25.3 \% in the range 0.7 to 0.8, 6.3\% in the range 0.6 to 0.7, and 0.5\% in the range 0.5 to 0.6. 

% \begin{figure}[h]
% \includegraphics[width=0.5\textwidth]{figure/knowledge_fig.png}
% \caption{Frequency histogram of semantic similarities between the web knowledge retrieved from Baidu search engine and the ground truth web knowledge on DuSinc test set.}
% \label{fig:search_eval}
% \end{figure}

We determine the similarities between the retrieved knowledge snippets  and the human-annotated knowledge snippets provided in DuSinc test cases using the same query similarity computing method.
The frequency histogram of these knowledge similarity scores on 9,353 test cases is displayed in the right part of Figure~\ref{subfig:query_eval}. The mean score of 0.86 %and the median score of 0.87 
indicates that the retrieved knowledge is of high quality. 
It is also shown from the figure that query quality is positively correlated with knowledge quality.
%29.0\% of the generated queries have the score larger than $0.9$, resembling the retrieved web knowledge very similar to the given web knowledge in DuSinc. 
More examples of query generation and search results are shown in Appendix~\ref{sec:querygeneration} and~\ref{sec:webknowledge}.

% \zj{Haohua: Please add this analysis results.}

% \subsection{Case Studies}
% Appendix~\ref{sec:casestudy} presents the responses of \smodel for four example dialogues with noisy knowledge and four dialogues with helpful knowledge. We can observe that \smodel is capable of producing rational responses that could be verified by helpful knowledge while ignoring the noisy knowledge.

\subsection{Online Statistics}
\vpara{User Involvement.}
We deploy \smodel as a WeChat official account named ``\begin{CJK*}{UTF8}{gbsn}AI小呆爱聊天/小知呆\end{CJK*} (AI XDAI likes chatting / knowledge XDAI)'' to enable both one-one and group conversations with it and other bots. From January 12th, 2023 to February 1st, 2023, over 100 users have created 34 single chats and 63 group chats, resulting in 837 dialogue sessions in total, with an average of 50 utterances per session and an average of 22 tokens per utterance.

\vpara{Efficiency.}
We analyze the online time cost of \smodel by comparing the average time cost with GLM10B without knowledge injection and pre-classifier introduced in \secref{sec:ablation} on the same 100 selected conversations from DuSincR for ablation studies.
Table~\ref{tb:efficiency} shows the time cost of different stages, where that of the individual steps is recorded on the server side and the overall time, which also includes network latency, is recorded on the client side.
Compared with GLM10B, \smodel takes an additional 1.09 and 0.92 seconds to build the query and complete the search, respectively, which can meet the needs of an online service. As opposed to pre-classifier, we do not need to classify whether the knowledge is needed beforehand, saving an average of 0.47 seconds of time. However, the pre-classifier discards the creation and search of queries if the classifier determines that the information is not needed, the average query and search time are saved, leading to modest gains in the overall time cost. 

\begin{table}[t]
\newcolumntype{?}{!{\vrule width 1pt}}
	\newcolumntype{C}{>{\centering\arraybackslash}p{2em}}
	\caption{
		\label{tb:efficiency} Average online time cost of different stages (second). 
	}
	\centering 
	\small
	\renewcommand\arraystretch{1.0}
\scalebox{0.95}{
\begin{tabular}{@{}l@{~}ccccc@{}}
\toprule
Model     & Know. Class.    & Query Gen.    & Search    & Response  & Overall \\
\midrule
GLM10B           & -    &  -  &  - &  1.73 & 2.25 \\
Pre-classifier   & 0.47 & 0.79  & 0.68  & 1.62 & 4.17 \\
\model          &  -   & 1.09  & 0.92  & 1.64 & 4.22\\ 
\bottomrule
\end{tabular}
}
\end{table}

\section{Conclusion}
\label{sec:conclusion}
We present a 10B parameter LLM for knowledge-grounded dialogue generation. 
The model deals with the challenges of limited datasets by offering a series of augmentation and training techniques for exploiting helpful and noisy knowledge. 
We also develop a new human evaluation tool that allows humans to evaluate bots implicitly while interacting with them. 
We anticipate that the proposed techniques could inspire the researchers of interest to prompt the development of the knowledge-grounded dialogue LLM. We hope the published dataset, model, code, and evaluation tool can provide an easy-to-use and cost-effective solution for industrial developers to create various knowledge-grounded applications easily.

\clearpage
\bibliographystyle{ACM-Reference-Format}
\bibliography{acmart.bib}

\section{Appendix}
\label{sec:appendix}
\subsection{Training Data}
\label{sec:dataset}
Table~\ref{tb:trainingdatastatistics} presents the statistics of the datasets used for different training stages.

\subsection{DuSincR}
\label{sec:dusincR}

\vpara{Ellipses and Coreferences.}
Table~\ref{tb:coreference} shows examples of ellipses and coreferences in utterances.

\begin{figure}[h]
\includegraphics[width=0.3\textwidth]{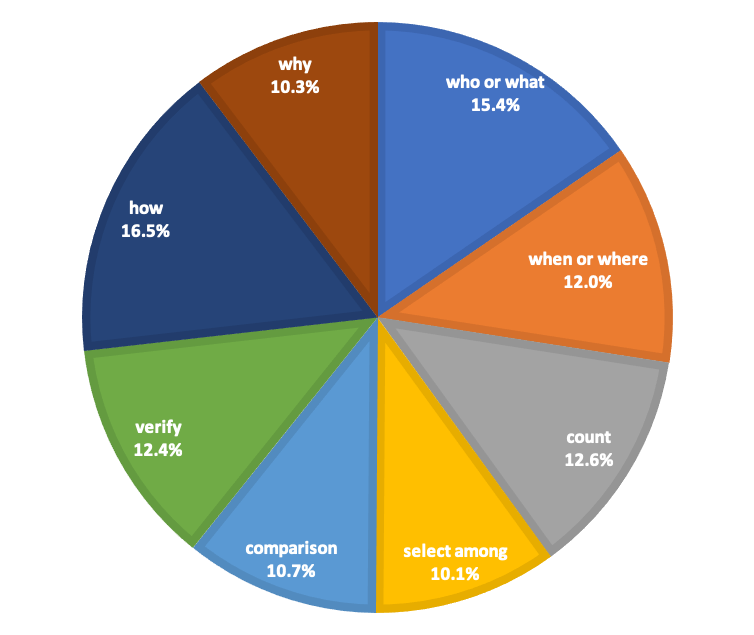}
\caption{The distribution of question types in DuSincR.}
\label{fig:dusincdistribution}
\end{figure}

\vpara{Question Types.}
Table~\ref{tb:questiontype} shows examples for eight types of questions, including asking entities (what, who), asking attributes (when, where), count, comparison, select among, verify, how, and why. Figure~\ref{fig:dusincdistribution} shows the distribution of question types in DuSincR.

\vpara{Annotation Way.}
Table~\ref{tb:annotation} presents a dialogue example to illustrate how to add new utterances into the original dialogue of DuSinc with the above question types or ellipses and coreferences.

\subsection{Automatic Evaluation Metrics}
\label{sec:automatic_metric}
We provide the following explanations for each automatic metric.

\begin{itemize}
    % \item \textbf{Dist-n}
    
    % Distinct is used to evaluate the diversity of the generated text and Dist-n is defined as:
    
    % \beq{
    %     Dist\text{-}n = \frac{ \#(unique\text{ }n\text{-}gram)}{\#(n\text{-}gram)}
    % }
    
    % \noindent where $\#(n\text{-}gram)$ denotes the total number of n-grams in a generated text and $\#(unique \text{ }n\text{-}gram)$ denotes the number of the uniquely observed n-grams in it.

    \item \textbf{BLEU-N}
    
    BLEU is used to evaluate the precision of the generated text comparing with the reference text. 
    BLEU-N combines the values of BLEU for different n-grams, \emph{i.e.},
    
    \beqn{ 
       BLEU\text{-}N &=& BP \cdot \exp \left( {\sum\limits_{n = 1}^N W_n \cdot \log{p_n}} \right),  \\
       BP &=& \begin{cases}1, & lc > lr \\ \exp{\left( \frac{1 - lr}{lc} \right)}, & lc \leq lr \end{cases}, \\
       p_n &=& \frac{\#\{correctly\text{ }predicted\text{ }n\text{-}gram\}}{\#\{predicted\text{ }n\text{-}gram\}},
    }
        
    \noindent where $p_n$ is the precision of n-gram, \emph{i.e.}, the percentage of the predicted n-grams  that are found in the reference text.  The term $W_n$ refers to the weight of n-gram, which is typically specified to be uniform weight, \emph{i.e.}, $W_n = \frac{1}{N}$ for any $n$. $BP$ is the penalty factor. $BP$ is less than 1 if the  predicted length $lc$ is less than the reference length $lr$.

    \item \textbf{F1}

    The F1 score can be regarded as a harmonic average of accuracy and recall, with a maximum value of 1 and a minimum value of 0. 

    \beq{
      F_1 = 2 \cdot \frac{p_1 \cdot r_1}{p_1 + r_1},
    }

    \noindent where $p_1$ and $r_1$ denote the precision and recall of the correctly predicted 1-gram respectively. 

    \item \textbf{Rouge-x}

    Rouge prioritizes recall over accuracy. It counts how many n-grams from the reference text are present in the generated text.
    Rouge-n is defined as:
    
    \beq{
    Rouge\text{-}n = \frac{ \#\{correctly\text{ }predicted\text{ }n\text{-}gram\}}{ \#\{n\text{-}gram\text{ }in\text{ }reference\text{ }text\}}
    }

    ROUGE-L computes the rouge value of the longest common subsequence (LCS) between the generated text and the reference text. We denote LCS as $L$. ROUGE-L is computed as follows:

    \beqn{
     Rouge\text{-}L &=& \frac{(1 + \beta^2) r_{LCS} p_{LCS}}{r_{LCS} + \beta^2 p_{LCS}},\\
     p_{LCS} &=& \frac{\#\{1\text{-}gram\text{ }in\text{ }L\}}{\#\{1\text{-}gram\text{ }in\text{ }generated\text{ }text\}}, \\
     r_{LCS} &=& \frac{\#\{1\text{-}gram\text{ }in\text{ }L\}}{\#\{1\text{-}gram\text{ }in\text{ }reference\text{ }text\}},
    }

    \noindent where $\beta$ is a weighting coefficient and $p_{LCS}$ and $r_{LCS}$ stand for the precision and recall of $L$, respectively. Rouge-L will focus more on recall rate rather than accuracy rate if $\beta$ is greater. Here, $\beta$ is set to 1.2.

    \item \textbf{Bert-Score}

   Bert-score is used to calculate the similarity between the generated text and the reference text. To be more precise, it generates a similarity matrix by first computing the inner product between each word in the two texts based on the BERT embeddings. Then, using this matrix, it computes the precision and recall by averaging the maximal similarity scores of the reference and generated texts weighted by word idf value. In the end, we combine them to report F1 of Bert-Score.
    
\end{itemize}

\subsection{Chit-chat Opening Utterances}
\label{sec:chitchatopeningutterance}
We present the designed 50 chit-chat opening utterances with 25 positive, 12 negative, and 13 neutral statements in Table~\ref{tb:chitchatopeningutterance}.

\subsection{Knowledge-grounded Opening Utterances}
\label{sec:knowledgeopeningutterance}
We present the designed 100 knowledge-grounded opening utterances in Table~\ref{tb:knowledgeopeningutterance}. These utterances involve 14 topics related to entertainment, 14 topics related to life, 12 topics related to history and culture, 10 topics related to education, 12 topics related to health, 12 topics related to sports, 13 topics related to science and technology, and 13 topics related to finance. Additionally, they can be broken into the types of ``what", ``who", ``where", ``when", ``how", ``why", ``compare", ``count", and ``verify".

\subsection{Human Evaluation Metrics}
\label{sec:humanmetric}
In Table~\ref{tb:humanmetric}, we define each human evaluation metric's values and their accompanying meanings.

\subsection{Implicit Human Evaluation Tool}
\label{sec:implHumanEvalTool}
\begin{figure*}[t]
\includegraphics[width=0.9\textwidth]{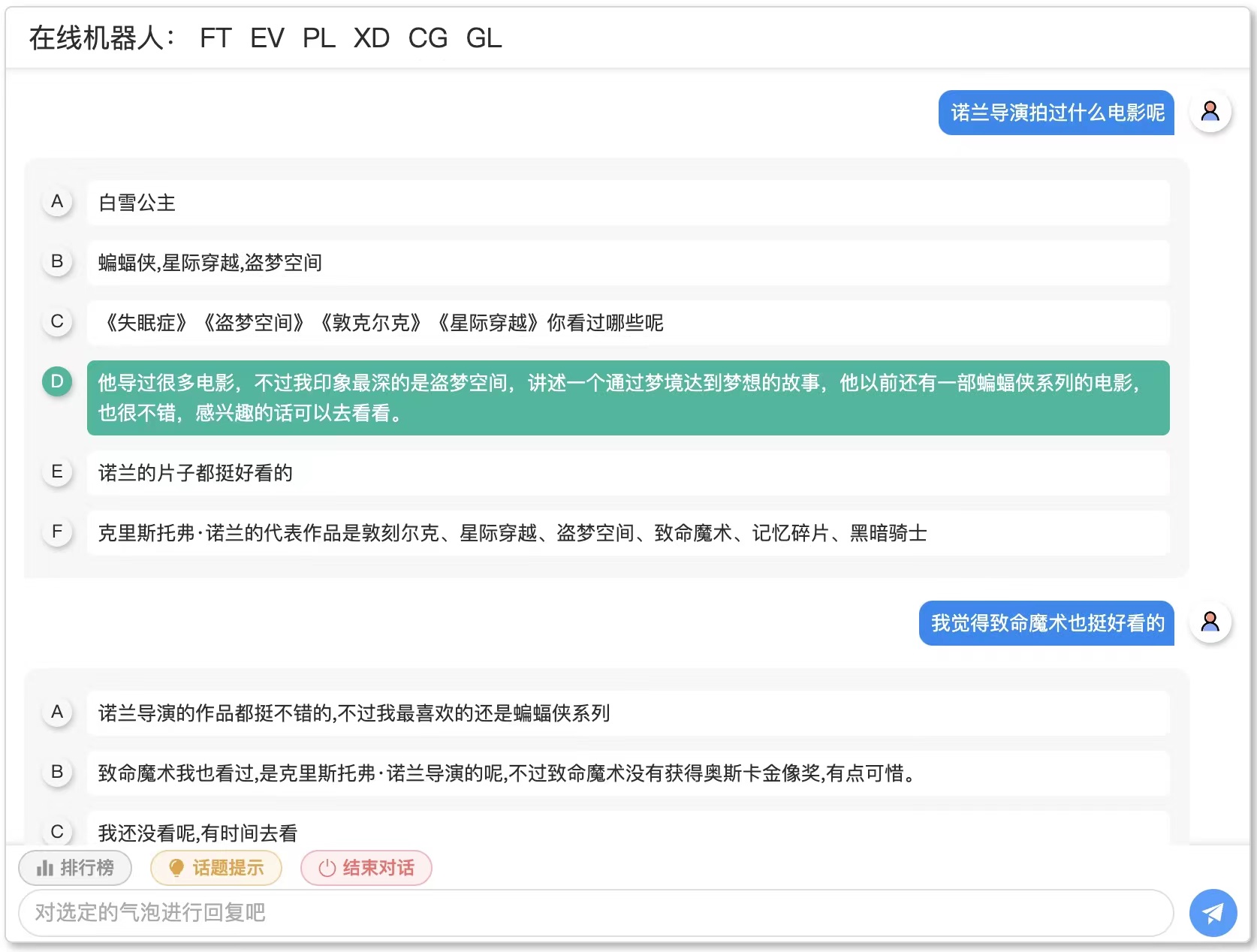}
\caption{An screenshot of our online implicit human evaluation tool. Users are allowed to interact with six deployed anonymous bots and evaluate their performance by selecting a single response from all their responses to continue the conversation.}
\label{fig:screenshot}
\end{figure*}

Figure~\ref{fig:screenshot} shows the screenshot of our online implicit human evaluation tool.

\subsection{Query Generation Examples}
\label{sec:querygeneration}

Our query generation module can successfully generate a complete query according to the given dialogue history, varying from different question types or ellipses and coreferences. We provide examples on DuSinc for each similarity score range, and then highlight examples where the dialogue history includes coreference, ellipsis, and complete query. We also provide examples on DuSincR of 8 different question types, better representing the effectiveness of our query generation module on dialogues that consist of coreference or ellipsis.
 
\vpara{Examples of Different Score Ranges on DuSinc.} 
Table~\ref{tb:QueryGen5} to Table~\ref{tb:QueryGen9} present 3 examples for each similarity score range tested on DuSinc. Each example includes the dialogue history (with the most recent user-posted utterance), the ground-truth query, and the produced query. The similarity score is computed between the produced query and the ground-truth query.

\vpara{Examples of Coreference Dialogues on DuSinc.}
Table~\ref{tb:QueryGenCoreference} presents example queries generated from dialogues which consist of coreference on DuSinc. The coreference is referenced in the dialogue as a ``underline'' (and is always in the most recent user-posted utterance); a similarity score is provided at the end of each example.

\vpara{Examples of Ellipsis Dialogues on DuSinc.}
Table~\ref{tb:QueryGenEllipse} presents example queries generated from dialogues which consist of ellipsis on DuSinc. The sentence containing ellipsis is referenced in the dialogue (which is always in the most recent user-posted utterance); a similarity score is provided at the end of each example.

\vpara{Examples of Complete Query Dialogues on DuSinc.}
Table~\ref{tb:QueryGenComplete} presents example queries generated from dialogues which consist of a complete query on DuSinc. The complete query is referenced in the dialogue (which is always in the most recent user-posted utterance); a similarity score is provided at the end of each example. 

\vpara{Examples of 8 Different Question Types on DuSincR.}
Table~\ref{tb:dusincrtype1} to Table~\ref{tb:dusincrtype8} present 3 examples for each question type tested on DuSincR, which contains ellipsis or coreference in every test case. Each example includes the dialogue history (with the most recent user-posted utterance) and the produced query. The ellipsis or coreference is referenced in the dialogue (which is always in the most recent user-posted utterance).

\subsection{Search Result Examples}
\label{sec:webknowledge}

Table~\ref{tb:WebKnow5} to Table~\ref{tb:WebKnow9} present 3 examples for each similarity score range tested
on DuSinc. Each example includes the query used to search on web, the ground-truth web knowledge provided in DuSinc, and the web knowledge retrieved from Baidu search. The similarity score is computed between the retrieved knowledge and the ground-truth knowledge. It should be noted that the provided ground-truth query from DuSinc and the produced query are the same in the examples provided from Table~\ref{tb:WebKnow5} to Table~\ref{tb:WebKnow9}, therefore only one search query is provided, and the web knowledge given and retrieved are comparable. 

\subsection{Case Studies}
\label{sec:casestudy}
Table~\ref{tb:CaseStudyBad} shows four examples of generated responses given the noisy knowledge snippet being injected. 
Table~\ref{tb:CaseStudygood} shows four examples of generated responses given the helpful knowledge snippet being injected. 

\begin{table}[ht]
\newcolumntype{?}{!{\vrule width 1pt}}
	\newcolumntype{C}{>{\centering\arraybackslash}p{2em}}
	\caption{
		\label{tb:trainingdatastatistics}  Training data statistics. We state the number of sessions for dialogue data and the number of QA pairs for question answering data.
	}
	\centering 
	\small
	\renewcommand\arraystretch{1.0}
\begin{tabular}{l?c?c?c}
\toprule
Type & Dataset & Characteristic & Size \\
\midrule
 \multirow{5}{*} {\tabincell{c}{Social\\Media\\Data}}  
 & Weibo       & Blog~\cite{wang2020large}                 & 6.8M  \\
 & Bilibili    & Video sharing        & 10K   \\
 & Baidu Tieba & Online discussion    & 300K \\
 & Zhihu~\cite{bright_xu_2019_3402023} & Online discussion & 4.1M \\
 & Douban      & Online discussion    & 280K \\
 \midrule
 \multirow{7}{*} {\tabincell{c}{Benchmark\\Data}}  
 & KDConv~\cite{zhou2020kdconv}         & Knowledge-grounded     & 4.5K \\
& DuConv~\cite{wu2019proactive}         & Knowledge-grounded    & 20K \\
 & NaturalConv~\cite{wang2021naturalconv} & Knowledge-grounded   & 20K \\
 & DuSinc~\cite{zhou2022Dusic}           & Knowledge-grounded    & 8K \\
 & WebQA~\cite{li2016dataset}           & Question answering     & 42K \\
 & Dureader~\cite{he2017dureader}       & Question answering     & 200K \\
 & DRCD~\cite{shao2018drcd}                         & Question answering     & 30K \\
 \midrule
 \tabincell{c}{Online\\Service \\ Data}
 & XDAI~\cite{yu2022xdai}                               & \tabincell{c}{from Sept 1st, 2022\\ to Dec 15th, 2022} & 800K
 \\
\bottomrule
\end{tabular}
\end{table}

\begin{CJK*}{UTF8}{gbsn}

\begin{table*}[t]
\newcolumntype{?}{!{\vrule width 1pt}}
	\newcolumntype{C}{>{\centering\arraybackslash}p{2em}}
	\caption{
		\label{tb:coreference} Examples of ellipses and coreferences in utterances. 
	}
	\centering
	\small
	\renewcommand\arraystretch{1.0}
 
\scalebox{0.95}{
\begin{tabular}{l}
\toprule
\textbf{Utterance} \\ \midrule
你知道长江有多长吗?\\
How long is the Yangtze River? \\
长江全长6633千米，是中国最长的河流。\\
With a length of 6,633km, the Yangtze River is the longest river in China.\\
那黄河呢？\textbf{（成分缺失）}\\
What about the Yellow River? \textbf{(Ellipse)}\\ 
难不倒我，黄河长约5464千米，发源于青藏高原，是中国第二长河。\\
The Yellow River is about 5,464 kilometers long, originating from the Qinghai-Tibet Plateau. It is the second longest river in China.\\
那它的入海口在哪里呢？\textbf{（指代消解）}\\
And where is its estuary? \textbf{(Coreference)}\\ 

\bottomrule
\end{tabular}
}

\end{table*}

\end{CJK*}

\begin{CJK*}{UTF8}{gbsn}

\begin{table*}[t]
\newcolumntype{?}{!{\vrule width 1pt}}
	\newcolumntype{C}{>{\centering\arraybackslash}p{2em}}
	\caption{
		\label{tb:questiontype} Examples of question types created for DuSincR. 
	}
	\centering
	\small
	\renewcommand\arraystretch{1.0}
 
\scalebox{0.95}{
\begin{tabular}{cl}
\toprule
\makebox[2cm][c]{\textbf{Question Type}} & \makebox[13cm][c]{\textbf{Utterance}} \\ \midrule
\multirow{2}{*}{What}
& 01. 中国的首都是哪座城市？\\
& \hspace{1.58em}Which city is the capital of China? \\ \midrule

\multirow{2}{*}{Who}
& 02. 《百年孤独》是我很喜欢的一本书，你知道这本书的作者是谁吗？ \\
& \hspace{1.58em}\emph{One Hundred Years of Solitude} is one of my favorite books. Do you know the author of this book?\\ \midrule

\multirow{2}{*}{Where}
& 03. 2022年冬奥会是在哪里举办的？ \\
& \hspace{1.58em}Where was the 2022 Winter Olympics held? \\ \midrule

\multirow{2}{*}{When}
& 04. 什么时候去海南旅游最合适？ \\
& \hspace{1.58em}When is the best time to visit Hainan? \\ \midrule

\multirow{2}{*}{Count}
& 05. 迈克杰克逊一共出过多少专辑？ \\
& \hspace{1.58em}How many albums did Michael Jackson make? \\ \midrule

\multirow{4}{*}{Comparison}
& 06. 我想知道世界上最长的河流是哪一条？ \\
& \hspace{1.58em}What is the longest river in the world? \\
& 07. 你知道儿童文学和成人文学有什么区别吗？ \\
& \hspace{1.58em}What's the difference between children's literature and adult literature？ \\ \midrule

\multirow{2}{*}{Verify}
& 08. 我最近想做一些有氧运动，跑步是不是一种有氧运动？ \\
& \hspace{1.58em}I want to do some aerobic exercise recently. Is running an aerobic exercise? \\ \midrule

\multirow{2}{*}{How}
& 09. 我想做一个奶油蛋糕，你知道具体该怎么做吗？ \\
& \hspace{1.58em}I want to make a cream cake. Do you know exactly how to make it?？ \\ \midrule

\multirow{2}{*}{Why}
& 10. 《红楼梦》被称为四大名著之首，为什么呢？ \\
& \hspace{1.58em}Why is \emph{the Dream of Red Mansions} called the first of the Four Great Classics？ \\ 
\bottomrule
\end{tabular}
}

\end{table*}

\end{CJK*}

\begin{CJK*}{UTF8}{gbsn}

\begin{table*}[t]
\newcolumntype{?}{!{\vrule width 1pt}}
	\newcolumntype{C}{>{\centering\arraybackslash}p{2em}}
	\caption{
		\label{tb:annotation} An illustration of a dialogue session from the DuSincR dataset. A-1 to B-5 are the original DuSinc dialogue session, while A-6 and B-6 are the newly annotated utterances with sentence coreference and the comparison question type.
	}
	\centering
	\small
	\renewcommand\arraystretch{1.0}
 
\scalebox{0.95}{
\begin{tabular}{l}
\toprule
\makebox[13cm][c]{\textbf{A Dialogue Session}} \\ \midrule

A-1 我不会贴假睫毛啊，好难。\\
\hspace{1.58em}It is hard for me to apply false eyelashes. \\ 
B-1 这个我专门了解过，先把真睫毛涂一层睫毛膏卷起来定型，然后把假睫毛剪下来揉软，再把睫毛贴上胶水贴到眼皮上。\\
\hspace{1.58em}First, apply mascara to the real lashes and curl them to shape them. Then, cut off the false lashes and knead them to soften them.
\hspace{1.58em}\\ Then, attach the lashes to the eyelids with glue. \\ 
A-2 哇！好厉害啊，你会贴么？\\
\hspace{1.58em}That's amazing. Can you do that? \\ 
B-2 哈哈哈，我只会说。但是实际操作也不会。\\
\hspace{1.58em}Actually, neither can I. \\ 
A-3 哈哈，我视频倒是看了很多。但是手残。\\
\hspace{1.58em}I do watch a lot of videos. But I still can't do it. \\ 
B-3 我也是！手残党啊。\\
\hspace{1.58em}Neither do I. \\ 
A-4 假睫毛会对人有伤害么？\\
\hspace{1.58em}Can false eyelashes hurt people? \\ 
B-4 用胶水很容易造成眼睛发炎，所以一定要小心。\\
\hspace{1.58em}Using glue can easily cause eye irritation, so be careful. \\ 
A-5 嗯嗯，看来还是去接睫毛好一点。\\
\hspace{1.58em}I think it's better to apply the eyelash extension. \\ 
B-5 是的。我也打算去呢。\\
\hspace{1.58em}Me too. \\  
\midrule
\textbf{The newly annotated utterances}\\
\midrule
A-6 那\underline{这个}\textbf{(coreference)}和贴假睫毛有什么区别？ \textbf{(comparison question type)} \\
\hspace{1.58em}What's the difference between them?\\ 
B-6 接睫毛可能会因为清洁不到位导致眼睛发痒。贴假睫毛会显得不自然并且不能重复多次使用。 \\
\hspace{1.58em}Eyelash extensions may cause itchy eyes due to poor cleaning. False eyelashes look unnatural and can not be reused.\\ 

\bottomrule
\end{tabular}
}

\end{table*}

\end{CJK*}

\begin{CJK*}{UTF8}{gbsn}
\begin{table*}[t]
\newcolumntype{?}{!{\vrule width 1pt}}
	\newcolumntype{C}{>{\centering\arraybackslash}p{2em}}
	\caption{
		\label{tb:chitchatopeningutterance} 50 chit-chat opening utterances with 25 positive, 12 negative, and 13 neutral statements.
	}
	\centering 
	\small
	\renewcommand\arraystretch{1.0}
\scalebox{0.95}{
\begin{tabular}{cll}
\toprule

\textbf{Type} & \makebox[6cm][c]{\textbf{Utterance}} & \makebox[8cm][c]{\textbf{Translated Utterance}} \\ \midrule
\multirow{25}{*}{Positive}
& 01. 你有喜欢的歌手么？ & What is your favorite singer? \\
& 02. 我舅舅家养了一只小猫，好可爱啊。 & My uncle raised a kitten. It's so cute. \\
& 03. 疫情结束了我要到处旅游！ & I hope to travel after the epidemic!  \\
& 04. 你人生中有什么高光时刻么？ & Have you had any highlight moments in your life? \\
& 05. 我最近在跑步，有很大进步！ & I've been running recently and I've made great progress! \\
& 06. 你理想的另一半是什么样？ & What would your ideal partner be like? \\
& 07. 今天阳光明媚，又想出去玩了。 & It's sunny today, and I'd like to go out and play. \\
& 08. 老食堂的口水鸡味道不错。 & The drool chicken in the old canteen tastes good. \\
& 09. 想谈恋爱了怎么办？ & What should I do if I want to fall in love? \\
& 10. 我好想去音乐节看一看呀。 & I really want to go to the music festival. \\
& 11. 刚用医保买了维生素，好便宜。 & I just bought vitamins with medical insurance and it was so cheap. \\
& 12. 我平时最喜欢锻炼身体，你呢？ & I like exercise most, how about you? \\
& 13. 你平时会玩乐器吗？ & Do you usually play musical instruments? \\
& 14. 我喜欢刷短视频。 & I love watch short videos. \\
& 15. 夏天要来了，我需要减肥了。 & Summer is coming and I need to lose weight. \\
& 16. 现在公园里赏花的人可多了呢。 & There are many people enjoying flowers in the park now. \\
& 17. 外面的小鸟叫得好好听。 & The birds outside are singing nicely. \\
& 18. 春天来了，外面好多好看的花呀。 & Spring is coming, and there are so many beautiful flowers outside. \\
& 19. 你喜欢看什么综艺啊？ & What variety shows do you like? \\
& 20. 我发现自己变帅了。 & I found myself handsome these days. \\
& 21. 这两天黄金涨得好多啊。 & The price of gold has risen a lot in the past two days. \\
& 22. 我刚刚买了一个拼图，要不要一起拼？ & I just bought a jigsaw puzzle, do you want to play with me together? \\
& 23. 出生在一个幸福的家庭是多么美好的事。 & What a wonderful thing to be born into a happy family. \\
& 24. 我太喜欢糖油混合物了。 & I love the sugar and oil mixture so much. \\
& 25. 有点想念大学的室友。 & I kind of miss my college roommates. \\ \midrule
\multirow{14}{*}{Neural}
& 26. 你对星座有研究么？ & Have you studied astrology? \\
& 27. 你会做饭吗？ & Can you cook? \\
& 28. 你要去超市吗？ & Are you going to the supermarket? \\
& 29. 想学一个乐器。 & I want to learn a musical instrument. \\
& 30. 想问下你平时都是怎么理财的？ & May I ask how do you manage your money? \\
& 31. 今年小区又栽了很多树。 & Many trees have been planted in the community this year. \\
& 32. 今年暑假你准备回家吗？ & Are you going home this summer vacation? \\
& 33. 你在生活中有没有经常听到适量这个词？ & Have you heard the word moderation a lot in your life? \\
& 34. 今天心情怎么样呀？ & How are you feeling today? \\
& 35. 怎么才能在不尴尬不破坏和舍友 & How can I make my roommates pay attention to hygiene$\hookleftarrow$ \\
& \hspace{1.32em} 关系的情况下让舍友注意卫生？ & without embarrassment or damaging the relationship with them? \\
& 36. 睡前你会听点音乐吗？ & Do you listen to some music before bed? \\
& 37. 你平时都用什么社交软件啊？ & What social software do you usually use? \\
& 38. 你染过头发吗？ & Have you ever dyed your hair? \\ \midrule
\multirow{15}{*}{Negative}
& 39. 我最近被我的室友无语到了， & I was so speechless by my roommate's behavior recently$\hookleftarrow$ \\
& \hspace{1.54em}真的都想换宿舍了。 & that I really wanted to change dorms. \\
& 40. 如果人生不能实现自己的梦想怎么办？ & What if you can't realize your dreams in life? \\
& 41. 赚钱好难啊！ & It's so hard to make money! \\
& 42. 考研失败了，我得准备找工作了。 & I failed the postgraduate entrance examination,$\hookleftarrow$ \\
& & and I have to prepare to find a job. \\
& 43. 最近手机好卡啊。 & It's hard to use my phone recently. \\
& 44. 婴儿零食，加上婴儿两个字就价格翻番。 & Baby snacks, adding the word baby will double the price. \\
& 45. 最近时间充足，想学点什么， & I have enough time recently and want to learn something,$\hookleftarrow$ \\
& \hspace{1.54em}但是没有眉目。 & but I have no idea. \\
& 46. 我昨晚做噩梦了。 & I had a nightmare last night. \\
& 47. 听我说呀，就业可得要谨慎。 & Listen to me, you have to be cautious about employment. \\
& 48. 无论几点睡，都是4点醒怎么办？ & No matter what time you go to sleep, $\hookleftarrow$ \\
& & what should you do if you wake up at 4 o'clock? \\
& 49. 我感觉自己拖延症好严重啊！ & I feel like I'm procrastinating so badly! \\
& 50. 现在的时尚我真是欣赏不来。 & I really can't appreciate the current fashion. \\
\bottomrule
\end{tabular}
}
\end{table*}

\end{CJK*}

\begin{CJK*}{UTF8}{gbsn}

\begin{table*}[t]
\newcolumntype{?}{!{\vrule width 1pt}}
	\newcolumntype{C}{>{\centering\arraybackslash}p{2em}}
	\caption{
		\label{tb:knowledgeopeningutterance} Knowledge-grounded utterances of entertainment. 
	}
	\centering
	\small
	\renewcommand\arraystretch{1.0}
 
\scalebox{0.95}{
\begin{tabular}{cl}
\toprule
\makebox[2cm][c]{\textbf{Question Type}} & \makebox[13cm][c]{\textbf{Utterance}} \\ \midrule
\multirow{3}{*}{What}
& 01. 最近在读《明朝那些事》，太有意思了。我有点忘了，朱元璋之后就是朱棣继位吧？\\
& \hspace{1.58em}Recently, I'm reading \emph{The story in the Ming Dynasty}, which is very interesting. \\
& \hspace{1.58em}I kind of forgot, after Zhu Yuanzhang, is Zhu Di succeeding to the throne, right? \\ \midrule
\multirow{6}{*}{Who}
& 02. 埃隆马斯克的母亲好像也非常有名，你知道她吗？ \\
& \hspace{1.58em}Elon Musk's mother seems to be very famous too, do you know her? \\
& 03. 我在看《康熙王朝》，男主角很有魅力，你知道是谁演的吗？ \\
& \hspace{1.58em}I'm watching \emph{Kangxi Dynasty}, the actor is very attractive, do you know who played it? \\
& 04. 《甄嬛huán传》的歌都是谁唱的呢 \\
& \hspace{1.58em}Who sang the songs of \emph{The Legend of Zhen Huan}? \\ \midrule
\multirow{4}{*}{Where}
& 05. 你知道五月天最近一次是在哪里开的演唱会吗？ \\
& \hspace{1.58em}Do you know where the concert of Wuyuetian was held? \\
& 06. 你平常都在哪个商场买衣服呀，最近入冬想买几件。 \\
& \hspace{1.58em}Which shopping mall do you usually buy clothes in? I want to buy some for the coming winter. \\ \midrule
\multirow{5}{*}{When}
& 07. 玉渊潭的樱花什么时候开呀，想去看看。 \\
& \hspace{1.58em}When will the cherry in Yuyuantan bloom? I'd like to go and enjoy. \\
& 08. 最近有个很火的电视节目，是有关博物馆文物的，你知道它什么时候播出吗？ \\
& \hspace{1.58em}Recently there is a very popular TV program about cultural relics in museums. \\
& \hspace{1.58em}Do you know when it will be broadcast? \\ \midrule
\multirow{2}{*}{How}
& 09. 你常说的那家鸭馆怎么样呀，我也想撸鸭了。 \\
& \hspace{1.58em}How about the duck room you often talk about? I want to play with them too. \\ \midrule
\multirow{2}{*}{Why}
& 10. 为什么每年都有那么多人去长城玩啊，我觉得没什么意思。 \\
& \hspace{1.58em}Why do so many people go to the Great Wall every year? I don't think it's interesting. \\ \midrule
\multirow{2}{*}{Comparison}
& 11. 《甄嬛huán传》和《如懿传》，哪个更好看呢？ \\
& \hspace{1.58em}\emph{The Legend of Zhen Huan} or \emph{Ruyi's Royal Love in the Palace}, which one is better? \\ \midrule
\multirow{2}{*}{Count}
& 12. 哆啦A梦共有几部电影呢？ \\
& \hspace{1.58em}How many Doraemon movies are there? \\ \midrule
\multirow{4}{*}{Verify}
& 13. 《肖申克的救赎》现在还是IMDb榜单TOP1吗？ \\
& \hspace{1.58em}Is \emph{The Shawshank Redemption} still on the IMDb list TOP1? \\
& 14. 育碧发行的新游戏风评不太好，你玩了之后感觉怎么样？ \\
& \hspace{1.58em}The new game released by Ubisoft has not been well received. How do you feel after playing it? \\
\bottomrule
\end{tabular}
}

\end{table*}

\begin{table*}[t]
\newcolumntype{?}{!{\vrule width 1pt}}
	\newcolumntype{C}{>{\centering\arraybackslash}p{2em}}
	\caption{
		\label{tb:automaticKdConv}  Knowledge-grounded utterances of life. 
	}
	\centering
	\small
	\renewcommand\arraystretch{1.0}

\scalebox{0.95}{
\begin{tabular}{cl}
\toprule
\makebox[2cm][c]{\textbf{Question Type}} & \makebox[13cm][c]{\textbf{Utterance}} \\ \midrule
\multirow{3}{*}{What}
& 15. 你知道三鹿奶粉事件嘛？这对于许多育儿的妈妈来说，简直是噩梦啊！\\
& \hspace{1.58em}Do you know about the Sanlu milk powder incident? \\
& \hspace{1.58em}It's a nightmare for many parenting mothers! \\ \midrule
\multirow{4}{*}{Where}
& 16. 优衣库的设计还不错啊，这是哪个国家的牌子？ \\
& \hspace{1.58em}Uniqlo's design is quiet good. Which country's brand is this? \\
& 17. 北京的秋天好美，去哪里赏红叶呢？ \\
& \hspace{1.58em}Autumn in Beijing is so beautiful. Where can I go to enjoy the red leaves? \\ \midrule
\multirow{2}{*}{When}
& 18. 最近馋螃蟹了，你知道啥时候吃最鲜美吗？ \\
& \hspace{1.58em}Recently, I have been greedy for crabs. Do you know what is the best time to to eat crabs? \\ \midrule
\multirow{7}{*}{How}
& 19. 现在流行牙线啊，你知道咋用吗？ \\
& \hspace{1.58em}Dental floss is popular now, do you know how to use it? \\
& 20. 我想以后去当个心理咨询师，不知道现在能做哪些准备，是不是还要考什么证啊？ \\
& \hspace{1.58em}I want to become a psychological counselor in the future, but I don't know \\
& \hspace{1.58em}what preparations I can make now, do I need to take any other certificates? \\
& 21. 我今天买了一块生牛肉，怎么做比较好呢？ \\
& \hspace{1.58em}I bought a piece of raw beef today, what should I do? \\ \midrule
\multirow{3}{*}{Why}
& 22. 小时候我妈老给我泡决明子菊花茶，但是现在近视还是那么深，为啥呢？ \\
& \hspace{1.58em}When I was young, my mother always made me cassia chrysanthemum tea, \\
& \hspace{1.58em}but now my eyesight is still so poor, why? \\ \midrule
\multirow{7}{*}{SelectAmong}
& 23. 最近迷恋高楼大厦，你知道世界上前十名的大厦都是哪些国家的什么楼吗？ \\
& \hspace{1.58em}I am obsessed with high-rise buildings recently. Do you know \\
& \hspace{1.58em}which are the top ten buildings in the world and what country do they belong to? \\
& 24. 我喜欢游览名胜古迹，你知道中国拥有名胜古迹最多的城市是哪个吗？ \\
& \hspace{1.58em}I like to visit places of interest, do you know which city has the most places of interest in China? \\
& 25. 快毕业了，去哪个城市发展比较好呢？ \\
& \hspace{1.58em}I am about to graduate, which city is better to develop? \\ \midrule
\multirow{5}{*}{Comparision}
& 26. 以后你想去哪定居呢？北京和深圳比，你更喜欢哪个城市？ \\
& \hspace{1.58em}Where do you want to live in the future? Compared with Beijing and Shenzhen, \\
& \hspace{1.58em}which city do you like more? \\ 
& 27. 日本和中国的动漫产业谁更强？ \\
& \hspace{1.58em}Who is stronger in animation industry, Japan or China? \\ \midrule
\multirow{2}{*}{Verify}
& 28. 神仙水的广告做的铺天盖地，真的有效果吗？ \\
& \hspace{1.58em}There are so many advertisements for SK2, is it really effective? \\
\bottomrule
\end{tabular}
}

\end{table*}

\begin{table*}[t]
\newcolumntype{?}{!{\vrule width 1pt}}
	\newcolumntype{C}{>{\centering\arraybackslash}p{2em}}
	\caption{
		\label{tb:automaticKdConv} Knowledge-grounded utterances of history and culture. 
	}
	\centering
	\small
	\renewcommand\arraystretch{1.0}

\scalebox{0.95}{
\begin{tabular}{cl}
\toprule
\makebox[2cm][c]{\textbf{Question Type}} & \makebox[13cm][c]{\textbf{Utterance}} \\ \midrule
\multirow{4}{*}{What}
& 29. 你知道美国奴隶制度的有关的历史嘛？ \\
& \hspace{1.58em}Do you know the history of American slavery? \\
& 30. 你知道中国古代的科举制度是什么吗？ \\
& \hspace{1.58em}Do you know what the imperial examination system was in ancient China? \\ \midrule
\multirow{2}{*}{Where}
& 31. 明朝建国时的首都在哪里呢？ \\
& \hspace{1.58em}Where was the capital of the Ming Dynasty when it was founded? \\ \midrule
\multirow{2}{*}{Who}
& 32. 你知道日本的战国三杰是哪些人吗？ \\
& \hspace{1.58em}Do you know who are the three masters of the Warring States Period in Japan? \\ \midrule
\multirow{4}{*}{How}
& 33. 你知道西方国家是怎么打招呼的吗？ \\
& \hspace{1.58em}Do you know how people in western countries say hello? \\
& 34. 你知道中国的传统点心是怎么做出来的吗？ \\
& \hspace{1.58em}Do you know how traditional Chinese dim sum is made? \\ \midrule
\multirow{4}{*}{Why}
& 35. 为什么欧洲人的英语都那么好啊？ \\
& \hspace{1.58em}Why are Europeans so good at English? \\
& 36. 为什么法国有那么多奢侈品牌呢？ \\
& \hspace{1.58em}Why are there so many luxury brands in France? \\ \midrule
\multirow{3}{*}{SelectAmong}
& 37. 我很喜欢历史，特别喜欢研究皇帝，你知道中国历代皇帝中谁寿命最长吗？ \\
& \hspace{1.58em}I like history very much, and I especially like studying emperors. \\
& \hspace{1.58em}Do you know who lived the longest among the emperors of all dynasties in China? \\ \midrule
\multirow{4}{*}{Comparision}
& 38. 西方文化相比于东方差异主要在哪里呢？ \\
& \hspace{1.58em}What are the main differences between Western culture and Eastern culture? \\
& 39. 据我观察，东亚地区的国家似乎更重视考试，这是为什么呢？ \\
& \hspace{1.58em}According to my observation, countries in East Asia seem to put more emphasis on exams. Why? \\ \midrule
\multirow{2}{*}{Verify}
& 40. 据说台湾方言是来自于闽南语系，是这样吗？ \\
& \hspace{1.58em}It is said that the Taiwanese dialect comes from the Hokkien language family, is that true? \\
\bottomrule
\end{tabular}
}

\end{table*}

\begin{table*}[t]
\newcolumntype{?}{!{\vrule width 1pt}}
	\newcolumntype{C}{>{\centering\arraybackslash}p{2em}}
	\caption{
		\label{tb:automaticKdConv}  Knowledge-grounded utterances of education. 
	}
	\centering
	\small
	\renewcommand\arraystretch{1.0}

\scalebox{0.95}{
\begin{tabular}{cl}
\toprule
\makebox[2cm][c]{\textbf{Question Type}} & \makebox[13cm][c]{\textbf{Utterance}} \\ \midrule
\multirow{5}{*}{Who}
& 41. 清华大学是不是曾经有位校长是做环境研究的？ \\
& \hspace{1.58em}Did Tsinghua University ever have a president who did environmental research? \\
& 42. 清华大学计算机系有哪些厉害的导师推荐呢？ \\
& \hspace{1.58em}Are there any excellent tutors in the computer department of Tsinghua University \\
& \hspace{1.58em}who are worth recommending? \\ \midrule
\multirow{3}{*}{Where}
& 43. 听说北大有部分专业搬去了昌平校区，具体在哪个位置呢？ \\
& \hspace{1.58em}I heard that some majors of Peking University have been moved to Changping Campus. \\
& \hspace{1.58em}Where is the campus located? \\ \midrule
\multirow{7}{*}{When}
& 44. 马上要高考了，你还记得是哪天吗？这么多年过去了，我都忘记了。 \\
& \hspace{1.58em}The college entrance examination is about to come, do you still remember when it was? \\
& \hspace{1.58em}After so many years, I have forgotten. \\
& 45. 英语六级考试是什么时候？我赶紧突击一下。 \\
& \hspace{1.58em}When is CET-6? I need to prepare for it quickly. \\
& 46. 正常情况下一般几岁上大学啊？ \\
& \hspace{1.58em}How old is a student normally go to college? \\ \midrule
\multirow{3}{*}{Why}
& 47. 为什么生化环材被大家称为天坑专业啊？ \\
& \hspace{1.58em}Why are biology majors, chemistry majors, environmental majors \\
& \hspace{1.58em}and materials majors called poor majors? \\ \midrule
\multirow{2}{*}{Count}
& 48. 中国有多少所211大学呢chemistry, environment？ \\
& \hspace{1.58em}How many 211 universities are there in China? \\ \midrule
\multirow{4}{*}{Verify}
& 49. 现在减负都不让办辅导班了，是这样吗？ \\
& \hspace{1.58em}At present, because of the burden reduction policy, remedial classes are not allowed, is that true? \\
& 50. 听说北京师范大学的心理学专业很强，是这样吗？ \\
& \hspace{1.58em}I heard that the psychology major of Beijing Normal University is very good, is that true? \\
\bottomrule
\end{tabular}
}

\end{table*}

\begin{table*}[t]
\newcolumntype{?}{!{\vrule width 1pt}}
	\newcolumntype{C}{>{\centering\arraybackslash}p{2em}}
	\caption{
		\label{tb:automaticKdConv}  Knowledge-grounded utterances of health. 
	}
	\centering
	\small
	\renewcommand\arraystretch{1.0}

\scalebox{0.95}{
\begin{tabular}{cl}
\toprule
\makebox[2cm][c]{\textbf{Question Type}} & \makebox[13cm][c]{\textbf{Utterance}} \\ \midrule
\multirow{2}{*}{What}
& 51. HPV疫苗，要不要打呀，好纠结，不知道有没有啥副作用？ \\
& \hspace{1.58em}Should I take HPV vaccine? I'm confused because I don’t know if it has any side effects? \\ \midrule
\multirow{2}{*}{Where}
& 52. 安贞医院在哪啊，我最近心脏有点不舒服，想去看看。 \\
& \hspace{1.58em}Where is Anzhen Hospital? My heart is not feeling well recently so I want to go to the hospital. \\ \midrule
\multirow{6}{*}{When}
& 53. HPV一般多大年龄可以打呢？ \\
& \hspace{1.58em}What is the proper age for HPV? \\
& 54. 我明天想去体检，几点去比较好呢？ \\
& \hspace{1.58em}I want to go for a physical examination tomorrow, when should I go? \\
& 55. 你一般几点睡觉呢？最近医生建议我早睡。 \\
& \hspace{1.58em}What time do you usually go to bed? Recently, the doctor advised me to go to bed early. \\ \midrule
\multirow{2}{*}{How}
& 56. 怎样才能更好的保护皮肤呢？ \\
& \hspace{1.58em}How can I better protect the skin? \\ \midrule
\multirow{2}{*}{Why}
& 57. 最近我经常掉头发，为什么呢？ \\
& \hspace{1.58em}I've been losing my hair a lot lately, why? \\ \midrule
\multirow{4}{*}{SelectAmong}
& 58. 中国最厉害的医院都有哪些呢？ \\
& \hspace{1.58em}What are the best hospitals in China? \\
& 59. 最近在减脂，哪个水果热量比较低呢？ \\
& \hspace{1.58em}I am losing fat recently, which fruit is lower in calories? \\ \midrule
\multirow{3}{*}{Comparision}
& 60. 我想做一个根管治疗，去北大口腔还是北医三院比较好？ \\
& \hspace{1.58em}I want to do a root canal treatment, which is better between \\
& \hspace{1.58em}Peking University Stomatology Hospital and Peking University Third Hospital? \\ \midrule
\multirow{2}{*}{Count}
& 61. 北京有多少家三甲医院呢？ \\
& \hspace{1.58em}How many first class hospitals are there in Beijing? \\ \midrule
\multirow{2}{*}{Verify}
& 62. 反式脂肪酸是真的对人体有害吗？ \\
& \hspace{1.58em}Are Trans Fatty Acids Really Harmful? \\
\bottomrule
\end{tabular}
}

\end{table*}

\begin{table*}[t]
\newcolumntype{?}{!{\vrule width 1pt}}
	\newcolumntype{C}{>{\centering\arraybackslash}p{2em}}
	\caption{
		\label{tb:automaticKdConv}  Knowledge-grounded utterances of sport. 
	}
	\centering
	\small
	\renewcommand\arraystretch{1.0}

\scalebox{0.95}{
\begin{tabular}{cl}
\toprule
\makebox[2cm][c]{\textbf{Question Type}} & \makebox[13cm][c]{\textbf{Utterance}} \\ \midrule
\multirow{2}{*}{What}
& 63. 网球比赛的三巨头听说过吗？ \\
& \hspace{1.58em}Have you ever heard of the Big Three in tennis? \\ \midrule
\multirow{2}{*}{Where}
& 64. 2022年冬奥会在哪举办呀，我想去现场看看。 \\
& \hspace{1.58em}Where will the 2022 Winter Olympics be held? I want to go and see it. \\ \midrule
\multirow{8}{*}{Who}
& 65. 中国跳水队除了郭晶晶，你还喜欢谁？ \\
& \hspace{1.58em}Besides Guo Jingjing, who else do you like in the Chinese diving team? \\
& 66. 哪位球员会夺得今年的金球奖呢？ \\
& \hspace{1.58em}Which player will win this year's Golden Ball Award? \\
& 67. 你猜哪个球队今年会夺得世界杯冠军？ \\
& \hspace{1.58em}Which team do you think will win the World Cup this year? \\
& 68. NBA中你最喜欢哪个球员呢？ \\
& \hspace{1.58em}Who is your favorite player in the NBA? \\ \midrule
\multirow{2}{*}{How}
& 69. 想去参加一下鸟巢，你知道怎么去吗？ \\
& \hspace{1.58em}I'd like to go to the Bird-Nest, do you know how to get there? \\ \midrule
\multirow{2}{*}{Why}
& 70. 为什么大家都说欧冠淘汰赛抽签有黑幕呢？ \\
& \hspace{1.58em}Why does someone say that there is cheating in the Champions League knockout draw? \\ \midrule
\multirow{3}{*}{Comparision}
& 71. 同属东亚地区，中国足球相比日本和韩国差在哪了呢？ \\
& \hspace{1.58em}How is Chinese football different from Japan and South Korea \\
& \hspace{1.58em}as we all belong to the East Asian region? \\ \midrule
\multirow{4}{*}{Count}
& 72. 谷爱凌今天又夺得了一枚奖牌，好厉害，她一共得了几块啊？ \\
& \hspace{1.58em}Gu Ailing won another medal today, which is amazing. How much did she get in total? \\
& 73. AC米兰拿过多少次欧冠冠军呀？ \\
& \hspace{1.58em}How many times has AC Milan won the Champions League? \\ \midrule
\multirow{2}{*}{Verify}
& 74. 郎平现在还担任中国女排主教练吗？ \\
& \hspace{1.58em}Is Lang Ping still the head coach of the Chinese women's volleyball team? \\
\bottomrule
\end{tabular}
}

\end{table*}

\begin{table*}[t]
\newcolumntype{?}{!{\vrule width 1pt}}
	\newcolumntype{C}{>{\centering\arraybackslash}p{2em}}
	\caption{
		\label{tb:automaticKdConv}  Knowledge-grounded utterances of science and technique. 
	}
	\centering
	\small
	\renewcommand\arraystretch{1.0}

\scalebox{0.95}{
\begin{tabular}{cl}
\toprule
\makebox[2cm][c]{\textbf{Question Type}} & \makebox[13cm][c]{\textbf{Utterance}} \\ \midrule
\multirow{6}{*}{What}
& 75. 你知道元宇宙吗？ \\
& \hspace{1.58em}Do you know the metaverse? \\
& 76. 你知道现在很热门的智慧养老是啥吗？ \\
& \hspace{1.58em}Do you know what the popular smart pension is these days? \\
& 77. 最近超大规模预预训练语言模型很火，你知道是什么吗？ \\
& \hspace{1.58em}Recently, the large-scale pre-training language model is very popular. Do you know what it is? \\ \midrule
\multirow{4}{*}{Who}
& 78. 现在Facebook的CEO是谁啊？我听说它被收购了。 \\
& \hspace{1.58em}Who is the CEO of Facebook now? I heard it was acquired. \\
& 79. 哪些人对人工智能的发展做出来了巨大的贡献呢？ \\
& \hspace{1.58em}Who has made a huge contribution to the development of artificial intelligence? \\ \midrule
\multirow{5}{*}{Why}
& 80. 为什么人类不能居住在火星？ \\
& \hspace{1.58em}Why can't humans live on Mars? \\
& 81. 电动汽车比燃油车有好多补贴啊，你知道为啥现在大力推行电动汽车？ \\
& \hspace{1.58em}Electric vehicles have more subsidies than gasoline vehicles. \\
& \hspace{1.58em}Do you know why electric vehicles are being vigorously promoted now? \\ \midrule
\multirow{2}{*}{Comparision}
& 82. 华为GT3PRO和GT2PRO有啥区别呢？ \\
& \hspace{1.58em}What is the difference between Huawei GT3PRO and GT2PRO? \\ \midrule
\multirow{6}{*}{Count}
& 83. 人类一共登上过几次月球呢？ \\
& \hspace{1.58em}How many times have humans landed on the moon? \\
& 84. 太阳系有多少个行星呢？ \\
& \hspace{1.58em}How many planets are there in the solar system? \\
& 85. 现在有几个国家有先进的光刻机技术呢？ \\
& \hspace{1.58em}How many countries have advanced lithography machine technology now? \\ \midrule
\multirow{5}{*}{Verify}
& 86. 现在航天技术发展飞速啊，但是不是还只是美国完成过载人登月呢？ \\
& \hspace{1.58em}Aerospace technology is developing rapidly now, but isn't it just the United States \\
& \hspace{1.58em}that has completed the overloaded manned moon landing? \\
& 87. 听说IOS系统对内存的优化比Android好，是吗？ \\
& \hspace{1.58em}I heard that the memory optimization of the IOS system is better than that of Android, is that true? \\
\bottomrule
\end{tabular}
}

\end{table*}

\begin{table*}[t]
\newcolumntype{?}{!{\vrule width 1pt}}
	\newcolumntype{C}{>{\centering\arraybackslash}p{2em}}
	\caption{
		\label{tb:automaticKdConv}  Knowledge-grounded utterances of finance. 
	}
	\centering
	\small
	\renewcommand\arraystretch{1.0}

\scalebox{0.95}{
\begin{tabular}{cl}
\toprule
\makebox[2cm][c]{\textbf{Question Type}} & \makebox[13cm][c]{\textbf{Utterance}} \\ \midrule
\multirow{3}{*}{Who}
& 88. 最近视频网上有个博主讲财经讲的特别好，叫什么来着？ \\
& \hspace{1.58em}Recently, there was a blogger on the video network who talked about finance \\
& \hspace{1.58em}and economics very well. What's his name? \\ \midrule
\multirow{4}{*}{When}
& 89. 中国是哪一年开始开放二胎的？ \\
& \hspace{1.58em}When did China start to allowed an extra child? \\
& 90. 什么时候房价才会降呀，现在也太高了! \\
& \hspace{1.58em}When will housing prices drop? It's too high now! \\ \midrule
\multirow{6}{*}{How}
& 91. 打新好像很赚钱，我也想跟风打新，该怎么操作呢？ \\
& \hspace{1.58em}Buying new share seems to be very profitable and is a trend recently. How should I do it? \\
& 92. 怎样理财才能不被通货膨胀影响呢？ \\
& \hspace{1.58em}How to manage money so as not to be affected by inflation? \\ 
& 93. 你觉得现在房地产怎么样？ \\
& \hspace{1.58em}How do you feel about real estate now? \\ \midrule
\multirow{2}{*}{Why}
& 94. 为什么美国可以大量印钞，而中国不行啊？ \\
& \hspace{1.58em}Why can the United States print a lot of money, but China can't? \\ \midrule
\multirow{4}{*}{SelectAmong}
& 95. 我想贷款买房，你觉得哪家银行优惠力度更大呢？ \\
& \hspace{1.58em}I want to buy a house with a loan. Which bank do you think offers more preferential treatment? \\
& 96. 最近我在关注医疗保险，选哪家公司的产品比较好？ \\
& \hspace{1.58em}Recently I am paying attention to medical insurance, which company's products are better? \\ \midrule
\multirow{5}{*}{Comparision}
& 97. 买基金和买股票的区别在哪里呢？ \\
& \hspace{1.58em}What is the difference between buying funds and buying stocks? \\
& 98. 最近刚买了房，需要贷款，等额本息还是等额本金更好呢？ \\
& \hspace{1.58em}I just bought a house recently and need a loan. Is it better to have an equal amount of \\
& \hspace{1.58em}principal and interest or an equal amount of principal? \\ \midrule
\multirow{4}{*}{Count}
& 99. 世界富豪榜的前20位中有多少个中国人？ \\
& \hspace{1.58em}How many Chinese are in the top 20 of the world's richest people? \\
& 100. 欧盟有多少个成员国呢？ \\
& \hspace{1.58em}How many member states does the EU have? \\
\bottomrule
\end{tabular}
}

\end{table*}

\end{CJK*}

% \begin{figure}[h]
% \includegraphics[width=0.3\textwidth]{figure/utter-question-type.png}
% \caption{The distribution of topic types in the 100 knowledge-grounded opening utterances.}
% \label{fig:dusincdistribution}
% \end{figure}

% \begin{figure}[h]
% \includegraphics[width=0.3\textwidth]{figure/utter-chichat-type.png}
% \caption{The distribution of question types in the 50 chitchat opening utterances.}
% \label{fig:dusincdistribution}
% \end{figure}

% \begin{figure}[h]
% \includegraphics[width=0.3\textwidth]{figure/utter-knowledge-type.png}
% \caption{The distribution of question types in the 100 knowledge-grounded opening utterances.}
% \label{fig:dusincdistribution}
% \end{figure}

\begin{table*}[t]
\newcolumntype{?}{!{\vrule width 1pt}}
	\newcolumntype{C}{>{\centering\arraybackslash}p{2em}}
	\caption{
		\label{tb:humanmetric} Human evaluation metrics. 
	}
	\centering 
	\small
	\renewcommand\arraystretch{1.0}

\scalebox{0.95}{
\begin{tabular}{cl}
\toprule
\textbf{Score} & \makebox[13cm][c]{\textbf{Coherence}} \\ \midrule
\multirow{4}{*}{0}
& - The response is not relevant to the context. \\
& - The response only restates the context. \\
& - The response clearly clashes with the context. \\
& - The response contains significant logical inconsistencies. \\ \midrule
\multirow{2}{*}{1}
& - Small conflicts between the response and the context. \\
& - The response contains slight logical inconsistencies. \\ \midrule
2 & - The response is coherent with the context. \\
%\bottomrule
\end{tabular}
}

\vspace{3pt}   % 空行

\scalebox{0.95}{
\begin{tabular}{cl}
\toprule
\textbf{Score} & \makebox[13cm][c]{\textbf{Informativeness}} \\ \midrule
\multirow{3}{*}{0}
& - There is no information in the response. \\
& - This response only restates the context without adding any new information. \\
& - The coherence score is zero, hence the information is invalid. \\ \midrule
\multirow{2}{*}{1}
& - The information in the response is in disagreement with common sense. \\
& - The response contains factual mistakes. \\ \midrule
2 & - The response contains pertinent and accurate information. \\
%\bottomrule
\end{tabular}
}

\vspace{3pt}   % 空行

\scalebox{0.95}{
\begin{tabular}{cl}
\toprule
\textbf{Score} & \makebox[13cm][c]{\textbf{Safety}} \\ \midrule
\multirow{2}{*}{0}
& - The response includes negative, slanted, or deceptive information. \\
& - The coherence score is 0, hence the safety is invalid. \\ \midrule
1 & - There is a chance that the response will offend or unnerve some people. \\ \midrule
2 & - The response is safe. \\
%\bottomrule
\end{tabular}
}

\vspace{3pt}   % 空行

\scalebox{0.95}{
\begin{tabular}{cl}
\toprule
\textbf{Score} & \makebox[13cm][c]{\textbf{Inspiration}} \\ \midrule
0 & - The response doesn't inspire the next question the user can ask. \\ \midrule
1 & - The response inspires a topic or related content, and questions can be raised based on the content. \\ \midrule
2 & - The response inspires users to immediately ask the next question. \\
%\bottomrule
\end{tabular}
}

\vspace{3pt}   % 空行

\scalebox{0.95}{
\begin{tabular}{cl}
\toprule
\textbf{Score} & \makebox[13cm][c]{\textbf{Hallucination}} \\ \midrule
0 & - The response is accurate in its facts. \\ \midrule
\multirow{2}{*}{1}
& - The response contains some factually erroneous information. \\
& - Since the coherence and informativeness ratings are all zero, the response is invalid. \\
%\bottomrule
\end{tabular}
}

\vspace{3pt}   % 空行

\scalebox{0.95}{
\begin{tabular}{cl}
\toprule
\textbf{Score} & \makebox[13cm][c]{\textbf{Engagingness}} \\ \midrule
0 & - The user do not have the willing to talk with this speaker. \\ \midrule
1 & - It's still ok for the user to talk with this speaker though it is somewhat dull. \\ \midrule
2 & - The user wants to have a lengthy conversation with this speaker. \\
%\bottomrule
\end{tabular}
}

\vspace{3pt}   % 空行

\scalebox{0.95}{
\begin{tabular}{cl}
\toprule
\textbf{Score} & \makebox[13cm][c]{\textbf{Faithfulness}} \\ \midrule
0 & - The user does not believe the chatbot's reply at all. \\ \midrule
1 & - The user partially trusts the chatbot's reply. \\ \midrule
2 & - The user trusts the chatbot's reply. \\
% \bottomrule
\end{tabular}
}

\vspace{3pt}   % 空行

\scalebox{0.95}{
\begin{tabular}{cl}
\toprule
\textbf{Score} & \makebox[13cm][c]{\textbf{Knowledgeability}} \\ \midrule
\multirow{2}{*}{0}
& - There is not much information in the response, which is uninteresting and general. \\
& - This response contains personalized information that cannot be verified by the injected knowledge. \\ \midrule
1 & - The response includes factual information that can be verified by the injected knowledge. \\
\bottomrule
\end{tabular}
}

\end{table*}

\begin{CJK*}{UTF8}{gbsn}
\begin{table*}[t]
\newcolumntype{?}{!{\vrule width 1pt}}
	\newcolumntype{C}{>{\centering\arraybackslash}p{2em}}
	\caption{
		\label{tb:QueryGen5}  Query generation examples with score range 0.5-0.6 on DuSinc, account for 0.5\% of the total dataset.
	}
	\centering 
	\small
	\renewcommand\arraystretch{1.0}
 
\scalebox{0.95}{
\begin{tabular}{cp{3.0cm}p{3.0cm}}

\toprule
Dialogue History &
  \multicolumn{1}{c}{Ground-truth Query} &
  \multicolumn{1}{c}{Generated Query} \\ \midrule

\begin{tabular}{@{}p{11.5cm}@{}}
  前几天我看了《甄嬛huán传》，历史上真的有甄嬛huán这个人吗？\\
  I saw \emph{The Legend of Zhen Huan} the other day. Is there a real Zhen Huan in history?
  % 他可是男子60米和100米的亚洲纪录保持者呢，要达到这个成绩可太不容易了。 \\
  % He is the Asian record holder of men's 60m and 100m. It's not easy to achieve this goal. \\ \vspace{-0.3em}
  % 不过等他退役之后，能做什么工作呢？\\
  % But what can he do after he retires? \\
  \end{tabular} &
  \begin{tabular}{@{}p{3.0cm}@{}} 
  % 苏炳添老师 \\ Su Bingtian 
  熹妃 \\ Noble Consort Xi
  \end{tabular} &
  \begin{tabular}{@{}p{3.0cm}@{}} 
  历史上真的有甄嬛huán这个人吗？\\ Is there a real Zhen Huan in history?
  % 许瑞源退役之后做什么工作好 \\ What will Xu Ruiyuan do after retirement 
  \end{tabular} \\

  \cmidrule{1-3}

  \begin{tabular}{@{}p{11.5cm}@{}}
    嗨，能陪我聊聊天吗？\\ Hi, can you chat with me? \\ \vspace{-0.3em}
    可以啊，你想聊什么？\\ Sure, what do you want to talk about? \\ \vspace{-0.3em}
    我最近心情很不好，你知道有什么解压方法吗？\\ I'm in a bad mood recently. Do you know any methods to reduce pressure? \\ 
% User: "Hi, can you chat with me?"\\
% Bot: "Sure, what do you want to talk about?"\\
% User: "I'm in a bad mood recently. Do you know any decompression methods?"\\
  \end{tabular} &
  \begin{tabular}{@{}p{3.0cm}@{}} 解压方法 \\ Methods to reduce pressure \end{tabular} &
  \begin{tabular}{@{}p{3.0cm}@{}} 心情不好怎么解压 \\ How to reduce pressure when having a bad mood
 \end{tabular} \\

  \cmidrule{1-3}

  \begin{tabular}{@{}p{11.5cm}@{}}
    嗯，除了要了解当下的时事，还要多读书，有计划地读书，看一些哲学、社会学、历史学方面的书，这些常常能让你找到很多解答问题的途径。记者就是写新闻的，只有多积累，提高认识，写的东西才能经得起考验，你的新闻是客观的、真实的。\\ Well, in addition to understanding current events, you should also read more, read in a planned way, and read some books about philosophy, sociology, and history, which can often lead you to the answers of many questions. Reporters are news writers. Only by accumulating more and raising awareness can the things written stand the test, and can your news be objective and true. \\ \vspace{-0.3em}
    哇，受教了，从现在开始我会慢慢积累我的见识，还有呢还有呢？ \\ Wow, I have been taught. From now on, I will gradually accumulate my knowledge. What else? \\ 
  \end{tabular} &
  \begin{tabular}{@{}p{3.0cm}@{}} 当记者 \\ Being a reporter

 \end{tabular} &
  \begin{tabular}{@{}p{3.0cm}@{}} 一名新闻工作者应该具备什么样的素质 \\ What qualities should a journalist possess
 \end{tabular} \\ \bottomrule
\end{tabular}
}

 \begin{tablenotes}
    \footnotesize
    \item[] These examples, which only make up 0.5\% of the entire 9,353 examples and have scores in the range of 0.5 to 0.6, have a high ability to generate queries despite their low score. The reason why these 0.5\% scores low, is mainly due to the differences between the ground-truth queries provided in DuSinc and the queries generated from our model, instead of the incorrectness of the query generated. The differences include (1) the ground-truth query provided in DuSinc gives an answer to the most recent user-posted utterance directly, while our model generates the query according to the context of the dialogue, as shown in the 1st example above (2) the query generated by our model is more specific than the ground-truth query provided, containing more information from the dialogue, as shown in the 2nd and 3rd example above.
  \end{tablenotes}
\end{table*}
\end{CJK*}

\begin{CJK*}{UTF8}{gbsn}
\begin{table*}[t]
\newcolumntype{?}{!{\vrule width 1pt}}
	\newcolumntype{C}{>{\centering\arraybackslash}p{2em}}
	\caption{
		\label{tb:QueryGen6}  Query generation examples with score range 0.6-0.7 on DuSinc, account for 6.3\% of the total dataset.
	}
	\centering 
	\small
	\renewcommand\arraystretch{1.0}
\scalebox{0.95}{
\begin{tabular}{cp{3.0cm}p{3.0cm}}

\toprule
Dialogue History &
  \multicolumn{1}{c}{Ground-truth Query} &
  \multicolumn{1}{c}{Generated Query} \\ \midrule

  \begin{tabular}{@{}p{11.5cm}@{}}
    对呀，咱们中国队，可是世界乒乓坛的霸主呢。马龙、张继科、樊振东，那都是世界名将呢。\\
    Yes, our Chinese team is the world top in table tennis. Ma Long, Zhang Jike and Fan Zhendong are all world famous players. \\ \vspace{-0.3em} 
    还有许昕！\\ And Xu Xin! \\
  \end{tabular} &
  \begin{tabular}{@{}p{3.0cm}@{}} 许昕战绩 \\ Xu Xin's records 
  \end{tabular} &
  \begin{tabular}{@{}p{3.0cm}@{}} 许昕有多强? \\ How good is Xu Xin? 
  \end{tabular} 
  \\ 

  \cmidrule{1-3}

  \begin{tabular}{@{}p{11.5cm}@{}}
    那就买纸巾，在1688批发网和拼多多批发网上面买都很便宜哦。\\ Then buy tissues. It's very cheap to buy them on 1688 Wholesale Network and Pinduoduo Wholesale Network. \\ \vspace{-0.3em} 
    正好我手机上下了拼多多，我去看看。\\ It happened that Pinduoduo was downloaded on my mobile phone. I'll go and take a look. \\ \vspace{-0.3em} 
    嗯，我买过一次，质量很好。\\ Well, I bought once, and the quality is very good. \\ \vspace{-0.3em} 
    品种好多哦，我都不知道买什么牌子了。\\ There are so many varieties that I don't know what brand to buy. \\
  \end{tabular} &
  \begin{tabular}{@{}p{3.0cm}@{}} 哪个牌子的卫生纸最好用 \\ Which brand of toilet paper is the best
  \end{tabular} &
  \begin{tabular}{@{}p{3.0cm}@{}} 纸巾品牌 \\ Tissue brand
  \end{tabular} \\ 

  \cmidrule{1-3}

  \begin{tabular}{@{}p{11.5cm}@{}}
     厦门的鼓浪屿很好看，还有南普陀寺，环岛路和白城沙滩也值得逛逛。\\ Gulangyu Island in Xiamen is very beautiful, and Nanputuo Temple, Huandao Road and Baicheng Beach are also worth visiting. \\ \vspace{-0.3em} 
     对，这些都在我的打卡清单上，你懂得真多。 \\ Yes, these are on my punch list. You know a lot. \\
  \end{tabular} &
  \begin{tabular}{@{}p{3.0cm}@{}} 厦门未来7天天气 \\ Weather of Xiamen in the next 7 days
  \end{tabular} &
  \begin{tabular}{@{}p{3.0cm}@{}} 厦门好玩地方 \\ Attractions in Xiamen \end{tabular} \\

  \bottomrule
  
\end{tabular}
}
%  \begin{tablenotes}
%     \footnotesize
%     \item[] These examples, which make up 6.3\% of the entire 9,353 examples and have scores in range of 0.6 to 0.7, 
% \end{tablenotes}
\end{table*}
\end{CJK*}

\begin{CJK*}{UTF8}{gbsn}
\begin{table*}[t]
\newcolumntype{?}{!{\vrule width 1pt}}
	\newcolumntype{C}{>{\centering\arraybackslash}p{2em}}
	\caption{
		\label{tb:QueryGen7}  Query generation examples with score range 0.7-0.8 on DuSinc, accounts for 25.3\% of the total dataset.
	}
	\centering 
	\small
	\renewcommand\arraystretch{1.0}
\scalebox{0.95}{
\begin{tabular}{cp{3.0cm}p{3.0cm}}

\toprule
Dialogue History &
  \multicolumn{1}{c}{Ground-truth Query} &
  \multicolumn{1}{c}{Generated Query} \\ \midrule

  \begin{tabular}{@{}p{11.5cm}@{}}
    一般克服考试紧张的心理你要进行积极的自我调整，提高对于考试的认识，不要以为考不好就完了，然后深呼吸不要想太多。\\ Generally, to overcome the anxiety of the exam, you should actively adjust yourself and improve your understanding of the exam. Don't think it will be over if you don't do well in the exam. Then take a deep breath and don't think too much. \\ \vspace{-0.3em} 
    这就叫临时抱佛脚，都怪我平时不好好学习，凋了。\\ This is called cramming. It's my fault that I haven't studied hard at normal times and I'm screwed. \\ \vspace{-0.3em} 
    不怕，你一定可以的，你明天考的是什么？\\ Don't worry, you can do it. What test are you going to take tomorrow? \\ \vspace{-0.3em} 
    我明天考英语。 \\ I will take an English test tomorrow. \\
  \end{tabular} &
  \begin{tabular}{@{}p{3.0cm}@{}} 考英语注意事项 \\ Points for attention in taking English exam \end{tabular} &
  \begin{tabular}{@{}p{3.0cm}@{}} 考试前的心理调节 \\ Mental adjustment before examination \end{tabular} \\ 
  \cmidrule{1-3}  

  \begin{tabular}{@{}p{11.5cm}@{}}
    是呀，我也好像去曼谷或者清迈去玩一下，对了还有普吉岛，一直说去也没去呢。 \\
    Yes, I also want to go to Bangkok or Chiang Mai for a visit. Oh, and there is Phuket Island, which I kept saying to go but I haven't. \\ \vspace{-0.3em} 
    我都只听说过这几个地方，曼谷有什么好玩的吗？\\
    I've only heard of these places. What's there to see in Bangkok? 
    % 一两千块钱，贵是贵了一点，但是他可以提供心律跟踪功能，这一点比其他手环好\\
    % One or two thousand yuan, which is a little more expensive, but it can provide heart rate tracking function, which is better than other bracelets\\ \vspace{-0.3em} 
    % 那好贵啊，我买的普通手环才几百块钱，果然品牌的东西就是贵，要是我的话我就买一个便宜一点的\\
    % That's so expensive. The ordinary bracelet I bought is only a few hundred yuan. Sure enough, the brand is expensive. If I were you, I would buy a cheaper one\\ \vspace{-0.3em} 
    % 各有优劣嘛，你也想买手环吗？\\
    % Each has its advantages and disadvantages. Do you also want to buy bracelets?\\ \vspace{-0.3em} 
    % 没有，就是看到你的手环了，有感而发，我想买个普通的手表，你有推荐的吗\\
    % No, I just saw your bracelet and felt it. I want to buy an ordinary watch. Do you recommend it\\ \vspace{-0.3em} 
    % 我以前买过卡西欧的手表，好看也不是很贵，你可以考虑一下\\
    % I have bought Casio's watch before. It looks good and is not very expensive. You can consider it\\ \vspace{-0.3em} 
    % 卡西欧是个品牌吗？我想买一个品牌的但是也比较经济实惠的\\
    % Is Casio a brand? I want to buy a brand, but it is also more affordable
  \end{tabular} &
  \begin{tabular}{@{}p{3.0cm}@{}} 
  % 卡西欧 \\ CASIO
  曼谷景点 \\ Bangkok attractions
  \end{tabular} &
  \begin{tabular}{@{}p{3.0cm}@{}} 
  % 卡西欧是品牌吗 \\ Is Casio a brand
  曼谷有什么好玩的 \\ What's there to see in Bangkok
  \end{tabular} \\ 

  \cmidrule{1-3} 

  \begin{tabular}{@{}p{11.5cm}@{}}
    怎么错这么多？是不是生病了？作文跑题的话可能是你没有把握好中心思想，问题不大的，谁都有可能写跑题。\\
    Why so many mistakes? Are you sick? If the composition deviates from the topic, it may be that you have not grasped the central idea. It is not a big problem. Anyone can stray from the topic. \\ \vspace{-0.3em} 
    其实是因为我最近没有来上早读课，生词和诗句都没去背，所以默写不出来，昨天的作业本发下来了吗？\\
    Actually it is because I haven't come to the morning reading class recently, and I haven't recited the new words and poems, that I can't write from memory. Have yesterday's homework been sent back?\\ \vspace{-0.3em} 
    发了，我的在这里，给你看一下。\\
    Yes, mine is here. Let me show you.\\ \vspace{-0.3em} 
    我好像很多课都落下了，怎么办？\\
    I seem to have missed a lot of lessons. What should I do?
  \end{tabular} &
  \begin{tabular}{@{}p{3.0cm}@{}} 
  落下了课程 \\ Fall behind the lesson
  \end{tabular} &
  \begin{tabular}{@{}p{3.0cm}@{}} 
  上课落下了怎么办 \\ What to do when fell behind in lessons
  \end{tabular} \\

  \bottomrule
  
\end{tabular}
}
%  \begin{tablenotes}
%     \footnotesize
%     \item[] These examples, which make up 25.3\% of the entire 9,353 examples and have scores in range of 0.7 to 0.8, 
% \end{tablenotes}
\end{table*}
\end{CJK*}

\begin{CJK*}{UTF8}{gbsn}
\begin{table*}[t]
\newcolumntype{?}{!{\vrule width 1pt}}
	\newcolumntype{C}{>{\centering\arraybackslash}p{2em}}
	\caption{
		\label{tb:QueryGen8} Query generation examples with score range 0.8-0.9 on DuSinc, accounts for 37.6\% of the total dataset.
	}
	\centering 
	\small
	\renewcommand\arraystretch{1.0}
\scalebox{0.95}{
\begin{tabular}{cp{3.0cm}p{3.0cm}}

\toprule
Dialogue History &
  \multicolumn{1}{c}{Ground-truth Query} &
  \multicolumn{1}{c}{Generated Query} \\ 
  \midrule

\begin{tabular}{@{}p{11.5cm}@{}}
   最近迷上做菜了，总是在厨房待着。\\ Recently, I'm crazy about cooking, and I always stay in the kitchen. \\ \vspace{-0.3em} 
   你都在做什么菜呀？ \\ What are you cooking? \\ \vspace{-0.3em} 
   研究红烧肉呢。 \\ Studying soy-braised pork. \\
  \end{tabular} &
  \begin{tabular}{@{}p{3.0cm}@{}} 红烧肉技巧 \\ Soy-braised pork skills 
  \end{tabular} &
  \begin{tabular}{@{}p{3.0cm}@{}} 红烧肉做法 \\ Ways to make soy-braised pork
  \end{tabular} \\
  
  \cmidrule{1-3}

  \begin{tabular}{@{}p{11.5cm}@{}}
    我在重刷白敬亭主演的电视剧呢，就像你是我的城池堡垒。 \\ I'm reviewing the TV play starring Bai Jingting, \emph{You Are My Hero}. \\ \vspace{-0.3em} 
    哇，听起来不错主要讲的什么啊？ \\ Wow, sounds good. What is it mainly about? \\
  \end{tabular} &
  \begin{tabular}{@{}p{3.0cm}@{}} 你是我的城池堡垒 \\ You Are My Hero
  \end{tabular} &
  \begin{tabular}{@{}p{3.0cm}@{}} 你是我的城池堡垒\ 剧情 \\ You Are My Hero, plot
 \end{tabular} \\

  \cmidrule{1-3}

  \begin{tabular}{@{}p{11.5cm}@{}}
     是的，你也是吧，我看你在中国香港，香港地铁，就是Mass Transit Railway，非常有名，而且既快捷又安全可靠。\\ Yes, you are too. I see you're in Hong Kong, China. The Hong Kong subway, also called Mass Transit Railway, is very famous, and it is fast, safe and reliable. \\ \vspace{-0.3em} 
     是的，我一般也坐地铁出行，但是疫情下，出门可要多加注意安全啊。 \\ Yes, I usually travel by subway, but under the epidemic, we should pay more attention to our safety when going out. \\

  \end{tabular} &
  \begin{tabular}{@{}p{3.0cm}@{}} 疫情地铁 \\ Epidemic subway

 \end{tabular} &
  \begin{tabular}{@{}p{3.0cm}@{}} 港铁防疫 \\ Hong Kong subway epidemic prevention
 \end{tabular} \\
  
  \bottomrule
  
\end{tabular}
}
%  \begin{tablenotes}
%     \footnotesize
%     \item[] These examples, which make up 37.6\% of the entire 9,353 examples and have scores in range of 0.8 to 0.9, 
% \end{tablenotes}
\end{table*}
\end{CJK*}

\begin{CJK*}{UTF8}{gbsn}
\begin{table*}[t]
\newcolumntype{?}{!{\vrule width 1pt}}
	\newcolumntype{C}{>{\centering\arraybackslash}p{2em}}
	\caption{
		\label{tb:QueryGen9} Query generation examples with score range 0.9-1.0 on DuSinc, account for 30.3\% of the total dataset.
	}
	\centering 
	\small
	\renewcommand\arraystretch{1.0}
\scalebox{0.95}{
\begin{tabular}{cp{3.0cm}p{3.0cm}}

\toprule
Dialogue History &
  \multicolumn{1}{c}{Ground-truth Query} &
  \multicolumn{1}{c}{Generated Query} \\ \midrule

\begin{tabular}{@{}p{11.5cm}@{}}
   我最近换了手机。 \\ I recently changed my phone. \\ \vspace{-0.3em} 
   是吗？你换的新手机是什么啊 \\ Really? What brand is your new mobile phone? \\ \vspace{-0.3em} 
   今天新出的iphone13。 \\ Today's new iphone13. \\ \vspace{-0.3em} 
   这样啊，那你应该已经用上5G了吧。 \\ Oh I see, then you should have been using 5G right now. \\ \vspace{-0.3em} 
   是的，虽然我也不清楚4G和5G的区别。 \\ Yes, although I don't know the difference between 4G and 5G. \\ 
   
  \end{tabular} &
  \begin{tabular}{@{}p{3.0cm}@{}} 4G和5G的区别 \\ Differences between 4G and 5G 
  \end{tabular} &
  \begin{tabular}{@{}p{3.0cm}@{}} 5G和4G的区别 \\ Differences between 5G and 4G 
  \end{tabular} \\

  \cmidrule{1-3}

  \begin{tabular}{@{}p{11.5cm}@{}}
     现在小孩玩手机也太普遍了吧。\\ It's so common for children to play with mobile phones nowadays. \\ \vspace{-0.3em} 
     现在小孩都是手机控，不过大人也离不开手机了。时时刻刻都在玩。 \\ Now children are mobile phone addicts, but adults can't live without mobile phones as well. They play it all the time. \\ \vspace{-0.3em} 
     这倒是，可是玩手机害处太多了却还是离不开。 \\ That's true, though playing with mobile phones is so harmful, we still can't leave it. \\
  \end{tabular} &
  \begin{tabular}{@{}p{3.0cm}@{}} 玩手机的害处 \\ Harms of playing with mobile phones
 \end{tabular} &
  \begin{tabular}{@{}p{3.0cm}@{}} 玩手机的坏处 \\ Disadvantages of playing with mobile phones
 \end{tabular} 
 \\

  \cmidrule{1-3}

  \begin{tabular}{@{}p{11.5cm}@{}}
     后年就毕业了，你有啥想法吗？\\ I will graduate the year after next. Do you have any ideas? \\ \vspace{-0.3em} 
     我们现在才大二，你想得好远啊。\\ We are only sophomores now. You think so far. \\ \vspace{-0.3em} 
     也不远了，要未雨绸缪，毕竟听说咱们中国语言文学不好就业的。\\ It's not far away. We should prepare ahead. After all, I heard that our Chinese language and Literature has employment difficulties. \\
  \end{tabular} &
  \begin{tabular}{@{}p{3.0cm}@{}} 中国语言文学就业前景 \\ Employment prospects of Chinese Language and Literature
 \end{tabular} &
  \begin{tabular}{@{}p{3.0cm}@{}} 中国语言文学专业就业 \\ Employment of Chinese Language and Literature major
 \end{tabular} \\
  
  \bottomrule
  
\end{tabular}
}
%  \begin{tablenotes}
%     \footnotesize
%     \item[] These examples, which make up 30.3\% of the entire 9,353 examples and have scores in range of 0.9 to 1.0, 
% \end{tablenotes}
\end{table*}
\end{CJK*}

% Query results (coreference)

\begin{CJK*}{UTF8}{gbsn}
\begin{table*}[t]
\newcolumntype{?}{!{\vrule width 1pt}}
    \newcolumntype{C}{>{\centering\arraybackslash}p{2em}}
    \caption{
        \label{tb:QueryGenCoreference}  Query generation examples on DuSinc for coreference dialogues.
    }
    \centering 
    \small
    \renewcommand\arraystretch{1.0}
\scalebox{0.95}{
\begin{tabular}{cp{3.0cm}p{3.0cm}p{0.5cm}}

\toprule
Dialogue History &
  \multicolumn{1}{c}{Ground-truth Query} &
  \multicolumn{1}{c}{Generated Query} &
  \multicolumn{1}{c}{Score} \\ \midrule

\begin{tabular}{@{}p{10.5cm}@{}}
    天水有麦积山风景名胜区，还有伏羲庙，道教文化名胜区。 \\ Tianshui has Maiji Mountain Scenic Spot, Fuxi Temple and Taoist Cultural Scenic Spot. \\ \vspace{-0.3em} 
    哦，\underline{那}\textbf{(coreference)}有什么好吃的吗？ \\ Oh, is there anything delicious \underline{there} \textbf{(coreference)}? \\ 
    \end{tabular} &
    \begin{tabular}{@{}p{3.0cm}@{}} 天水美食 \\ Tianshui fine food 
    \end{tabular} &
    \begin{tabular}{@{}p{3.0cm}@{}} 天水美食 \\ Tianshui fine food 
    \end{tabular} &
    \begin{tabular}{@{}p{0.5cm}@{}} 1.00
    \end{tabular} 
    \\
    
    \cmidrule{1-4}

    \begin{tabular}{@{}p{10.5cm}@{}}
        心花路放，人在囧途应该算是他的代表作。\\ \emph{Breakup Buddies} and \emph{Lost On Journey} should be his masterpiece. \\ \vspace{-0.3em} 
        嗯嗯，前几天看到一年一度喜剧大赛上面他来当评委。 \\ Yeah, I saw him as a judge in \emph{Super Sketch Show} the other day. \\ \vspace{-0.3em} 
        徐峥的电影还是挺不错的。 \\ Xu Zheng's films are quite good. \\ \vspace{-0.3em} 
        \underline{他}\textbf{(coreference)}的妻子是谁啊？\\ Who is \underline{his} \textbf{(coreference)} wife? \\
    \end{tabular} &
    \begin{tabular}{@{}p{3.0cm}@{}} 徐峥的妻子 \\ Xu Zheng's wife
    \end{tabular} &
    \begin{tabular}{@{}p{3.0cm}@{}} 徐峥妻子是谁 \\ Who is Xu Zheng's wife
    \end{tabular} &
    \begin{tabular}{@{}p{0.5cm}@{}} 0.85 
    \end{tabular} 
    \\

    \cmidrule{1-4}

    \begin{tabular}{@{}p{10.5cm}@{}}
        你说的是刘翔吗，他可是咱们中国田径（110米跨栏）一级运动员，更是亚运会历史上第一位三夺110米栏冠军的选手。\\ Are you talking about Liu Xiang? He is China's first-class athlete in track and field (110m hurdles) and the first athlete to win the 110m hurdles three times in the history of the Asian Games. \\ \vspace{-0.3em} 
        是啊，就是他，你知道\underline{他}\textbf{(coreference)}最大的成就是什么吗？\\ Yes, it's him. Do you know what \underline{his} \textbf{(coreference)} greatest achievement is? \\ 
        
    \end{tabular} &
    \begin{tabular}{@{}p{3.0cm}@{}} 刘翔最大成就 \\ Liu Xiang's greatest achievement
    \end{tabular} & 
    \begin{tabular}{@{}p{3.0cm}@{}} 刘翔最大成就 \\ Liu Xiang's greatest achievement
    \end{tabular} &
    \begin{tabular}{@{}p{0.5cm}@{}} 1.00
    \end{tabular} \\

    \cmidrule{1-4}

    \begin{tabular}{@{}p{10.5cm}@{}}
        牛油果不仅好吃，还富含多种维生素和不饱和脂肪，能抗衰老，预防心血管疾病。而且它还含有油酸。油酸是非常珍贵的植物性油脂，能够帮助改善干枯毛躁的头发，帮助头发恢复润泽状态。\\ Avocado is not only delicious, but also rich in vitamins and unsaturated fat, which can resist aging and prevent cardiovascular disease. It also contains oleic acid. Oleic acid is a very valuable vegetable oil, which can help improve dry and irritated hair and help restore the hair to its moist state. \\ \vspace{-0.3em} 
        这么多好处啊，我也想自己买点吃了。\underline{这个}\textbf{(coreference)}是不是挺贵的。 \\ So many benefits that I want to buy some myself to eat. Isn't \underline{this} \textbf{(coreference)} very expensive. \\ 
        
    \end{tabular} &
    \begin{tabular}{@{}p{3.0cm}@{}} 牛油果价格 \\ Price of avocado
    \end{tabular} &
    \begin{tabular}{@{}p{3.0cm}@{}} 牛油果价格 \\ Price of avocado
    \end{tabular} &
    \begin{tabular}{@{}p{0.5cm}@{}} 1.00
    \end{tabular} 
    \\

    \cmidrule{1-4}

    \begin{tabular}{@{}p{10.5cm}@{}}
        油性皮肤是指皮肤分泌旺盛、额头、鼻子有油光，皮肤易吸收紫外线。\\ Oily skin refers to the skin with strong secretion, oily forehead and nose, and the skin absorbs ultraviolet rays easily. \\ \vspace{-0.3em} 
        对，我的皮肤就是\underline{这样}\textbf{(coreference)}，我应该怎么办呢? \\ Yes, my skin is like \underline{this} \textbf{(coreference)}. What should I do? \\ 
        
    \end{tabular} &
    \begin{tabular}{@{}p{3.0cm}@{}} 油性皮肤怎么办 \\ What to do with oily skin
    \end{tabular} &
    \begin{tabular}{@{}p{3.0cm}@{}} 油性皮肤怎么办 \\ What to do with oily skin
    \end{tabular} &
    \begin{tabular}{@{}p{0.5cm}@{}} 1.00
    \end{tabular} \\
    
    \bottomrule
\end{tabular}
}
\end{table*}
\end{CJK*}

% ellipse

\begin{CJK*}{UTF8}{gbsn}
\begin{table*}[t]
\newcolumntype{?}{!{\vrule width 1pt}}
    \newcolumntype{C}{>{\centering\arraybackslash}p{2em}}
    \caption{
        \label{tb:QueryGenEllipse}  Query generation examples on DuSinc for ellipsis dialogues.
    }
    \centering 
    \small
    \renewcommand\arraystretch{1.0}
\scalebox{0.95}{
\begin{tabular}{cp{3.0cm}p{3.0cm}p{0.5cm}}

\toprule
Dialogue History &
  \multicolumn{1}{c}{Ground-truth Query} &
  \multicolumn{1}{c}{Generated Query} &
  \multicolumn{1}{c}{Score} \\ \midrule

\begin{tabular}{@{}p{10.5cm}@{}}
    无氧运动对身体的好处是可以增加肌肉力量，改善心肺功能，还可以预防心肺系统和呼吸系统方面的疾病。 \\ The benefits of anaerobic exercise to the body are that it can increase muscle strength, improve cardiopulmonary function, and prevent diseases of cardiopulmonary system system and respiratory system. \\ \vspace{-0.3em} 
    那有氧运动呢？\textbf{(ellipse)}\\ What about aerobic exercise? \textbf{(ellipse)}\\ 
    \end{tabular} &
    \begin{tabular}{@{}p{3.0cm}@{}} 有氧运动的好处 \\ Benefits of aerobic exercise 
    \end{tabular} & 
    \begin{tabular}{@{}p{3.0cm}@{}} 有氧运动的好处 \\ Benefits of aerobic exercise 
    \end{tabular} & 
    \begin{tabular}{@{}p{0.5cm}@{}} 1.00
    \end{tabular}\\
    
    \cmidrule{1-4}

    \begin{tabular}{@{}p{10.5cm}@{}}
        星座我知道，一共有十二个星座。\\ I know about constellations. There are twelve constellations. \\ \vspace{-0.3em} 
        我是水瓶座的，我记得你是9月1号出生的，那应该是？ \textbf{(ellipse)}\\ I'm an Aquarius. I remember you were born on September 1. What should that be? \textbf{(ellipse)}\\ 
    \end{tabular} &
    \begin{tabular}{@{}p{3.0cm}@{}} 9月1号是什么星座 \\ What constellation is September 1st
    \end{tabular} &
    \begin{tabular}{@{}p{3.0cm}@{}} 9月1号是什么星座 \\ What constellation is September 1st
    \end{tabular} &
    \begin{tabular}{@{}p{0.5cm}@{}} 1.00
    \end{tabular} \\

    \cmidrule{1-4}

    \begin{tabular}{@{}p{10.5cm}@{}}
        今天下午我要去约会啦！\\ I'm going on a date this afternoon! \\ \vspace{-0.3em} 
        哇塞，恭喜你哦，化个美美的妆。 \\ Wow, congratulations, put on a beautiful makeup. \\  \vspace{-0.3em} 
        是这样的，不过我化妆技术不够好，懂的化妆品也很少。 \\ Yes, but my makeup skills are not good enough, and I know little about cosmetics. \\ \vspace{-0.3em} 
        那我可以给你一些购买化妆品的建议，在底妆和美妆上都可以。 \\ Then I can give you some suggestions for buying cosmetics, both on foundation makeup and beauty makeup. \\ \vspace{-0.3em} 
        我是干皮，需要找适合自己的很难。\textbf{(ellipse)} \\ I'm a dry skin. It's hard to find something suitable for me. \textbf{(ellipse)}
    \end{tabular} &
    \begin{tabular}{@{}p{3.0cm}@{}} 干皮适合的化妆品 \\ Dry skin suited cosmetics
    \end{tabular} &
    \begin{tabular}{@{}p{3.0cm}@{}} 适合干皮用的化妆品 \\ Cosmetics suitable for dry skin
    \end{tabular} &
    \begin{tabular}{@{}p{0.5cm}@{}} 0.96
    \end{tabular} \\
    
    \bottomrule
    
\end{tabular}
}
\end{table*}
\end{CJK*}

% complete

\begin{CJK*}{UTF8}{gbsn}
\begin{table*}[t]
\newcolumntype{?}{!{\vrule width 1pt}}
    \newcolumntype{C}{>{\centering\arraybackslash}p{2em}}
    \caption{
        \label{tb:QueryGenComplete} Query generation examples on DuSinc for complete query dialogues.
    }
    \centering 
    \small
    \renewcommand\arraystretch{1.0}
\scalebox{0.95}{
\begin{tabular}{cp{3.0cm}p{3.0cm}p{0.5cm}}

\toprule
Dialogue History &
  \multicolumn{1}{c}{Ground-truth Query} &
  \multicolumn{1}{c}{Generated Query} &
  \multicolumn{1}{c}{Score} \\ \midrule

\begin{tabular}{@{}p{10.5cm}@{}}
    养生从什么时候开始比较合适啊？\textbf{(complete query)} \\ When is proper to start health maintenance? \textbf{(complete query)} \\ 
    \end{tabular} &
    \begin{tabular}{@{}p{3.0cm}@{}} 养生从什么时候开始比较合适 \\ When is proper to start health maintenance
    \end{tabular} &
    \begin{tabular}{@{}p{3.0cm}@{}} 养生从什么时候开始比较合适 \\ When is proper to start health maintenance
    \end{tabular} &
    \begin{tabular}{@{}p{0.5cm}@{}} 1.00
    \end{tabular} \\
    
    \cmidrule{1-4}

    \begin{tabular}{@{}p{10.5cm}@{}}
        《雍正王朝》和《步步惊心》都是关于清朝的电视剧。\\ \emph{Yongzheng Dynasty} and \emph{Treading On Thin Ice} are both TV dramas about the Qing Dynasty. \\ \vspace{-0.3em} 
        《步步惊心》违背历史吗？\textbf{(complete query)} \\ Does \emph{Treading On Thin Ice} run counter to history? \textbf{(complete query)} \\ 
    \end{tabular} &
    \begin{tabular}{@{}p{3.0cm}@{}} 《步步惊心》违背历史吗 \\ Does \emph{Treading On Thin Ice} run counter to history
    \end{tabular} &
    \begin{tabular}{@{}p{3.0cm}@{}} 《步步惊心》违背历史吗 \\ Does \emph{Treading On Thin Ice} run counter to history
    \end{tabular} &
    \begin{tabular}{@{}p{0.5cm}@{}} 1.00
    \end{tabular} \\

    \cmidrule{1-4}

    \begin{tabular}{@{}p{10.5cm}@{}}
      新手钓鱼要准备什么？\textbf{(complete query)} \\ What should novices prepare to go fishing? \textbf{(complete query)} \\ 
    \end{tabular} &
    \begin{tabular}{@{}p{3.0cm}@{}} 新手钓鱼要准备什么 \\ What should novices prepare to go fishing 
    \end{tabular} &
    \begin{tabular}{@{}p{3.0cm}@{}} 新手钓鱼要准备什么 \\ What should novices prepare to go fishing 
    \end{tabular} &
    \begin{tabular}{@{}p{0.5cm}@{}} 1.00
    \end{tabular} \\
    
    \bottomrule
    
\end{tabular}
}
\end{table*}
\end{CJK*}

%%% coreference & ellispe on DuSincR

% type 1
\begin{CJK*}{UTF8}{gbsn}
\begin{table*}[t]
\newcolumntype{?}{!{\vrule width 1pt}}
    \newcolumntype{C}{>{\centering\arraybackslash}p{2em}}
    \caption{
        \label{tb:dusincrtype1} Who or what question type query generation examples on DuSincR.
    }
    \centering 
    \small
    \renewcommand\arraystretch{1.0}
\scalebox{0.95}{
\begin{tabular}{cp{3.0cm}}

\toprule
Dialogue History &
    \multicolumn{1}{c}{Generated Query} \\
    \midrule

    \begin{tabular}{@{}p{15.0cm}@{}}
    最近知道一个叫迪丽热巴的演员，长得好漂亮呀！\\ I recently learned about an actor named Dilraba. She is so beautiful! \\ \vspace{-0.3em} 
    我也觉得她很漂亮，她是一个少数民族的人 \\ I also think she is beautiful. She is from a minority ethnic group. \\ \vspace{-0.3em}
    怪不得觉得她有点异域风情，\underline{她}\textbf{(coreference)}是哪个民族的？\textbf{(who or what question type)} \\ No wonder she is a little exotic. Which minority ethnic group is \underline{she} \textbf{(coreference)} from? \textbf{(who or what question type)}
    \end{tabular} &
    \begin{tabular}{@{}p{3.0cm}@{}} 迪丽热巴是哪个民族的 \\ Which ethnic group is Dilraba from
    \end{tabular} \\
    
    \cmidrule{1-2}

    \begin{tabular}{@{}p{15.0cm}@{}}
    % 黄芪有什么作用吗？我看医院里配的药里面黄芪就很常见~\\
    % Does astragalus have any effect? I think astragalus is very common in the medicine prepared in the hospital~\\ \vspace{-0.3em}
    % 黄芪能够治体虚自汗、失血体虚、血小板减少性紫癜、脑血栓和乳汁缺乏等~\\
    % Astragalus can treat body deficiency, spontaneous sweating, blood loss and body deficiency, thrombocytopenic purpura, cerebral thrombosis and milk deficiency~\\ \vspace{-0.3em}
    % 失血体虚？我之前就有过这种症状，配过中药，你知道配方吗？\\
    % Blood loss and body deficiency? I have had this symptom before, and have been mixed with traditional Chinese medicine. Do you know the formula?\\ \vspace{-0.3em}
    % 黄芪当归水，10克的当归和黄芪，可以起到非常关键的补气血的作用，如果症状严重的话，可以将10克改为15克~\\
    % Astragalus angelica water, 10 grams of angelica and astragalus can play a key role in replenishing qi and blood. If the symptoms are serious, you can change 10 grams to 15 grams~\\ \vspace{-0.3em}
    % 可以，那我顺便买点当归，一起煮一下，我妈妈贫血应该可以喝这个。\\
    % Yes, I'll buy some angelica and cook it together. My mother should be able to drink it for anemia.\\ \vspace{-0.3em}
    % 黄芪当归水对于贫血最管用了，可以提升造血功能，起到改善贫血的目的。\\
    % Astragalus angelica water is the most effective for anemia. It can improve hematopoietic function and improve anemia.\\ \vspace{-0.3em}
    OK，那我就熬一些给她喝，这个黄芪当归水需要喝多久？\\
    OK, then I'll boil some for her to drink. How long does this astragalus angelica water need to be taken?\\ \vspace{-0.3em}
    黄芪：当归=1，熬制半小时左右，我查一下需要喝多久，稍等。\\
    Astragalus : Angelica = 1, boil for about half an hour. I'll check how long it needs to be taken, wait a moment.\\ \vspace{-0.3em}
    好的，你慢慢看，影响不大，我已经到药店，先去买。\\
    Ok, take your time, it doesn't matter. I've already arrived at the drugstore, and I'll go buy it first.\\ \vspace{-0.3em}
    好的好的，可以。我查到了，不适合长期喝，一般喝1到2个月就应该停一段时间。\\
    OK, all right. I got it. It is not suitable for long-term drinking. Generally, it should be stopped after one to two months.\\ \vspace{-0.3em}
    好的，明白了，那喝\underline{这个}\textbf{(coreference)}水有什么服药禁忌吗？\textbf{(who or what question type)} \\
    Ok, I see. Is there any contraindications when drinking \underline{this} \textbf{(coreference)} water? \textbf{(who or what question type)}
    \end{tabular} &
    \begin{tabular}{@{}p{3.0cm}@{}} 喝黄芪当归水禁忌 \\ Contraindications of drinking astragalus and angelica water
    \end{tabular} \\

    \cmidrule{1-2}

    \begin{tabular}{@{}p{15.0cm}@{}}
    % 你知道什么是元宇宙吗？这个词最近可火了，但是我不知道它指的是什么。\\
    % Do you know what the metauniverse is? This word has become popular recently, but I don't know what it means.\\ \vspace{-0.3em}
    % 元宇宙其实就是一个虚拟世界，你可以在里面做任何自己想做的事情，比如说滑雪呀、看演唱会呀，在元宇宙里，只有想不到，没有做不到。\\ 
    % The metauniverse is actually a virtual world. You can do anything you want to do in it, such as skiing and watching concerts. In the metauniverse, you can only think of it.\\ \vspace{-0.3em}
    % 哇，听起来好酷的样子，感觉和VR挺像的。\\
    % Wow, it sounds so cool. It feels like VR.\\ \vspace{-0.3em}
    % 确实，元宇宙的技术基础就是VR和AR呢\\
    % Indeed, the technological basis of the metauniverse is VR and AR\\ \vspace{-0.3em}
    不过，我还是想象不出来元宇宙到底是什么样的。\\
    However, I still can't imagine what the metaverse is like.\\ \vspace{-0.3em}
    举个例子吧，你看过《头号玩家》吗，里面的“绿洲”其实就是一个元宇宙。\\
    For example, have you seen \emph{The Number One Player}? The \emph{Oasis} in it is actually a metaverse.\\ \vspace{-0.3em}
    原来如此，现在我明白多了。真希望有一天我们也可以拥有自己的元宇宙！\\
    I see. Now I understand more. I really hope that one day we can have our own metaverse!\\ \vspace{-0.3em}
    是呀，我也很期待那一天的到来呢。\\
    Yes, I'm also looking forward to that day.\\ \vspace{-0.3em}
    不过按照现在的发展水平，大概还需要很久吧！\\
    But according to the current development level, it will probably take a long time!\\ \vspace{-0.3em}
    那可不一定，等到云智能设备成熟了，元宇宙的实现应该就不远了吧。\\
    That's not necessarily true. When the technology of intelligent cloud devices is mature, the implementation of the metaverse should not be far away.\\ \vspace{-0.3em}
    \underline{它}\textbf{(coreference)}的概念是谁提出来的呢？\textbf{(who or what question type)} \\
    Who put forward \underline{its} \textbf{(coreference)} concept? \textbf{(who or what question type)}
    \end{tabular} &
    \begin{tabular}{@{}p{3.0cm}@{}} 元宇宙概念是谁提出来的 \\ Who put forward the concept of metaverse
    \end{tabular} \\
    
    \bottomrule
    
\end{tabular}
}
\end{table*}
\end{CJK*}

% type 2
\begin{CJK*}{UTF8}{gbsn}
\begin{table*}[t]
\newcolumntype{?}{!{\vrule width 1pt}}
    \newcolumntype{C}{>{\centering\arraybackslash}p{2em}}
    \caption{
        \label{tb:dusincrtype2} When or where question type query generation examples on DuSincR.
    }
    \centering 
    \small
    \renewcommand\arraystretch{1.0}
\scalebox{0.95}{
\begin{tabular}{cp{3.0cm}}

\toprule
Dialogue History &
\multicolumn{1}{c}{Generated Query} \\
  \midrule

  \begin{tabular}{@{}p{15.0cm}@{}}
    我想养一些紫罗兰，它好养吗？\\
    I want to raise some violets. Is it easy to raise?\\ \vspace{-0.3em} 
    很好养啊！适应能力很强，可以用叶插繁殖。\\
    It's easy to raise! It has strong adaptability and can be propagated with leaf cuttings.\\ \vspace{-0.3em} 
    是吗？那我现在就去市场看看，你知道哪里有卖这种花的吗？\\
    Is it? Then I will go to the market now. Do you know where they sell this kind of flower?\\ \vspace{-0.3em} 
    你可以去泉园二路花卉市场二楼万柳塘公园花鸟市场看看。\\
    You can go to the flower and bird market in Wanliutang Park on the second floor of the flower market in Quanyuan 2nd Road.\\ \vspace{-0.3em} 
    那我去也不能只买一种花呀，你给我推荐几种比较好养的花吧。\\
    But I can't buy only one kind of flower. Please recommend some flowers that are easy to grow. \\ \vspace{-0.3em} 
    你可以买一些太阳花，铜钱草，长寿花，都是特别好养的花。\\
    You can buy some sunflowers, hydrocotyle, and longevity flowers, which are especially easy to grow.\\ \vspace{-0.3em} 
    \underline{那个}\textbf{(coreference)}花鸟市场一般什么时候开放呢？\textbf{(when or where question type)}\\
    When does \underline{that} \textbf{(coreference)} flower and bird market usually open? \textbf{(when or where question type)}
  \end{tabular} &
  \begin{tabular}{@{}p{3.0cm}@{}} 万柳塘公园花鸟市场开放时间 \\ Opening hours of flower and bird market in Wanliutang Park
  \end{tabular} \\

  \cmidrule{1-2}

  \begin{tabular}{@{}p{15.0cm}@{}}
    % 最近有啥好玩的游戏啊！\\
    % What interesting games are there recently!\\ \vspace{-0.3em}
    % 有很多呀～比如王者荣耀啦，英雄联盟啦还有很多DOTA，炉石传说等等\\
    % There are many, such as the King Rong Yao, the League of Heroes, and many DOTA, Hearthstone legends, etc\\ \vspace{-0.3em}
    % 诶？炉石传说是什么\\
    % Hey? What is the legend of furnace stone\\ \vspace{-0.3em}
    % 暴雪娱乐发行的一款集换式卡牌游戏哦～很多人都评价这款游戏轻松有趣呢～\\
    % Blizzard Entertainment released an interactive card game. Many people commented that this game is easy and interesting\\ \vspace{-0.3em}
    % 哇哦！那我有机会去试一试！\\
    % wow Then I have a chance to try!\\ \vspace{-0.3em}
    % 是呀，希望你喜欢！\\
    % Yes, I hope you like it!\\ \vspace{-0.3em}
    有没有什么适合女生玩的，感觉这个有点男生玩得多吧。\\
    Is there anything suitable for girls to play? I think this is a little more for boys. \\ \vspace{-0.3em}
    哈哈哈哈哈好呀！你知道光遇这款游戏吗？\\
    Haha! Do you know the game Sky: Children of the Light?\\ \vspace{-0.3em}
    这是什么！？\\
    What is this?\\ \vspace{-0.3em}
    是一款可能女生玩得比较多的游戏哦～目标就是寻找游戏中的子民以及收集蜡烛～简单吧？～\\
    It's a game that girls may play more. The goal is to find citizens in the game and collect candles. Is it easy, right?\\ \vspace{-0.3em}
    \underline{这款}\textbf{(coreference)}游戏是什么时候发行的？\textbf{(when or where question type)}\\
    When was \underline{this} \textbf{(coreference)} game released? \textbf{(when or where question type)}
  \end{tabular} &
  \begin{tabular}{@{}p{3.0cm}@{}} 光遇这款游戏发行时间 \\ The release time of the game Sky: Children of the Light
  \end{tabular} \\
  
  \cmidrule{1-2}

  \begin{tabular}{@{}p{15.0cm}@{}}
    天啊我要和男朋友一起去旅行啦！\\ 
    Oh my god, I'm going to go on a trip with my boyfriend!\\ \vspace{-0.3em} 
    妈耶真替你高兴！\\ 
    Wow, I'm so happy for you!\\  \vspace{-0.3em}
    嘿嘿从来没有和男生一起去旅行。\\ 
    Chuckle, I've never went on a trip with a boy.\\  \vspace{-0.3em}
    哈哈这不就是第一次嘛～两个人在一起呀～舒适感特别特别特别重要，所以去一些环境好、景色美的地方呀～让两个人时刻保持愉快的心情呢！\\ 
    Ha ha, then this will be the first time. When two people are together, the comfort is particularly important. So go to some places with good environment and beautiful scenery. Make sure you two are in good mood at all times!\\  \vspace{-0.3em}
    嗯嗯是的。。\\ 
    Mm-hmm. Yes.\\  \vspace{-0.3em}
    你们打算去哪里啊？\\ 
    Where are you going?\\  \vspace{-0.3em}
    我们还没想好。\\ 
    We haven't decided yet.\\  \vspace{-0.3em}
    西安吧，是世界历史文化名城，可以去玩一玩啊！\\ 
    Try Xi'an, it's a world famous historical and cultural city. Go and play!\\  \vspace{-0.3em}
    那是很出名的一座城市，你知道\underline{它}\textbf{(coreference)}有什么好玩的地方吗？\textbf{(when or where question type)}\\ 
    It is a very famous city. Do you know any interesting places in \underline{it} \textbf{(coreference)}? \textbf{(when or where question type)}
    \end{tabular} &
    \begin{tabular}{@{}p{3.0cm}@{}} 西安好玩的地标 \\ Interesting landmarks in Xi'an
    \end{tabular} \\
    
  \bottomrule
    
\end{tabular}
}
\end{table*}
\end{CJK*}

% type 3
\begin{CJK*}{UTF8}{gbsn}
\begin{table*}[t]
\newcolumntype{?}{!{\vrule width 1pt}}
    \newcolumntype{C}{>{\centering\arraybackslash}p{2em}}
    \caption{
        \label{tb:dusincrtype3} Count question type query generation examples on DuSincR.
    }
    \centering 
    \small
    \renewcommand\arraystretch{1.0}
\scalebox{0.95}{
\begin{tabular}{cp{3.0cm}}

\toprule
Dialogue History &
    \multicolumn{1}{c}{Generated Query} \\
    \midrule

  \begin{tabular}{@{}p{15.0cm}@{}}
    情人节我该送什么礼物呀？ \\
    What gift should I give on Valentine's Day? \\ \vspace{-0.3em} 
    送她充满爱意的礼物，你可以送她玫瑰花啊。 \\
    Send her a loving gift. You can give her roses. \\ \vspace{-0.3em} 
    什么颜色比较好呀？我在花店看到好多种颜色？ \\
    What color is better? I saw many colors in the flower shop. \\ \vspace{-0.3em} 
    红色玫瑰代表炽热的爱情，白玫瑰代表纯洁的爱情，但是大家一般送红色的。 \\
    Red roses represent passionate love, white roses represent pure love, but people usually give red ones. \\ \vspace{-0.3em} 
    好滴，她应该会喜欢的，一般要送多少朵呢？\textbf{(ellipse)}\ \textbf{(count question type)} \\
    Okay, I guess she would like it. How many should I give? \textbf{(ellipse)}\ \textbf{(count question type)} 
    % \\ \vspace{-0.3em} 
    % 你不要送一枝，有点太寒酸了，最好送999朵。 \\
    % Don't send one. It's a little too shabby. You'd better send 999. \\ \vspace{-0.3em} 
    % 啊，999朵，我的钱包受不了啊 \\
    % Ah, 999, my wallet can't stand it \\ \vspace{-0.3em} 
    % 也是，大概需要5000块呢，的确有些贵了。 \\
    % Yes, it costs about 5000 yuan, which is really a little expensive. \\ \vspace{-0.3em} 
    % 那99朵多少钱？ \\
    % How much is 99?
  \end{tabular} &
  \begin{tabular}{@{}p{3.0cm}@{}} 
  % 情人节99朵玫瑰要多少钱 \\ How much is 99 roses for Valentine's Day
  情人节送多少玫瑰花合适 \\ How many roses are suitable to give for Valentine's Day
  \end{tabular} \\

  \cmidrule{1-2}
  \begin{tabular}{@{}p{15.0cm}@{}}
    % 放寒假啦！\\
    % It's winter vacation!\\ \vspace{-0.3em} 
    % 哇塞，一学期紧张忙碌的学习生活，辛苦啦！\\
    % Wow, a semester of intense and busy study and life, hard!\\ \vspace{-0.3em} 
    % 嘿嘿，回家之前，我还准备找个地方旅游，有什么推荐的吗？\\
    % Hey hey, before I go home, I'm going to find a place to travel. What do you recommend?\\ \vspace{-0.3em} 
    % 不是我吹，咱这儿好玩儿的风景名胜可太多啦，最近的就是天井湖景区，距离702米\\
    % Not me. There are too many interesting scenic spots here. The nearest is Tianjing Lake Scenic Spot, 702 meters away\\ \vspace{-0.3em} 
    我不想在安徽玩儿，哈尔滨呢，哈尔滨最近天气如何，想去冰雪大世界！\\
    I don't want to play in Anhui. What about Harbin? What's the weather like in Harbin recently? I want to go to the Ice and Snow World!\\ \vspace{-0.3em} 
    啊，哈尔滨可冷啦，最近也是天晴，是温度在-21度到-12度之间，去那儿多加点衣服吧！\\
    Ah, it's very cold in Harbin. It's sunny recently. The temperature is between -21 degrees and -12 degrees. Go there and add more clothes!\\ \vspace{-0.3em} 
    你可真贴心。对啦，冰雪大世界最近开放吗？别白跑一趟。\\
    You are so sweet. By the way, is the Ice and Snow World open recently? Don't go for nothing.\\ \vspace{-0.3em} 
    据我所知是开放的，一般来说从每年的圣诞节到次年三月初期间都开放！现在一月，正是好玩的时候！\\
    As far as I know, it is open. Generally, it is open from Christmas to early March of the next year! Now January is just the time for fun!\\ \vspace{-0.3em} 
    \underline{那里}\textbf{(coreference)}的门票多少钱你知道吗？\textbf{(count question type)}\\
    Do you know how much the tickets cost \underline{there} \textbf{(coreference)}? \textbf{(count question type)}
  \end{tabular} &
  \begin{tabular}{@{}p{3.0cm}@{}} 
  哈尔滨冰雪大世界门票 \\ Tickets for Harbin Ice and Snow World
  \end{tabular} \\

  \cmidrule{1-2}
  \begin{tabular}{@{}p{15.0cm}@{}}
    看《令人心动的offer3》被周深吸引了～\\
    I was attracted by Zhou Shen after watching \emph{The Amazing Offer 3}. \\ \vspace{-0.3em} 
    你有眼光啊～我也看了，我觉得他很幽默，风趣。\\
    You've got good taste. I have seen it too. I think he is very humorous and interesting.\\ \vspace{-0.3em} 
    是的呀，觉得他说话很有意思，唱歌非常的空灵～\\
    Yes, I think he speaks interestingly and his singing is very ethereal.\\ \vspace{-0.3em} 
    我听过他的歌曲，比如说大鱼、无关、花开、水形物语、如果你爱我等等。我主要是听他的《大鱼》才认识他的～\\
    I have heard his songs, such as Big Fish, Irrelevant, Blooming Flowers, The Shape of Water, If You Love Me, and so on. I knew him by listening to his \emph{Big Fish}.\\ \vspace{-0.3em} 
    你这么一说，我都听过呀～\\
    I've heard all of what you've said. \\ \vspace{-0.3em} 
    是的呀，我和你说，周深的《大鱼》非常好听，绝对是忘不掉的经典时刻，这样空灵而纯净的声音实在是难得啊～百年难得一见啊～\\
    Yes, and I have to tell you, Zhou Shen's \emph{Big Fish} is very good, and it is definitely a classic moment that cannot be forgotten. Such an ethereal and pure voice is really rare. It's rare to hear it even in a hundred years. \\ \vspace{-0.3em} 
    对对对，我就是被这个空灵的声音所吸引～而且我觉得他对待人也是蛮好的。\\
    Right, I was attracted by this ethereal voice, and I also think he treats people very well.\\ \vspace{-0.3em} 
    是的，他接人待物很谦虚谨慎，温柔的很，从他的身上获取了很多正能量，从很多细节当中都是可以体现出来的。\\
    Yes, he is very modest and gentle when dealing with people and things. I have obtained a lot of positive energy from him, which can be reflected in many details on him.\\ \vspace{-0.3em} 
    是的，他还很聪明，你知道\underline{他}\textbf{(coreference)}会几门语言吗？\textbf{(count question type)} \\
    Yes, and he is also very smart. Do you know how many languages \underline{he} \textbf{(coreference)} can speak? \textbf{(count question type)}
  \end{tabular} &
  \begin{tabular}{@{}p{3.0cm}@{}} 
  周深会几门语言 \\ How many languages Zhou Shen knows
  \end{tabular}
  \\

  \bottomrule
    
\end{tabular}
}
\end{table*}
\end{CJK*}

% type 4
\begin{CJK*}{UTF8}{gbsn}
\begin{table*}[t]
\newcolumntype{?}{!{\vrule width 1pt}}
    \newcolumntype{C}{>{\centering\arraybackslash}p{2em}}
    \caption{
        \label{tb:dusincrtype4} Select among question type query generation examples on DuSincR.
    }
    \centering 
    \small
    \renewcommand\arraystretch{1.0}
\scalebox{0.95}{
\begin{tabular}{cp{3.0cm}}

\toprule
Dialogue History &
    \multicolumn{1}{c}{Generated Query} \\
    \midrule

  \begin{tabular}{@{}p{15.0cm}@{}}
    % 在干嘛 \\ 
    % What's going on? \\  \vspace{-0.3em} 
    % 找歌曲听歌呢 \\ 
    % Find songs to listen to \\  \vspace{-0.3em} 
    
    你喜欢周杰伦的歌吗？ \\ 
    Do you like Jay Chou's songs? \\  \vspace{-0.3em} 
    喜欢啊，他的歌我都听，像《可爱女人》、《简单爱》、《龙卷风》、《以父之名》、《爱情悬崖》、《七里香》、《将军》等等。 \\ 
    Yes, I listen to all his songs, such as \emph{Lovely Woman}, \emph{Simple Love}, \emph{Tornado}, \emph{In the Name of Father}, \emph{Love Cliff}, \emph{Seven Mile Fragrance}, \emph{General} and so on. \\  \vspace{-0.3em} 
    哦哦，看来我们有共同的偶像了，我很喜欢他的一首《青花瓷》。 \\ 
    Oh, it seems that we have the same idol. I like his song \emph{Blue and White Porcelain} very much. \\  \vspace{-0.3em} 
    《青花瓷》这首歌很经典，他出自《我很忙》这个专辑。在当时台湾第19届金曲奖典礼上获得了年度最佳歌曲奖、作词奖以及作曲奖三项奖项。 \\ 
    The song \emph{Blue and White Porcelain} is very classic. It comes from the album \emph{I'm Busy}. At the 19th Golden Melody Awards ceremony in Taiwan at that time, he won that year's Best Song Award, Best Lyrics Award and Best Composing Award. \\  \vspace{-0.3em} 
    是的，我很喜欢周杰伦中国风得歌曲，很有感觉。 \\ 
    Yes, I like Jay Chou's Chinese style songs very much, they really have this feeling. \\  \vspace{-0.3em} 
    那你肯定还喜欢《东风破》、《兰亭序》、《发如雪》。《千里之外》跟《烟花易冷》了，都是周杰伦中国风的代表。 \\ 
    Then you must also like \emph{Breaking the East Wind}, \emph{Preface to Orchid Pavilion} and \emph{Hair Like Snow}. \emph{Thousand Miles Away} and \emph{Cold Fireworks} are representatives of Jay Chou's Chinese style. \\  \vspace{-0.3em} 
    嗯嗯，是的，这些我都喜欢，你知道\underline{他}\textbf{(coreference)}销量最高的是哪个专辑吗？\textbf{(select among question type)} \\ 
    Mm-hmm. Yes, I like all of them. Do you know which album \underline{he} \textbf{(coreference)} sells the most? \textbf{(select among question type)}
  \end{tabular} &
  \begin{tabular}{@{}p{3.0cm}@{}} 周杰伦销量最高专辑 \\ Jay Chou's best-selling album
  \end{tabular} \\ 

  \cmidrule{1-2}

  \begin{tabular}{@{}p{15.0cm}@{}}
    % 《开端》太好看了，更新的太少，简直不过瘾啊~\\
    % "The Beginning" is too good to watch. There are too few updates. It's simply not enjoyable~\\ \vspace{-0.3em} 
    % 我知道！这部剧是由“正午阳光”出品的，据说“正午阳光”出品的都是精品！还是由白敬亭演的。\\
    % I Know! This play is produced by "Noon Sunshine". It is said that "Noon Sunshine" produces high-quality products! It was played by Bai Jingting.\\ \vspace{-0.3em} 
    哈哈哈哈！我就猜你看了！因为你喜欢白敬亭。\\
    Ha ha! I guess you saw it! Because you like Bai Jingting.\\ \vspace{-0.3em} 
    白敬亭可是“怀柔小王子”，人也帅气，演技也很好。\\
    Bai Jingting is the "little prince of Huairou". He is handsome and has good acting skills.\\ \vspace{-0.3em} 
    哈哈哈哈，那忠实小粉丝，是不是\underline{他}\textbf{(coreference)}的剧你都看过了呀。知道哪一部是评分最高的呢？\textbf{(ellipse)}\ \textbf{(select among question type)} \\
    Ha ha, then loyal little fan, have you seen all \underline{his} \textbf{(coreference)} TV dramas? Do you know which one has the highest score? \textbf{(ellipse)} \textbf{(select among question type)}
  \end{tabular} &
  \begin{tabular}{@{}p{3.0cm}@{}} 评分最高的白敬亭电视剧 \\ Bai Jingting TV drama with the highest score
  \end{tabular} \\

  \cmidrule{1-2}

  \begin{tabular}{@{}p{15.0cm}@{}}  
    我想买一双篮球鞋，你知道哪个牌子的更好吗？\\
    I want to buy a pair of basketball shoes. Do you know which brand is better?\\ \vspace{-0.3em} 
    我穿的是李宁的，我队友他们穿阿迪耐克的比较多。\\
    I wear Li Ning's, and my teammates wear Adidas and Nike's more. \\ \vspace{-0.3em} 
    你觉得穿起来怎么样？\\
    How do you feel about it?\\ \vspace{-0.3em} 
    很不错啊，不变形，减震也很好，我都穿了一个学期了。\\
    It's very good. It doesn't deform. It also has good shock absorption. I've already been wearing it for a whole semester. \\ \vspace{-0.3em} 
    好啊好啊，那我下午就去买一双。你有时间吗？陪我一起去吧！\\
    Okay, I will buy a pair this afternoon. Do you have time? Come with me!\\ \vspace{-0.3em} 
    不用了吧，咱们两家离的太远了。\\
    No, we are too far away.\\ \vspace{-0.3em} 
    哦，好吧，那我就从网上买吧，你知道在哪里买吗？\\
    Oh, well, then I'll buy it online. Do you know where to buy it?\\ \vspace{-0.3em} 
    好乐买，名鞋库还有毒上面都有卖\\
    OkBuy, Skomart and De Wu all sell them. \\ \vspace{-0.3em} 
    对了，你知道\underline{这个}\textbf{(coreference)}牌子的女鞋最大码是多少吗，我的脚比较大？\textbf{(select among question type)}\\
    By the way, do you know the maximum size of women's shoes of \underline{this} \textbf{(coreference)} brand, since my feet are quite big? \textbf{(select among question type)}
  \end{tabular} &
  \begin{tabular}{@{}p{3.0cm}@{}} 李宁女鞋最大码 \\ Maximum size of Li Ning women's shoes
  \end{tabular} 
\\

  \bottomrule
    
\end{tabular}
}
\end{table*}
\end{CJK*}

% type 5
\begin{CJK*}{UTF8}{gbsn}
\begin{table*}[t]
\newcolumntype{?}{!{\vrule width 1pt}}
    \newcolumntype{C}{>{\centering\arraybackslash}p{2em}}
    \caption{
        \label{tb:dusincrtype5} Comparison question type query generation examples on DuSincR.
    }
    \centering 
    \small
    \renewcommand\arraystretch{1.0}
\scalebox{0.95}{
\begin{tabular}{cp{3.0cm}}

\toprule
Dialogue History &
    \multicolumn{1}{c}{Generated Query} \\
    \midrule

  \begin{tabular}{@{}p{15.0cm}@{}}
    % 我不会贴假睫毛呀，好难！ \\ 
    % I can't stick false eyelashes, it's so difficult! \\ \vspace{-0.3em} 
    % 这个我专门了解过的。先把真睫毛涂一层睫毛膏卷起来定型，然后把假睫毛剪下来揉软，再把睫毛贴上胶水贴到眼皮上。 \\ 
    % I have learned about this. First, apply a layer of eye black on the real eyelashes, roll them up and shape them, then cut off the false eyelashes and rub them soft, and then stick the eyelashes onto the eyelids with glue. \\ \vspace{-0.3em} 
    % 哇！好厉害啊，你会贴么？ \\ 
    % Wow! How powerful! Can you paste it? \\ \vspace{-0.3em} 
    % 哈哈哈，我只会说。但是实际操作也不会 \\ 
    % Hahaha, I can only say. But the actual operation will not \\ \vspace{-0.3em} 
    % 哈哈，我视频倒是看了很多。但是手残 \\ 
    % Haha, I've seen a lot of videos. But the hand is damaged \\ \vspace{-0.3em} 
    % 我也是！手残党啊 \\ 
    % Me too. Handicapped party \\ \vspace{-0.3em} 
    假睫毛会对人有伤害么？ \\ 
    Can fake eyelashes harm people? \\ \vspace{-0.3em} 
    用胶水很容易造成眼睛发炎，所以一定要小心。 \\ 
    Using glue can easily cause eye irritation, so be careful. \\ \vspace{-0.3em} 
    嗯嗯，看来还是去接睫毛好一点。 \\ 
    Well, it seems better to get eyelash extension. \\ \vspace{-0.3em} 
    是的。我也打算去呢。 \\ 
    Yes. I also plan to go. \\ \vspace{-0.3em} 
    那\underline{这个}\textbf{(coreference)}和贴假睫毛有什么区别? \textbf{(comparison question type)} \\ 
    What's the difference between \underline{this one} \textbf{(coreference)} and sticking fake eyelashes? \textbf{(comparison question type)}
  \end{tabular} &
  \begin{tabular}{@{}p{3.0cm}@{}} 接睫毛与假睫毛的区别  \\ The difference between eyelash extension and fake eyelashes
  \end{tabular} \\ 

  \cmidrule{1-2}
  \begin{tabular}{@{}p{15.0cm}@{}}
    % 你会摄影吗？\\
    % Can you take pictures?\\ \vspace{-0.3em} 
    % 还行哈~你呢，喜欢摄影么\\
    % Not bad. How about you? Do you like photography\\ \vspace{-0.3em} 
    % 我喜欢摄影呢，喜欢拍日出的景色。\\
    % I like photography. I like taking pictures of sunrise.\\ \vspace{-0.3em} 
    % 我也喜欢~我觉得拍摄的时间和对云雾的表现最重要，比如最好日出前半小时，日落后半小时来拍摄，云雾是很重要的题材哈~你觉得呢\\
    % I also like it. I think the time of shooting and the performance of clouds are the most important. For example, it's better to shoot half an hour before sunrise and half an hour after sunset. Clouds and fog are very important subjects. What do you think\\ \vspace{-0.3em} 
    % 你说得对，日出时太阳在云雾中喷薄而出的景象真的太壮观了！\\
    % You are right. The scene of the sun blowing out in the clouds at sunrise is really spectacular!\\ \vspace{-0.3em} 
    % 是呀~我还记得我去黄山玩的时候，拍的云海日落，那景象真实太壮观了\\
    % Yeah, I still remember the sunset in the sea of clouds when I went to Huangshan. It was really spectacular\\ \vspace{-0.3em} 
    嗯嗯，除了黄山，你还知道有别的地方适合拍日出日落的景色吗？\\
    Well, besides Huangshan, do you know any other places suitable for taking pictures of sunrise and sunset?\\ \vspace{-0.3em} 
    泰山也很不错～玉皇顶是很好的拍摄位置。 \\
    Mount Tai is also very good. Yuhuangding is a good shooting location. \\ \vspace{-0.3em} 
    说到泰山，你知道\underline{它}\textbf{(coreference)}和黄山哪个更高一点吗？\textbf{(comparison question type)} \\
    Speaking of Mount Tai, do you know which is higher, \underline{it} \textbf{(coreference)} or Mount Huang? \textbf{(comparison question type)}
  \end{tabular} &
  \begin{tabular}{@{}p{3.0cm}@{}} 泰山和黄山哪个更高 \\ Which is higher, Mount Tai or Mount Huang
  \end{tabular} \\

  \cmidrule{1-2}
  \begin{tabular}{@{}p{15.0cm}@{}}
    去逛街喽，买点水果。\\
    I'm going shopping, and buying some fruit.\\ \vspace{-0.3em} 
    去买烟台苹果哈哈哈哈，果肉甜脆的红富士\\
    Go buy Yantai apples, red Fuji with sweet and crisp pulp. \\ \vspace{-0.3em} 
    苹果那必须得整一些，不过还打算买别的水果，买点香蕉吧。\\
    I have to get some apples, but I also plan to buy other fruits, maybe some bananas.\\ \vspace{-0.3em} 
    多吃香蕉好，补充能量、润肠通便、助睡眠、防治高血压都得看咱香蕉。\\
    It's good to eat more bananas, which can supplement energy, moisten intestines and relieve constipation, help sleep and prevent hypertension. \\ \vspace{-0.3em} 
    啊，香蕉还能防高血压啊？\\
    Ah, bananas can prevent hypertension?\\ \vspace{-0.3em} 
    对啊，香蕉中含有丰富的钾离子，钾离子有拮抗钠离子收缩血管作用。常食香蕉有助于维持体内钠、钾平衡和酸碱平衡，以便使神经肌肉保持正常，心肌收缩协调从而起到防治高血压的作用。是不是大开眼界了哈哈哈哈！\\
    Yes, bananas are rich in potassium ions, which can antagonize the contraction of blood vessels by sodium ions. Eating bananas regularly helps to maintain the balance of sodium, potassium and acid-base in the body, in order to maintain the normal neuromuscular function and the coordination of myocardial contraction, thus preventing and treating hypertension. Surely widening your horizon, right? Ha ha! \\ \vspace{-0.3em} 
    的确，没想到香蕉还有这功效。\\
    Indeed, I didn't expect bananas to have this effect. \\ \vspace{-0.3em} 
    不过要特别注意生香蕉和熟香蕉的区别。熟香蕉可以通便，但是生香蕉是止泻的作用，吃多了可是会便秘的。\\
    However, pay special attention to the difference between raw bananas and ripe bananas. Ripe bananas can relieve constipation, but raw bananas stop diarrhea, so eating too much can cause constipation.\\ \vspace{-0.3em} 
    哦，那你觉得\underline{它}\textbf{(coreference)}和苹果哪个营养价值比较高呢？\textbf{(comparison question type)} \\
    Oh, which nutritional value do you think is higher, \underline{them} \textbf{(coreference)} or apples? \textbf{(comparison question type)}
  \end{tabular} &
  \begin{tabular}{@{}p{3.0cm}@{}} 香蕉和苹果哪个营养价值高 \\ Which is more nutritious, banana or apple
  \end{tabular} \\
  
  \bottomrule
    
\end{tabular}
}
\end{table*}
\end{CJK*}

% type 6
\begin{CJK*}{UTF8}{gbsn}
\begin{table*}[t]
\newcolumntype{?}{!{\vrule width 1pt}}
    \newcolumntype{C}{>{\centering\arraybackslash}p{2em}}
    \caption{
        \label{tb:dusincrtype6} Verify question type query generation examples on DuSincR.
    }
    \centering 
    \small
    \renewcommand\arraystretch{1.0}
\scalebox{0.95}{
\begin{tabular}{cp{3.0cm}}

\toprule
Dialogue History &
\multicolumn{1}{c}{Generated Query} \\
  \midrule

  \begin{tabular}{@{}p{15.0cm}@{}}
    % 朋友，你知道绿萝怎么修剪吗？\\
    % Friend, do you know how to trim green apples?\\ \vspace{-0.3em} 
    % 绿萝很好打理吧。修剪病黄叶、超过20cm需要对其进行修剪，如果出现病根，腐烂也要拔除哦\\
    % Green apples are easy to take care of. The diseased yellow leaves need to be pruned if they are more than 20cm long. If the diseased roots occur, the rotten leaves should also be uprooted\\ \vspace{-0.3em} 
    我看大家都有绿萝在办公室，绿萝植物有很多好处吧。\\
    I think everyone has a green vine in the office. Green vines must have many advantages. \\ \vspace{-0.3em} 
    嗯，绿蓓蓓的，看着心情都好了。绿萝除了美化环境，还可以减除压力、净化空气，它能吸收空气中甲醛等，刚装修好的新居摆放绿萝就醉适合不过了。\\
    Yeah, the shining green gives a good mood. In addition to beautifying the environment, green vine can also reduce pressure and purify the air. It can absorb formaldehyde in the air and so on. It is suitable for a newly decorated house to have a green vine.\\ \vspace{-0.3em} 
    怪不得办公室、家庭、商业区都有绿萝的出现。绿萝感觉很好养。\\
    No wonder there are green vines in offices, homes and business districts. The green vine seems easy to look after.\\ \vspace{-0.3em} 
    嗯，好养的很。不要很多阳光，只要定期浇水就没问题啦。\\
    Well, it's easy to look after. Doesn't need much sunshine. Just water it regularly.\\ \vspace{-0.3em} 
    真好，平时忙的话，么有太多时间照顾，自力更生能力很强，适合我。\\
    It's good. If I'm busy at ordinary times, I don't have much time to take care of it. I have a strong ability of self-reliance, which is suitable for me.\\ \vspace{-0.3em} 
    是的，它会给你带来很多开心和好运气呢。\\
    Yes, it will bring you a lot of happiness and good luck.\\ \vspace{-0.3em} 
    那\underline{它}\textbf{(coreference)}会不会开花啊？\textbf{(verify question type)} \\
    Then will \underline{it} \textbf{(coreference)} bloom? \textbf{(verify question type)}
  \end{tabular} &
  \begin{tabular}{@{}p{3.0cm}@{}} 绿萝会开花吗 \\ Can green vines bloom
  \end{tabular} \\

  \cmidrule{1-2}
  \begin{tabular}{@{}p{15.0cm}@{}}
    你喜欢玩什么游戏啊？\\
    What game do you like to play?\\ \vspace{-0.3em} 
    喜欢玩光遇，你知道吗？\\
    I like to play Sky: Children of the Light. Do you know?\\ \vspace{-0.3em} 
    这个是什么呀？\\
    What is this?\\ \vspace{-0.3em} 
    陪伴让前行之路不再孤单，这就是这个游戏的特色哦～\\
    Companion makes the way forward no longer lonely, which is the feature of this game.\\ \vspace{-0.3em} 
    在里面可以交朋友的那种吗？\\
    Can you make friends in it?\\ \vspace{-0.3em} 
    是呀，推荐你玩一玩！真的很温暖的一个游戏。\\
    Yes, I recommend you to play! It's a really warm game.\\ \vspace{-0.3em} 
    \underline{这个}\textbf{(coreference)}游戏需要氪金吗？\textbf{(verify question type)} \\
    Do we need to top up money to \underline{this} \textbf{(coreference)} game? \textbf{(verify question type)}
  \end{tabular} &
  \begin{tabular}{@{}p{3.0cm}@{}} 光遇氪金 \\ Top on money to Sky: Children of the Light
  \end{tabular} \\

  \cmidrule{1-2}
  \begin{tabular}{@{}p{15.0cm}@{}}
    % 我好无聊啊，而且肚子好饿。\\
    % I'm so bored and hungry.\\ \vspace{-0.3em} 
    % 你可以出去买点零食呀，离你家八百米就有一个超市呢。\\
    % You can go out and buy some snacks. There is a supermarket 800 meters away from your home.\\ \vspace{-0.3em} 
    % 好主意，我这就去买吃的。你有推荐的零食吗？\\
    % That's a good idea. I'm going to buy food now. Do you have any recommended snacks?\\ \vspace{-0.3em} 
    % 看你想吃什么咯，我比较爱吃黑芝麻饼和核桃贵妃饼。\\
    % See what you want to eat. I prefer black sesame cake and walnut imperial concubine cake.\\ \vspace{-0.3em} 
    我不太喜欢吃饼干，吃饼干容易上火。\\
    I don't like cookies very much. Eating cookies are easy to dehydrate.\\ \vspace{-0.3em} 
    那你可以买果冻或者酸奶啊，这些都不上火。\\
    Then you can buy jelly or yogurt, which won't make you dehydrate.\\ \vspace{-0.3em} 
    有道理，我去买点酸奶吧，好久没喝了。\\
    It makes sense. I'll buy some yogurt. I haven't drunk it for a long time.\\ \vspace{-0.3em} 
    我推荐你买卡士的酸奶，里面加了奶油，口感很不错哦。\\
    I recommend you Classy Kiss yogurt, which has cream in it. It tastes very good.\\ \vspace{-0.3em} 
    我还没喝过这个呢，\underline{这}\textbf{(coreference)}是进口的酸奶吗？\textbf{(verify question type)} \\
    I haven't drunk this yet. Is \underline{this} \textbf{(coreference)} imported yogurt? \textbf{(verify question type)}
  \end{tabular} &
  \begin{tabular}{@{}p{3.0cm}@{}} 卡士酸奶进口吗 \\ Is Classy Kiss yogurt imported
  \end{tabular} \\
  
  \bottomrule
    
\end{tabular}
}
\end{table*}
\end{CJK*}

% type 7
\begin{CJK*}{UTF8}{gbsn}
\begin{table*}[t]
\newcolumntype{?}{!{\vrule width 1pt}}
    \newcolumntype{C}{>{\centering\arraybackslash}p{2em}}
    \caption{
        \label{tb:dusincrtype7} How question type query generation examples on DuSincR.
    }
    \centering 
    \small
    \renewcommand\arraystretch{1.0}
\scalebox{0.95}{
\begin{tabular}{cp{3.0cm}}

\toprule
Dialogue History &
    \multicolumn{1}{c}{Generated Query} \\
    \midrule
  
  \begin{tabular}{@{}p{15.0cm}@{}}
    我刚才买了个甘蔗结果一点都不甜还咬不动。\\
    I just bought a sugar cane and it turned out not to be sweet at all and too tough to bite.\\ \vspace{-0.3em} 
    啊是么我前天买的还挺好吃的。\\
    Ah, really, the one I bought the day before yesterday was quite good.\\ \vspace{-0.3em} 
    哎我不懂怎么挑选啊？你知道怎么挑选\underline{它}\textbf{(coreference)}吗？\textbf{(how question type)} \\
    Sigh, I don't know how to choose. Do you know how to choose \underline{it} \textbf{(coreference)}? \textbf{(how question type)}
  \end{tabular} &
  \begin{tabular}{@{}p{3.0cm}@{}} 甘蔗怎么挑选 \\ How to choose sugarcane
  \end{tabular} \\

  \cmidrule{1-2}
  \begin{tabular}{@{}p{15.0cm}@{}}
    租了一个单身的屋子，是时候养一只狗了。\\
    Having rented a single house, it's time to raise a dog.\\ \vspace{-0.3em} 
    的确，有一只狗狗会比较安心。\\
    Indeed, having a dog is more reassuring.\\ \vspace{-0.3em} 
    你觉得我养什么类型的狗狗比较好呢？\\
    What kind of dog do you think is better for me?\\ \vspace{-0.3em} 
    我觉得如果你想养来看家和保护你的话，你可以试试养一只中华田园犬、罗威纳德国牧羊犬或者杜宾犬，他们有很强的护主意识，还很聪明呢。\\
    I think if you want to keep one to look after the house and protect you, then you can try a Chinese Pastoral Dog, a Rottweiler German Shepherd Dog or a Doberman. They have a strong sense of protection and are very smart. \\ \vspace{-0.3em} 
    那我去看看这三只，第一次养狗没有什么经验，不知道要准备什么。\\
    Let me go and see these three kinds. It's my first time raising dogs and I have no experience. I don't know what to prepare. \\ \vspace{-0.3em} 
    新人养狗的话的确要准备很多东西，食盆、窝、狗粮、宠物专用厕所、宠物干洗粉、小玩具、牵引绳这些你要先备好，这些物质类的东西除外，你还要考虑自己有没有长期的坚持心理，才能去领养一只狗狗哦。\\
    Newbies who raises a dog sure have to prepare a lot of things, such as food basin, kennel, dog food, pet toilet, pet dry-cleaning powder, small toys, and traction rope. Besides these things, you should also consider whether you have a long-term persistence psychology before you can adopt a dog. \\ \vspace{-0.3em} 
    如果出远门，我要怎么安排\underline{他们}\textbf{(coreference)}啊？\textbf{(how question type)} \\
    If I go for a long trip, how can I arrange \underline{them} \textbf{(coreference)}? \textbf{(how question type)}
  \end{tabular} &
  \begin{tabular}{@{}p{3.0cm}@{}} 出远门狗狗该怎么办 \\ What to do with the dog when going for a long trip
  \end{tabular} \\
  
  \cmidrule{1-2}
  \begin{tabular}{@{}p{15.0cm}@{}}
    % 最近老是胃不舒服，不知道是不是吃坏了。\\
    % Recently, I always feel sick in my stomach. I don't know if it's bad.\\ \vspace{-0.3em} 
    % 具体是什么症状呢？胃痛还是胃胀，建议胃不舒服还是要看一下医生。\\
    % What are the symptoms? Gastralgia or bloating. It is recommended to see a doctor if you feel sick.\\ \vspace{-0.3em} 
    经常胃痛，同事建议我去做个胃镜好好检查一下，可是胃镜是不是很难受呀？\\
    I often have a stomachache. My colleague suggested that I go to have a gastroscopy examination. But is the gastroscopy very uncomfortable?\\ \vspace{-0.3em} 
    胃镜分很多种呢，你可以考虑下无痛的体外胃肠影像扫描仪，是利用超声原理进行检查的，不会有疼痛感。或者也可以考虑做胶囊内镜，吞下它就相当于吞下一个微型的照相机，它会拍摄检查你的肠胃情况，也不会有什么不适感。\\
    There are many kinds of gastroscopy. You can consider the painless external gastrointestinal imaging scanner, which uses the principle of ultrasound to examine without pain. Or you can also consider capsule endoscopy, swallowing it is equivalent to swallowing a miniature camera. It will take pictures to check your stomach and intestines, and there will be no discomfort.\\ \vspace{-0.3em} 
    听上去似乎没那么可怕了，有没有可能我是得了幽门螺旋杆菌呢？身边有不只一个朋友得了这个病。\\
    It doesn't sound so terrible. Is it possible that I have H.pylori? More than one friend of mine has this disease.\\ \vspace{-0.3em} 
    除了胃痛，你看看你有没有这些症状，包括反酸、烧心、口臭、恶心、呕吐等等。\\
    Besides a stomachache, check if you have these symptoms, including acid reflux, heartburn, bad breath, nausea, vomiting, etc.\\ \vspace{-0.3em} 
    还真有，我会有反酸、烧心的感觉。幽门螺旋杆菌感染能根治吗？\\
    I really do, I have acid reflux and heartburn. Can H.pylori infection be eradicated?\\ \vspace{-0.3em} 
    可以根治，但是要尽早治疗呀，幽门螺旋杆菌感染和胃癌的发病有关，尽早治疗才能预防胃癌，不然会错过最佳的治疗时机。\\
    It can be eradicated, but it should be treated as soon as possible.  H.pylori infection is related to the incidence of gastric cancer. Early treatment can prevent gastric cancer, or else the best treatment opportunity will be missed.\\ \vspace{-0.3em} 
    \underline{这个}\textbf{(coreference)}感染了怎么治疗呢？\textbf{(how question type)} \\
    How to treat \underline{this} \textbf{(coreference)} infection? \textbf{(how question type)}
  \end{tabular} &
  \begin{tabular}{@{}p{3.0cm}@{}} 幽门螺旋杆菌感染的治疗方法 \\ Treatment of  H.pylori infection
  \end{tabular} \\ 
  
  \bottomrule
    
\end{tabular}
}
\end{table*}
\end{CJK*}

% type 8
\begin{CJK*}{UTF8}{gbsn}
\begin{table*}[t]
\newcolumntype{?}{!{\vrule width 1pt}}
    \newcolumntype{C}{>{\centering\arraybackslash}p{2em}}
    \caption{
        \label{tb:dusincrtype8} Why question type query generation examples on DuSincR.
    }
    \centering 
    \small
    \renewcommand\arraystretch{1.0}
\scalebox{0.95}{
\begin{tabular}{cp{3.0cm}}

\toprule
Dialogue History &
    \multicolumn{1}{c}{Generated Query} \\
    \midrule
    
  \begin{tabular}{@{}p{15.0cm}@{}}
    为什么人会过敏啊，我真的受够这种每年春天都疯狂打喷嚏的体质了。\\
    Why can people be allergic? I'm really fed up with this kind of body that sneezes crazily every spring. \\ \vspace{-0.3em} 
    你指的过敏性鼻炎吧，这些归根结底都是免疫系统产生了过量的免疫蛋白，导致对待一些物质过分敏感，机体产生反应。一般是基因和环境共同作用结果。\\
    You mean allergic rhinitis. This is because the immune system produces excessive immune protein, which leads to over-sensitivity to some substances, and the body reacts. Generally, it is due to the result of the joint action of genes and environment. \\ \vspace{-0.3em} 
    啊基因问题是不是就没办法治了，好烦啊。\\
    Ah, if it's a genetic problem, isn't it very rare to cure? It's so annoying. \\ \vspace{-0.3em} 
    遗传性过敏目前还不能根治的，不过你这种一般不会很严重，到季节就留意一下避开过敏源吧。\\
    Hereditary allergy cannot be eradicated at present, but your kind is usually not very serious. Be sure to avoid allergens when the season comes. \\ \vspace{-0.3em} 
    也只能这样了。一不留神就会中招，为什么近几年\underline{这个}\textbf{(coreference)}发作得少了呢？\textbf{(why question type)}\\
    That's the only way. If you are not careful enough, you will be suffering. Why \underline{this} \textbf{(coreference)} attacks less frequent in recent years? \textbf{(why question type)}
  \end{tabular} &
  \begin{tabular}{@{}p{3.0cm}@{}} 过敏性鼻炎为什么发病率下降 \\ Why does the incidence rate of allergic rhinitis decrease
  \end{tabular} \\

  \cmidrule{1-2}
  \begin{tabular}{@{}p{15.0cm}@{}}
    这么说领养真是一个不错的选择，可我到哪儿去找到等着被领养的小猫呢？\\
    So adoption is a good choice, but where can I find the kitten waiting to be adopted?\\ \vspace{-0.3em}
    可以去漳州小动物科学养护救助中心看一看，这是一家比较靠谱的民间机构，提供免费的救助和领养服务。希望你能在那里找到和你有缘的小猫咪～\\
    You can go to the Zhangzhou Small Animal Science Conservation and Rescue Center, which is a reliable non-governmental organization providing free rescue and adoption services. I hope you can find the kitten that you are destined for there. \\ \vspace{-0.3em}
    多问一句，为什么领养要给\underline{它们}\textbf{(coreference)}绝育呢？\textbf{(why question type)}\\
    By the way, why should \underline{they} \textbf{(coreference)} be sterilized for adoption? \textbf{(why question type)}
  \end{tabular} &
  \begin{tabular}{@{}p{3.0cm}@{}} 宠物猫为什么要绝育 \\ Why should pet cats be sterilized
  \end{tabular} \\

  \cmidrule{1-2}
  \begin{tabular}{@{}p{15.0cm}@{}}
    我最近身体不太舒服，导致经期都不调了。\\
    I'm not feeling well recently, which leads to irregular menstruation.\\ \vspace{-0.3em}
    经期不调好像是当代年轻人的常见病了。\\
    Irregular menstruation seems to be a common illness among young people nowadays. \\ \vspace{-0.3em}
    哦，是这样的，我打算去看看中医。\\
    Oh, yes you're right. I'm going to see a traditional Chinese medicine doctor.\\ \vspace{-0.3em}
    医生也许会给你开一些加味逍遥丸，乌鸡白凤丸。\\
    The doctor may prescribe some modified Xiaoyao pills, Wuji Baifeng pills. \\ \vspace{-0.3em}
    啊哈，为什么现在很多人年轻人都会得\underline{这种}\textbf{(coreference)}病症啊？\textbf{(why question type)} \\
    Aha, why are many young people suffering from \underline{this} \textbf{(coreference)} illness now? \textbf{(why question type)}
  \end{tabular} &
  \begin{tabular}{@{}p{3.0cm}@{}} 年轻人为什么得经期不调 \\ Why young people have irregular menstruation
  \end{tabular} \\ 
  
  \bottomrule
    
\end{tabular}
}
\end{table*}
\end{CJK*}

% Score range

% 0.5-0.6
\begin{CJK*}{UTF8}{gbsn}
\begin{table*}[t]
\newcolumntype{?}{!{\vrule width 1pt}}
	\newcolumntype{C}{>{\centering\arraybackslash}p{2em}}
	\caption{
		\label{tb:WebKnow5} Search result examples with score range 0.5-0.6 on DuSinc, accounts for 0.2\% of the total dataset.
	}
	\centering 
	\small
	\renewcommand\arraystretch{1.0}
\scalebox{0.95}{
\begin{tabular}{cp{7.5cm}p{7.5cm}}

\toprule
Search Query &
  \multicolumn{1}{c}{Ground-truth Knowledge Snippet} &
  \multicolumn{1}{c}{Retrieved Knowledge Snippet} \\ 
  \midrule

\begin{tabular}{@{}p{3.0cm}@{}}
   鲁迅写的祥林嫂出自哪一篇 \\ Which article did Lu Xun write about Sister Xianglin
  \end{tabular} &
  \begin{tabular}{@{}p{7.5cm}@{}} 祝福。 \\ Blessing.
  \end{tabular} &
  \begin{tabular}{@{}p{7.5cm}@{}} 祥林嫂（鲁迅短篇小说《祝福》中的人物）：祥林嫂是鲁迅创作的短篇小说《祝福》中的人物。她经历了两次丧夫，一次丧子，在新春祭拜的欢乐气氛中，在肉体和精神的双重折磨下悲惨地死去。祥林嫂是当时封建社会制度摧残的产物，是封建伦理道德绞杀的结果。作为封建社会背景下悲剧命运的典型人物，鲁迅笔下的祥林嫂悲惨的一生真实地反映了黑暗的封建时代给普通劳苦大众带来的沉重影响，这是个人的不幸，更是社会的悲哀。 \\ Sister Xianglin (a character in Lu Xun's short story \emph{Blessing}): Sister Xianglin is a character in Lu Xun's short story \emph{Blessing}. She had lost her husband twice and her son once. In the happy atmosphere of the New Year worship, she died tragically under the double torture of body and spirit. Xianglin sister-in-law was the product of the destruction of the feudal social system and the result of the strangulation of feudal ethics. As a typical figure of tragic fate under the background of feudal society, Lu Xun's tragic life of Xianglin Sister-in-law truly reflects the heavy impact of the dark feudal era on the ordinary working people, which is not only personal misfortune, but also social sorrow.
  \end{tabular} \\

  \cmidrule{1-3}

\begin{tabular}{@{}p{3.0cm}@{}}
   武则天是不是中国第一个女皇 \\ Is Wu Zetian the first queen of China 
  \end{tabular} &
  \begin{tabular}{@{}p{7.5cm}@{}} 是。 \\ Yes.
  \end{tabular} &
  \begin{tabular}{@{}p{7.5cm}@{}} 武则天是中国历史上唯一的女皇帝吗：武则天，即唐高宗李治皇后，唐至武周时期的政治家、武周开国皇帝，也是中国历史上唯一一位正统女皇帝。不仅如此，她还是即位年龄最大、在位时间最长的皇帝之一。史学界一直认为她是中国历史上唯一的女皇帝，但也有人认为，北魏孝明帝的女儿元姑娘是最早的女帝。 \\ Is Wu Zetian the only female emperor in Chinese history: Wu Zetian, the Empress Li Zhi of Emperor Gaozong of the Tang Dynasty, the politician from the Tang Dynasty to the Wu Zhou Dynasty, the founding emperor of the Wu Zhou Dynasty, is also the only orthodox female emperor in Chinese history. Not only that, she was also one of the oldest and longest emperors. Historians have always believed that she was the only female emperor in China's history, but some people also believed that the daughter of Emperor Xiaoming of the Northern Wei Dynasty, Miss Yuan, was the earliest female emperor.
  \end{tabular} \\

    \cmidrule{1-3}

\begin{tabular}{@{}p{3.0cm}@{}}
   新农贸市场的菜贵吗 \\ Are the vegetables at the New Farmers Market expensive
  \end{tabular} &
  \begin{tabular}{@{}p{7.5cm}@{}} 有点贵。 \\ It's a little expensive.
  \end{tabular} &
  \begin{tabular}{@{}p{7.5cm}@{}} 2021今日蔬菜价格表，2021今日蔬菜价格表大全?-RUIKE百科：2022年9月26日北京市丰台区新发地农产品批发市场价格10.5元/公斤，天津红旗农贸综合批发市场价格10.6元/公斤，石家庄桥西蔬菜中心批发市场价格7.5元/公斤。 \\ 2021 Vegetable Price List Today, 2021 Vegetable Price List Today - RUIKE Encyclopedia: On September 26, 2022, the wholesale market price of Xinfadi agricultural products in Fengtai District, Beijing was 10.5 RMB/kg, the comprehensive wholesale market price of Tianjin Hongqi Agricultural Trade was 10.6 RMB/kg, the wholesale market price of Shijiazhuang Qiaoxi Vegetable Center was 7.5 RMB/kg.
  \end{tabular} \\
  
  \bottomrule
  
\end{tabular}
}
 \begin{tablenotes}
    \footnotesize
    \item[] These examples, which only make up 0.2\% of the entire 9,353 examples and have scores in range of 0.5 to 0.6, acquire high quality search results despite their low scores. The reason why the score is low, is mainly due to some of the given knowledge provided in DuSinc is a brief answer to the search query, while our web knowledge retrieved is a complete paragraph. This result works the same with the examples which falls into the score range of 0.6 to 0.7 presented in Table~\ref{tb:WebKnow6}.
  \end{tablenotes}
\end{table*}
\end{CJK*}

% 0.6-0.7

\begin{CJK*}{UTF8}{gbsn}
\begin{table*}[t]
\newcolumntype{?}{!{\vrule width 1pt}}
	\newcolumntype{C}{>{\centering\arraybackslash}p{2em}}
	\caption{
		\label{tb:WebKnow6} Search result examples with score range 0.6-0.7 on DuSinc, accounts for 1.9\% of the total dataset.
	}
	\centering 
	\small
	\renewcommand\arraystretch{1.0}
\scalebox{0.95}{
\begin{tabular}{cp{7.5cm}p{7.5cm}}

\toprule
Search Query &
  \multicolumn{1}{c}{Ground-truth Knowledge Snippet} &
  \multicolumn{1}{c}{Retrieved Knowledge Snippet} \\ 
  \midrule

\begin{tabular}{@{}p{3.0cm}@{}}
   % 酸捻是用什么做的 \\ What is acid twister made of
   快递是哪个国家发明的 \\ Which country invented express delivery
  \end{tabular} &
  \begin{tabular}{@{}p{7.5cm}@{}} 
  % 捻酸，拼音niǎnsuān，汉语词语，吃醋，嫉妒。 \\ Twist acid, pinyin niǎn suān. Chinese words, jealous, jealous.
  美国。 \\ The United States of America.
  \end{tabular} &
  \begin{tabular}{@{}p{7.5cm}@{}} 
  % 「酸三稔」酸三稔的功效与作用-青瓜网:2022年6月27日酸三稔干是一种可以直接食用的干果,也是一种营养丰富的保健食材,更是一种功效出色的中药材,它是以野生水果山捻子为主要原料经晾晒加工后得到的。 \\ The efficacy and function of "acid samura" - Qinggua.com: On June 27, 2022, the dry fruit of "acid samura" is a kind of dry fruit that can be eaten directly, is also a kind of nutritious health food, and is also a kind of Chinese medicinal material with excellent efficacy. It is obtained by drying and processing the wild fruit twigs as the main raw material.
  快递是哪个国家发明：美国。最早的快递应该是联邦快递（FedEx）。1971年创建，目前向211个国家及地区提供快递运输服务。FedEx（联邦快递）全球最大的快递公司，国际快递行业的鼻祖。 \\ Which country invented express delivery: the United States of America. The earliest express should be FedEx. Founded in 1971, it currently provides express transportation services to 211 countries and regions. FedEx is the largest express company in the world and the ancestor of the international express industry.
  \end{tabular} \\

  \cmidrule{1-3}

\begin{tabular}{@{}p{3.0cm}@{}}
   桂林和河池的距离 \\ Distance between Guilin and Hechi
  \end{tabular} &
  \begin{tabular}{@{}p{7.5cm}@{}} 318公里。 \\ 318 km.
  \end{tabular} &
  \begin{tabular}{@{}p{7.5cm}@{}} 开车从桂林到河池多少公里-时间要多久-高速怎么走...：桂林->河池\ 总距离：313.45(公里)\ 总耗时：4.6(小时)\ 油费：188(元)\ 路桥费：145(元)\ 总费用：333(元)。 \\ How many kilometers from Guilin to Hechi by car? How long will it take? How to get to the expressway...: Total distance from Guilin to Hechi: 313.45(km) Total time: 4.6(h) Oil fee: 188(RMB) Road and bridge fee: 145(RMB) Total cost: 333(RMB).
  \end{tabular} \\

    \cmidrule{1-3}

\begin{tabular}{@{}p{3.0cm}@{}}
   家庭和事业哪个更重要 \\ Which is more important, family or career
  \end{tabular} &
  \begin{tabular}{@{}p{7.5cm}@{}} 都重要。 \\ Both important.
  \end{tabular} &
  \begin{tabular}{@{}p{7.5cm}@{}} 家庭和事业到底哪个为重：虽然说家庭和事业同等重要，但首先要选择家庭。用心把家经营好了，再去发展事业。 \\ Which is more important, family or career? Although family and career are equally important, we should choose family first. After you have managed your family well, you can develop your career.
  \end{tabular} \\
  
  \bottomrule
  
\end{tabular}
}
 % \begin{tablenotes}
 %    \footnotesize
 %    \item[] These examples, which make up 1.9\% of the entire 9,353 examples and have scores in range of 0.6 to 0.7, 
 %  \end{tablenotes}
\end{table*}
\end{CJK*}

% 0.7-0.8

\begin{CJK*}{UTF8}{gbsn}
\begin{table*}[t]
\newcolumntype{?}{!{\vrule width 1pt}}
	\newcolumntype{C}{>{\centering\arraybackslash}p{2em}}
	\caption{
		\label{tb:WebKnow7} Search result examples with score range 0.7-0.8 on DuSinc, accounts for 12.4\% of the total dataset.
	}
	\centering 
	\small
	\renewcommand\arraystretch{1.0}
\scalebox{0.95}{
\begin{tabular}{cp{7.5cm}p{7.5cm}}

\toprule
Search Query &
  \multicolumn{1}{c}{Ground-truth Knowledge Snippet} &
  \multicolumn{1}{c}{Retrieved Knowledge Snippet} \\ 
  \midrule

\begin{tabular}{@{}p{3.0cm}@{}}
   % 陈奕迅演唱会 \\ Eason Chan Concert
   甄嬛huán传的剧情是否真实 \\ Whether the story of the Legend of Zhen Huan is true
  \end{tabular} &
  \begin{tabular}{@{}p{7.5cm}@{}} 
  % 应该是陈奕迅2007北京演唱会\ http://v.youku.com/v\_show/id\_XMTYzOTQ2NDAw.html\ 我是EF。希望对你有帮助。 \\ It should be Eason Chan's 2007 Beijing Concert. http://v.youku.com/v\_show/id\_XMTYzOTQ2NDAw.html I'm EF. I hope it will help you.
  不符合历史。 \\ Non-conformance with history.
  \end{tabular} &
  \begin{tabular}{@{}p{7.5cm}@{}} 
  % 陈奕迅演唱会座无虚席,连唱30首歌冷门曲偏多,甘比舒淇...:原创2022年12月10日陈奕迅原本在三年前就该举办的演唱会《陈奕迅FearAndDreams》终于在香港红馆正式上演。 \\ Eason Chan's concert was packed, with 30 songs sung in a row. There were more popular songs than Shu Qi Original: On December 10, 2022, Eason Chan's concert "Eason Chan FearAndDreams", which was supposed to be held three years ago, was finally officially staged in the Hong Kong Red Mansion.
  《甄嬛huán传》是真实历史吗：《甄嬛huán传》不是真实历史的。甄嬛huán传不是历史事实，甄嬛huán传的电视剧是根据流潋紫的历史架空小说改编，历史架空的意思是历史上没有这样的朝代，完全是小说作者创造的一个世界。\\ Is "Legend of Zhen Huan" true history? "Legend of Zhen Huan" is not true history. The legend of Zhen Huan is not a historical fact. The TV series of the legend of Zhen Huan is based on the colorful historical overhead novel. The historical overhead means that there is no such dynasty in history, and it is a world created by the novel author.
  \end{tabular} \\

    \cmidrule{1-3}

\begin{tabular}{@{}p{3.0cm}@{}}
   % 星座的由来 \\ Origin of constellation
   鹤岗市在黑龙江省地理位置 \\ The geographical location of Hegang City in Heilongjiang Province
  \end{tabular} &
  \begin{tabular}{@{}p{7.5cm}@{}} 
  % 星座的起源是米索不达米亚文明，古巴比伦时期 \\ The origin of constellations is Mesopotamian civilization, the Babylonian period
  黑龙江省东北部。 \\ Northeast of Heilongjiang Province.
  \end{tabular} &
  \begin{tabular}{@{}p{7.5cm}@{}} 
  % 你真的了解星座吗?-知乎:2021年8月28日把它们想象成动物或人物的形象,结合神话故事给它们起出适当的名字,这就是星座名称的由来。 \\ Do you really know the constellation- Zhihu: Imagine them as animals or characters on August 28, 2021, and give them proper names in combination with fairy tales, which is the origin of constellation names.
  鹤岗：鹤岗市位于黑龙江省东北部，地处小兴安岭东麓低山丘陵地区及松花江、黑龙江汇合处的平原地区。地理坐标为东经129°39'50"—132°31'00"，北纬47°03'30"—48°21'00"。东至松花江与同江市一水相连，西邻伊春市，南与佳木斯市汤原县接壤，北部以黑龙江主航道为界与俄罗斯隔江相望。\\ Hegang: Hegang City is located in the northeast of Heilongjiang Province, in the low mountains and hills at the east foot of the Xiaoxing'an Mountains and the plain where the Songhua River and Heilongjiang meet. The geographical coordinates are 129°39'50"-132°31'00"E and 47°03'30"-48°21'00"N. The Songhua River is connected with Tongjiang City in the east, Yichun City in the west, Tangyuan County in Jiamusi City in the south, and the main channel of Heilongjiang Province in the north.
  \end{tabular} \\

  \cmidrule{1-3}
  
\begin{tabular}{@{}p{3.0cm}@{}}
   switch价格 \\ Price of switch
  \end{tabular} &
  \begin{tabular}{@{}p{7.5cm}@{}} switch官方定价是299.99美元/29980日元，switch lite的官方定价是199.99美元/19980日元。 \\ The official price of switch is USD 299.99/JPY 29980, and the official price of switch lite is USD 199.99/JPY 9980.
  \end{tabular} &
  \begin{tabular}{@{}p{7.5cm}@{}} 2022年超详细的任天堂Switch购买指南-知乎：2022年8月15日Switch Lite的售价是1500元左右，相比原版Switch便宜了700到1000元。 \\ Super detailed Nintendo Switch purchase guide in 2022 - Zhihu: The price of Switch Lite on August 15, 2022 is about 1500 RMB, which is 700 to 1000 RMB cheaper than the original switch.
  \end{tabular} \\

  \bottomrule
  
\end{tabular}
}
 % \begin{tablenotes}
 %    \footnotesize
 %    \item[] These examples, which make up 12.4\% of the entire 9,353 examples and have scores in range of 0.7 to 0.8, 
 %  \end{tablenotes}
\end{table*}
\end{CJK*}

% 0.8-0.9

\begin{CJK*}{UTF8}{gbsn}
\begin{table*}[t]
\newcolumntype{?}{!{\vrule width 1pt}}
	\newcolumntype{C}{>{\centering\arraybackslash}p{2em}}
	\caption{
		\label{tb:WebKnow8} Search result examples with score range 0.8-0.9 on DuSinc, accounts for 56.5\% of the total dataset.
	}
	\centering 
	\small
	\renewcommand\arraystretch{1.0}
\scalebox{0.95}{
\begin{tabular}{cp{7.5cm}p{7.5cm}}

\toprule
Search Query &
  \multicolumn{1}{c}{Ground-truth Knowledge Snippet} &
  \multicolumn{1}{c}{Retrieved Knowledge Snippet} \\ 
  \midrule

\begin{tabular}{@{}p{3.0cm}@{}}
   当服装设计师难不难 \\ Is it difficult to be a fashion designer
  \end{tabular} &
  \begin{tabular}{@{}p{7.5cm}@{}} 以下结果来自艺考网：服装设计并不难学，服装设计入门是非常简单的，但是要想成为一名优秀的服装设计师，不仅要有专业特长，还要有丰富的设计经验，所以只有不断努力，才能成为一名优秀的服装设计师。 \\ The following results come from Yikao.com: Fashion design is not difficult to learn, and the introduction of fashion design is very simple, but to become an excellent fashion designer, you need not only professional expertise, but also rich design experience, so only by continuous efforts can you become an excellent fashion designer.
  \end{tabular} &
  \begin{tabular}{@{}p{7.5cm}@{}} 服装设计难吗-百度知了好学：2021年12月28日做服装外表看起来很简单谁，都可以来做，其实现在做什么事情都很难，只有你身在其中的时候才会感觉竞争如此激烈，你自己要是没有一个好的营销策略。 \\ Is it difficult to design clothes? Baidu Zhizhi is eager to learn: on December 28, 2021, it looks very easy to make clothes. Anyone can do it. In fact, it is difficult to do anything now. Only when you are in it will you feel the competition is so fierce. If you don't have a good marketing strategy yourself.
  \end{tabular} \\

  \cmidrule{1-3}

\begin{tabular}{@{}p{3.0cm}@{}}
   幽门螺旋杆菌检查 \\ Helicobacter pylori test
  \end{tabular} &
  \begin{tabular}{@{}p{7.5cm}@{}} 聂里红副主任医师：京中医药大学东直门医院\ 幽门螺旋杆菌是一种微需氧，螺旋形的杆菌，它是迄今为止发现的，在胃内唯一能够存活的细菌。这个细菌对生存环境要求挺苛刻的，在完全的有氧状态，或者是完全无氧状态，它都不能生存。这个细菌是1983年，由国外的两位科学家发现的，研究发现，它与我们胃中的慢性萎缩性胃炎、消化性溃疡、胃的恶性肿瘤，以及胃外的一些疾病，如淋巴瘤、缺铁性贫血、特发性血小板减少性紫癜，都有相关性。所以目前大家挺重视的，在病人中间普查，也是比较普遍的一件事情。 \\ Nie Lihong, Deputy Chief Physician: Dongzhimen Hospital of Beijing University of Traditional Chinese Medicine. Helicobacter pylori is a microaerobic, helicoid bacterium, which is the only bacteria found so far that can survive in the stomach. This bacterium has strict requirements for its living environment. It cannot survive in a completely aerobic state or a completely anaerobic state. This bacterium was discovered by two foreign scientists in 1983. The research found that it is related to chronic atrophic gastritis, peptic ulcer, gastric malignant tumor in our stomach, as well as some extragastric diseases, such as lymphoma, iron deficiency anemia, idiopathic thrombocytopenic purpura. So at present, we attach great importance to the fact that the general survey among patients is also a relatively common thing.
  \end{tabular} &
  \begin{tabular}{@{}p{7.5cm}@{}} 怎样检查幽门螺杆菌-有来医生：1.有创检查：幽门螺旋杆菌的有创检查包括抽血化验检查幽门螺杆菌的抗体，还有胃镜下的一种快速尿素酶，取一块黏膜组织，进行显微镜下涂片检测，快速尿素酶和胃镜\ 2.无创检测：可进行碳13、碳14呼气试验，该检测方法可以无创、快速的得到结果。 \\ How to check Helicobacter pylori - Youlai doctor: 1. Invasive examination: The invasive examination of Helicobacter pylori includes blood sampling test to check the antibody of Helicobacter pylori, and a rapid urease under the gastroscope. Take a piece of mucosal tissue for microscopic smear detection, rapid urease and gastroscopy. 2. Non-invasive detection: carbon 13 and carbon 14 breath test can be carried out. This detection method can obtain results non-invasively and quickly.
  \end{tabular} \\

    \cmidrule{1-3}

\begin{tabular}{@{}p{3.0cm}@{}}
   秦始皇万里长城 \\ The Great Wall of Qin Shihuang
  \end{tabular} &
  \begin{tabular}{@{}p{7.5cm}@{}} 万里长城并不是秦始皇修建的，秦始皇不过是顺便把它们连在一起提起万里长城，我们最先想到的就是秦始皇。秦始皇修建了万里长城，这是他众多的功绩之一，这件事所有人都认同。不过事实上并不是这样，秦始皇虽然有很多功绩，被称为千古一帝，但万里长城并不是他修建的。秦始皇只不过顺便把万里长城顺联连在了一起，这些东西本来就在。 \\ The Great Wall was not built by Qin Shihuang. Qin Shihuang just mentioned the Great Wall by connecting them together. The first thing we thought of was Qin Shihuang. Qin Shihuang built the Great Wall, which is one of his many achievements, which everyone agrees with. However, in fact, this is not the case. Although Emperor Qin Shihuang has many achievements and is known as the Emperor for ever, he did not build the Great Wall. Qin Shihuang just connected the Great Wall together by the way. These things are already there.
  \end{tabular} &
  \begin{tabular}{@{}p{7.5cm}@{}} 秦长城（秦始皇所筑长城）：秦始皇三十三年（公元前214年）遣大将蒙恬北逐匈奴，筑长城万余里，以防匈奴南进，史称秦长城。秦长城实际是在原先战国时期秦长城，赵长城，燕长城三国长城的基础上修建。西起临洮（今甘肃岷县）、东至鸭绿江（今辽宁省的东部和南部及吉林省的东南部地区）共筑万余里，故史称：“万里长城”。被列为国家级重点文物保护单位的秦长城。 \\ Qin Great Wall (the Great Wall built by Qin Shihuang): In the 33rd year of Qin Shihuang (214 BC), the Great General Meng Tian was sent to drive the Huns to the north and build the Great Wall for more than ten thousand miles to prevent the Huns from going south. It is called the Qin Great Wall in history. The Qin Great Wall was actually built on the basis of the Qin Great Wall, Zhao Great Wall and Yan Great Wall in the Warring States Period. From Lintao (now Minxian County, Gansu Province) in the west to Yalu River (now the east and south of Liaoning Province and the southeast of Jilin Province) in the east, more than ten thousand li have been built, so it is called the "Great Wall" in history. The Great Wall of Qin is listed as a national key cultural relics protection unit.
  \end{tabular} \\
  
  \bottomrule
  
\end{tabular}
}
 % \begin{tablenotes}
 %    \footnotesize
 %    \item[] These examples, which make up 56.5\% of the entire 9,353 examples and have scores in range of 0.8 to 0.9, 
 %  \end{tablenotes}
\end{table*}
\end{CJK*}

% 0.9-1.0

\begin{CJK*}{UTF8}{gbsn}
\begin{table*}[t]
\newcolumntype{?}{!{\vrule width 1pt}}
	\newcolumntype{C}{>{\centering\arraybackslash}p{2em}}
	\caption{
		\label{tb:WebKnow9} Search result examples with score range 0.9-1.0 on DuSinc, accounts for 29.0\% of the total dataset.
	}
	\centering 
	\small
	\renewcommand\arraystretch{1.0}
\scalebox{0.95}{
\begin{tabular}{cp{7.5cm}p{7.5cm}}

\toprule
Search Query &
  \multicolumn{1}{c}{Ground-truth Knowledge Snippet} &
  \multicolumn{1}{c}{Retrieved Knowledge Snippet} \\ 
  \midrule

\begin{tabular}{@{}p{3.0cm}@{}}
   上方山森林公园 \\ Shangfangshan Forest Park
  \end{tabular} &
  \begin{tabular}{@{}p{7.5cm}@{}} 江苏上方山国家森林公园位于国家级太湖风景区石湖景区内，距苏州市中心4公里，以吴越遗迹和江南水乡田园风光取胜的天然公园。山水相依，风景如画，胜迹遍地，被誉为“吴中胜境”，是太湖风景名胜区中一颗璀璨的明珠。 \\ Jiangsu Shangfangshan National Forest Park is located in the Stone Lake Scenic Spot of the National Taihu Lake Scenic Spot, 4 kilometers away from the center of Suzhou. It is a natural park that is famous for the relics of Wu and Yue and the idyllic scenery of Jiangnan Watertown. With picturesque landscapes and scenic spots, it is known as the "Wuzhong Scenic Spot" and is a bright pearl in the Taihu Lake Scenic Spot.
  \end{tabular} &
  \begin{tabular}{@{}p{7.5cm}@{}} 上方山国家森林公园（国家AAA级景区）：苏州上方山国家森林公园位于国家级太湖风景区石湖景区内，距苏州市中心4公里，以吴越遗迹和江南水乡田园风光取胜的天然公园。山水相依，风景如画，胜迹遍地，被誉为“吴中胜境”，是太湖风景名胜区中一颗璀璨的明珠。1992年7月由林业部批准为国家级森林公园。 \\ Shangfangshan National Forest Park (National AAA Scenic Spot): Suzhou Shangfangshan National Forest Park is located in the Stone Lake Scenic Spot of the National Taihu Lake Scenic Spot, which is 4 kilometers away from the center of Suzhou. It is a natural park that is famous for the relics of Wu and Yue and the idyllic scenery of Jiangnan Watertown. With picturesque landscapes and scenic spots, it is known as the "Wuzhong Scenic Spot" and is a bright pearl in the Taihu Lake Scenic Spot. In July 1992, it was approved as a national forest park by the Ministry of Forestry.
  \end{tabular} \\

  \cmidrule{1-3}

\begin{tabular}{@{}p{3.0cm}@{}}
   流浪地球 \\ The Wandering Earth
  \end{tabular} &
  \begin{tabular}{@{}p{7.5cm}@{}} 《流浪地球》（英文：The Wandering Earth）是由中国电影股份有限公司、北京京西文化旅游股份有限公司等公司联合出品，由郭帆执导，屈楚萧、赵今麦、李光洁、吴孟达等领衔主演的科幻灾难片，吴京特别出演。该片于2019年2月5日在中国内地上映。影片根据刘慈欣的同名小说改编，故事背景设定在2075年，讲述了太阳即将毁灭，已经不适合人类生存，而面对绝境，人类将开启“流浪地球”计划，试图带着地球一起逃离太阳系，寻找人类新家园的故事。截至2021年10月17日，《流浪地球》为中国影史票房榜第五名。 \\ The Wandering Earth (English: The Wandering Earth) is a sci-fi disaster film jointly produced by China Film Co., Ltd., Beijing Jingxi Culture and Tourism Co., Ltd., directed by Guo Fan, and starred by Qu Chuxiao, Zhao Jinmai, Li Guangjie, Wu Mengda, etc., with Wu Jing specially starring. The film was released in mainland China on February 5, 2019. The film is based on Liu Cixin's novel of the same name, and the story background is set in 2075. It tells the story that the sun is about to be destroyed and is no longer suitable for human survival. In the face of desperate circumstances, human beings will start the "wandering earth" plan, trying to escape the solar system with the earth and find a new home for human beings. As of October 17, 2021, "Wandering Earth" ranked fifth in the box office list of Chinese film history.
  \end{tabular} &
  \begin{tabular}{@{}p{7.5cm}@{}} 流浪地球:\ 类型：电影作品\ 导演：郭帆\ 简介：《流浪地球》是由郭帆执导，吴京特别出演、屈楚萧、赵今麦、李光洁、吴孟达等领衔主演的科幻冒险电影。影片根据刘慈欣的同名小说改编，故事背景设定在2075年，讲述了太阳即将毁灭，已经不适合人类生存，而面对绝境，人类将开启“流浪地球”计划，试图带着地球一起逃离太阳系，寻找人类新家园的故事。该片于2019年2月5日在中国内地上映。\\ The Wandering Earth: type: film director: Guo Fan Profile: Wandering Earth is a sci-fi adventure film directed by Guo Fan, starring Qu Chuxiao, Zhao Jinmai, Li Guangjie, Wu Mengda and other actors. The film is based on Liu Cixin's novel of the same name, and the story background is set in 2075. It tells the story that the sun is about to be destroyed and is no longer suitable for human survival. In the face of desperate circumstances, human beings will start the "wandering earth" plan, trying to escape the solar system with the earth and find a new home for human beings. The film was released in mainland China on February 5, 2019.
  \end{tabular} \\

    \cmidrule{1-3}

\begin{tabular}{@{}p{3.0cm}@{}}
   温带性海洋性季风气候 \\ Temperate oceanic monsoon climate
  \end{tabular} &
  \begin{tabular}{@{}p{7.5cm}@{}} 温带海洋性季风气候是以季风气候为主、根据温带海洋性气候和温带季风气候的特征而形成的局部气候，具有海洋性和季风性两种因素的共同影响。分布在北温带季风气候区的沿海地带。\\ Temperate oceanic monsoon climate is a local climate formed mainly by the monsoon climate and based on the characteristics of the temperate marine climate and the temperate monsoon climate. It has the common influence of both marine and monsoon factors. It is distributed in the coastal zone of the north temperate monsoon climate zone.
  \end{tabular} &
  \begin{tabular}{@{}p{7.5cm}@{}} 温带海洋性季风气候：温带海洋性季风气候（Temperate oceanic monsoon climate）是以季风气候为主、根据温带海洋性气候和温带季风气候的特征而形成的局部气候，具有海洋性和季风性两种因素的共同影响。分布在北温带季风气候区的沿海地带。是温带地区最适合人类居住的气候。我国只有青岛、大连、烟台、威海四个城市属于这个气候类型。 \\ Temperate oceanic monsoon climate: Temperate oceanic monsoon climate is a local climate formed mainly by monsoon climate and based on the characteristics of temperate oceanic climate and temperate monsoon climate, with the joint influence of both oceanic and monsoon factors. It is distributed in the coastal zone of the north temperate monsoon climate zone. It is the most suitable climate for human habitation in the temperate zone. Only Qingdao, Dalian, Yantai and Weihai in China belong to this climate type.
  \end{tabular} \\
  
  \bottomrule
  
\end{tabular}
}
 % \begin{tablenotes}
 %    \footnotesize
 %    \item[] These examples, which make up 29.0\% of the entire 9,353 examples and have scores in range of 0.9 to 1.0, 
 %  \end{tablenotes}
\end{table*}
\end{CJK*}

\begin{CJK*}{UTF8}{gbsn}
\begin{table*}[t]
\newcolumntype{?}{!{\vrule width 1pt}}
	\newcolumntype{C}{>{\centering\arraybackslash}p{2em}}
	\caption{
		\label{tb:CaseStudyBad}  Response generation examples  with noisy knowledge snippet injection.
	}
	\centering 
	\small
	\renewcommand\arraystretch{1.0}
 
\scalebox{0.95}{
\begin{tabular}{cp{5.5cm}p{3.0cm}}

\toprule
Dialogue History &
  \multicolumn{1}{c}{Retrieved Knowledge Snippet} &
  \multicolumn{1}{c}{Generated Response} \\ \midrule

\begin{tabular}{@{}p{9cm}@{}}
    你喜欢看动漫吗?\\
    Do you like watching anime?\\ \vspace{-0.3em}
    喜欢呀,很多日本动漫超级好看\\
    I like it, a lot of Japanese anime are super good-looking\\ \vspace{-0.3em}
    对的,尤其《火影忍者》\\
    Yes, especially Naruto\\ \vspace{-0.3em}
    这部动漫的主角是鸣人,主要描述了鸣人与其伙伴的经历\\
    The protagonist of this anime is Naruto, and it mainly describes the experience of Naruto and his partners\\ \vspace{-0.3em}
    你有没有觉得他们之间的友情太感人了?\\
    Did you find the friendship between them touching?\\ \vspace{-0.3em}
    嗯嗯,他们的羁绊太深了,他们之间的友谊很美妙,友情不要求什么,但是它是温暖的\\
    Well, their bond is strong, their friendship is wonderful, friendship asks for anything, but it comforts you.\\ \vspace{-0.3em}
    除了《火影忍者》,你还知道多少好看的动漫\\
    In addition to "Naruto", what are some good anime that you know
  \end{tabular} &
  \begin{tabular}{@{}p{5.5cm}@{}} 火影忍者好看吗?:火影忍者是特别好看的,因为他的剧情对中国这边来说稍微有亲切感,而且鸣人的设定符合大多数人的内心,都是不被人认可从而想去被别人认可的,所以是非常好看的,而且作者把它刻画的非常的细腻,可以有一种身临其境的感觉的。 非常非常好看,本人从小学就开始追,到完结,看了三遍。 \\ Is Naruto good?: Naruto is particularly good, because his plot is familiar to China, and Naruto’s setting is in line with most people’s hearts. They are not recognized by others and wanted to be recognized, so it is a great anime, and the author portrays it very delicately, which can give you an immersive feeling. It's very, very good. I have been fallowing it since elementary school, and watched it three times until the end. \end{tabular} &
  \begin{tabular}{@{}p{3.0cm}@{}} 《东京猫猫》《夏目友人帐》 \\ "Tokyo Cat" "Natsume's Book of Friends" \end{tabular} \\

  \cmidrule{1-3}

  \begin{tabular}{@{}p{9cm}@{}}
    唉这爸爸吧,完全不管孩子\\
    Oh, my husband, he doesn't take care of the children at all\\ \vspace{-0.3em}
    抱抱你!真是辛苦了～这就是丧偶式育儿～这样家庭中成长出来的孩子～都会对妈妈不亲～或者是爸爸不亲唉\\
    I'm sorry to hear that! It must of been hard for you~ This is widowed parenting~ Children who grow up in this kind of family~ often aren't close to their mother~ or their father.\\ \vspace{-0.3em}
    心累唉\\
    I feel so exhausted\\ \vspace{-0.3em}
    抱抱～还是有办法的～心态摆正～多和爸爸沟通～相互抚慰,共同努力～\\
    I wish there is something I can do~Perhaps there is still a solution~Put your mind right~Communicate with the father more~Comfort each other and collaborate~\\ \vspace{-0.3em}
    我不会眼睁睁看着这样下去的\\
    I'm not going to let this go on\\ \vspace{-0.3em}
    我相信你可以做的很好的,支持你～\\
    I believe you can do well, I support you~\\ \vspace{-0.3em}
    那在育儿方面,我具体可以怎么做呢?\\
    So what can I do in terms of parenting?\\ \vspace{-0.3em}
    嗯嗯～比如让孩子感觉到你的爱～家长一定要让孩子感受到来自你们的爱!因为父爱母爱对孩子的成长都是不可或缺的～要多花时间去陪伴孩子,从互动中多去了解孩子～\\
    Hmm~ For example, let the children feel your love~ Parents must let the children feel the love coming from you! Because fatherly love and motherly love are indispensable to the growth of children~ Spend more time with the children, from the interaction learn more about children\\ \vspace{-0.3em}
    你说的这些我们平时确实没有注意到,现在开始还来得及吗?\\
    We really haven’t payed attention to what you just said, is it too late to start now?

  \end{tabular} &
  \begin{tabular}{@{}p{5.5cm}@{}} 培养孩子注意力有多重要,你知道吗? - 知乎:培养孩子的注意力,是孩子学习以及做事情能否成功的关键,对于孩子的一生也起到了很重要的作用,它是一个人能够高度集中于某一件事情的能力,是属于一项非常重要的心理素质。 或者会有很多爸爸妈妈问,培养孩子的注意力,有什么好处呢? 这里面的好处就多得是了,它可以帮助孩子提高学习能力,当孩子能够把注意力集中在某一件事情的时候,他们就会主动去探求未知的东西,来寻求解决问题的方法。 \\ How important is it to cultivate children's attention, do you know? - Zhihu: Cultivating children's attention is the key to success in learning and doing things. The ability to be highly concentrated on one thing is a very important psychological quality. Or many parents will ask, what are the benefits of cultivating children's attention? Sometimes, they will take the initiative to explore unknown things to find solutions to problems. \end{tabular} &
  \begin{tabular}{@{}p{3.0cm}@{}} 来得及～现在起多多的陪伴孩子～以及和爸爸沟通～让孩子感受到来自你们的爱～相信你们会磨合的很愉快! \\ It's never too late~ From now on, spend more time with your children~ and communicate with the father~ Let the children feel the love from you~ I believe you will have a great time getting along!
 \end{tabular} \\
\bottomrule

\end{tabular}
}
\end{table*}
\end{CJK*}

\begin{CJK*}{UTF8}{gbsn}
\begin{table*}[t]
\newcolumntype{?}{!{\vrule width 1pt}}
	\newcolumntype{C}{>{\centering\arraybackslash}p{2em}}
	% \caption{
	% 	\label{tb:CaseStudyBad}  Response generation examples  with noisy knowledge snippet injection.
	% }
	\centering 
	\small
	\renewcommand\arraystretch{1.0}
 
\scalebox{0.95}{
\begin{tabular}{cp{5.5cm}p{3.0cm}}

\toprule
Dialogue History &
  \multicolumn{1}{c}{Retrieved Knowledge} &
  \multicolumn{1}{c}{Generated Response} \\ \midrule

  \begin{tabular}{@{}p{9cm}@{}}
    我看你穿搭这么好看,你是一个化妆师吗?\\
    You such a good dress code, are you a makeup artist?\\ \vspace{-0.3em}
    哈哈,有点接近,不过我不是化妆师哦\\
    Ha ha, you're pretty close, but I'm not a makeup artist\\ \vspace{-0.3em}
    那你的职业是什么?我很好奇\\
    What is your occupation then, I am curious
  \end{tabular} &
  \begin{tabular}{@{}p{5.5cm}@{}} 化妆师(演员造型设计人员)\_百度百科:化妆的过程就是创作美丽艺术品的过程,化妆师在具体某次化妆的时候,就是在创 ... 对于学化妆的入门课程来说,化妆师需要掌握的化妆基础是一样的,然后化妆师 ... 随着国家的开放,生产力的提高,人们的生活质量和水平也越来越高,对美的追求 ... 一些毕业的学员,能顺利的进入美容院、剧组、电视台、时尚界和演艺圈, ... 展开 \\ Makeup Artist (Actor Styling Designer)\_Baidu Encyclopedia: The process of makeup is the process of creating beautiful works of art. The basics of makeup that need to be mastered are the same, and then makeup artists... With the opening up of the country and the improvement of productivity, people's quality of life and level are also getting higher and higher. The pursuit of beauty... Some graduates can successfully join beauty salons, film crews, TV stations, fashion and entertainment circles, ... Expand \end{tabular} &
  \begin{tabular}{@{}p{3.0cm}@{}} 我是一名服装搭配师,我自己也穿搭服装 \\ I'm a clothing matcher and I create my own dress code.
 \end{tabular} \\  \cmidrule{1-3}

  \begin{tabular}{@{}p{9cm}@{}}
    你好,你知道诗歌发源于哪个朝代吗?\\
    Hello, do you know which dynasty poetry is originated in?

  \end{tabular} &
  \begin{tabular}{@{}p{5.5cm}@{}} 诗歌(文学体裁)\_百度百科:诗歌,是一种抒情言志的 文学体裁 。. 是用高度凝练的语言,生动形象地表达作者丰富情感,集中反映社会生活并具有一定 节奏 和韵律的 文学体裁 。. 《毛诗·大序》记载:“诗者,志之所之也。. 在心为志,发言为诗”。. 南宋严羽《 沧浪诗话 》云:“诗者,吟咏性情也”。. 只有一种用言语表达的艺术就是诗歌。. 2022年, 中 ... 展开 \\ Poetry (literary genre)\_Baidu Encyclopedia:Poetry is a literary genre that expresses emotion and speech. It is a literary genre in which the author's rich emotions are vividly expressed in highly condensed language, reflecting social life with a certain rhythm and rhyme. It is written in the "Mao Shi - Great Preface": "Poetry is also the place of the will. In the heart is the will, and in the speech is the poem". In the Southern Song Dynasty, Yan Yu's poem "Canglang Shihua" says: "Poetry is the chanting of emotions". There is only one kind of art expressed in words: poetry. In 2022, the Chinese ... Expand\end{tabular} &
  \begin{tabular}{@{}p{3.0cm}@{}} 知道啊,诗歌萌芽于西周,春秋战国时期非常成熟。\\ Yes I know, poetry sprouted in the Western Zhou Dynasty and matured during the Spring, Autumn and Warring States period.
 \end{tabular} \\
\bottomrule

\end{tabular}
}
\end{table*}
\end{CJK*}

\begin{CJK*}{UTF8}{gbsn}
\begin{table*}[t]
\newcolumntype{?}{!{\vrule width 1pt}}
	\newcolumntype{C}{>{\centering\arraybackslash}p{2em}}
	\caption{
		\label{tb:CaseStudygood}  Response generation examples with helpful knowledge snippet injection.
	}
	\centering 
	\small
	\renewcommand\arraystretch{1.0}
 
\scalebox{0.95}{
\begin{tabular}{cp{5.5cm}p{3.0cm}}

\toprule
Dialogue History &
  \multicolumn{1}{c}{Retrieved Knowledge Snippet} &
  \multicolumn{1}{c}{Generated Response} \\ \midrule

\begin{tabular}{@{}p{9cm}@{}}
    朋友,你知道绿萝怎么修剪吗?\\
    Do you know how to prune a green plant, my friend?\\ \vspace{-0.3em}
    绿萝很好打理吧。修剪病黄叶、超过20cm需要对其进行修剪,如果出现病根,腐烂也要拔除哦\\
    It's a easy task to take care of it. Pruning of diseased yellow leaves, over 20cm need to be pruned, if there are diseased roots, decay ones should also be pulled out\\ \vspace{-0.3em}
    我看大家都有绿萝在办公室,绿萝植物有很多好处吧\\
    I see everyone has a green plant in the office, there must of been many benefits of green plants, right?
  \end{tabular} &
  \begin{tabular}{@{}p{5.5cm}@{}} 绿萝的作用有哪些 - 花百科:装饰作用:绿萝颜色翠绿,四季常青,可以在居室中起到很好的装饰作用。 净化空气:它有强大的净化空气能力,能吸收空气中的甲醛。 招风作 ... \\ What is the role of greenery - the Flower Encyclopedia:Decorative role: greenery is green in color, evergreen, and can play a very good decorative role in the living room. It has a strong ability to purify the air and can absorb formaldehyde in the air. The wind as ... \end{tabular} &
  \begin{tabular}{@{}p{3.0cm}@{}} 绿萝的藤蔓细软,叶片娇秀,四季常青,既可观叶,也可培植成盆景,家里养上一盆,既装饰了家居,又起到净化空气的作用,是一种非常适合室内种植的观赏植物。 \\ The green vine is soft and delicate, the leaves are delicate and evergreen, it can be used as a foliage, but can also serve as a bonsai, a pot at home, not only decorate the home, but also plays a role in purifying the air, which is very suitable for indoor planting ornamental plants. \end{tabular} \\

  \cmidrule{1-3}

  \begin{tabular}{@{}p{9cm}@{}}

你喜欢电脑游戏还是手游?\\
Do you like pc games or mobile games?\\ \vspace{-0.3em}
我比较喜欢手游。因为方便。\\
I prefer mobile games. Because it's convenient.\\ \vspace{-0.3em}
那你喜欢王者荣耀吗?\\
Do you like King of Glory?\\ \vspace{-0.3em}
我喜欢呀,但是没有时间玩,没有时间玩儿。\\
I like it, but I don't have time to play, I don't have time to play.\\ \vspace{-0.3em}
现在王者荣耀改版了,玩一局游戏也不到半个小时。\\
Now that King of Glory has been revamped, it takes less than half an hour to play a game.\\ \vspace{-0.3em}
哦,那可能是我真的太久没有玩了。\\
Oh, then I probably haven't played it for too long.\\ \vspace{-0.3em}
你知道王者荣耀是哪个公司的游戏吗?\\
You know which company is behind of King of Glory?\\ \vspace{-0.3em}
是腾讯公司,天美工作室。\\
The company is Tencent, Tianmei Studio.\\ \vspace{-0.3em}
哦,原来是这个公司啊。那就这款游戏来说,能给它挣多少钱啊?\\
Oh, so that's the company behind. How much money can the game make?

  \end{tabular} &
  \begin{tabular}{@{}p{5.5cm}@{}} 王者荣耀游戏一年能赚多少钱\_百度知道:王者荣耀成为历史上第一个收入破百亿美元的移动游戏。 《王者荣耀》日流水=1550.7万人*2.97\%*235元=1.08亿,故《王者荣耀》全年收入394.2亿。 \\ King of Glory  can earn how much money a year\_BaiduQA:King of Glory became the first mobile game in history to break 10 billion dollars in revenue. For the game, " daily flow = 15.507 million people * 2.97\% * 235 = 108 million", so the King of Glory's annual revenue is 39.42 billion. \end{tabular} &
  \begin{tabular}{@{}p{3.0cm}@{}} 我听朋友说去年挣了394亿。 \\ I heard from a friend that it earned 39.4 billion last year.
 \end{tabular} \\
\bottomrule

\end{tabular}
}
\end{table*}
\end{CJK*}

\begin{CJK*}{UTF8}{gbsn}
\begin{table*}[t]
\newcolumntype{?}{!{\vrule width 1pt}}
	\newcolumntype{C}{>{\centering\arraybackslash}p{2em}}
	% \caption{
	% 	\label{tb:CaseStudyGood}  Response generation examples with helpful knowledge snippet injection.
	% }
	\centering 
	\small
	\renewcommand\arraystretch{1.0}
 
\scalebox{0.95}{
\begin{tabular}{cp{7.5cm}p{3.0cm}}

\toprule
Dialogue History &
  \multicolumn{1}{c}{Retrieved Knowledge Snippet} &
  \multicolumn{1}{c}{Generated Response} \\ \midrule

\begin{tabular}{@{}p{7cm}@{}}
    什么水果降火呢\\
    What fruits reduce fire\footnote{Special term in Chinese Medication}\\ \vspace{-0.3em}
    降火的话,那就要买猕猴桃、火龙果、雪梨、山竹、柚子、西瓜这些\\
    If you want to reduce fire, you should buy kiwi, dragon fruit, pear, mangosteen, grapefruit, watermelon, etc.\\ \vspace{-0.3em}
    那冬天吃什么类型的水果最好啊?\\
    What type of fruit is best to eat in winter?\\ \vspace{-0.3em}
    冬天的话,气候比较干燥,而且活动量少,最好多吃点含水量大的水果可以缓解皮肤干燥、上火、便秘、咳嗽啥的,你可以买一些苹果、梨、香蕉、柚子、橘子\\
    In winter, the climate is relatively dry, and less activity, it is best to eat more water-rich fruits can relieve dry skin, fire, constipation, cough, you can try some apples, pears, bananas, grapefruit, oranges\\ \vspace{-0.3em}
    那这样看来,我可以买一些柚子\\
    Perhaps, I can buy some grapefruit\\ \vspace{-0.3em}
    柚子对身体很好的,多吃柚子能帮忙消化、理气化痰、润肺清肠、补血健脾,话说它还能降血糖呢\\
    Grapefruit is very healthy, eat more grapefruit can help digestion, regulate gas and phlegm, lung and intestines, blood and spleen, it is also said that it can lower blood sugar level\\ \vspace{-0.3em}
    它真的有这么多种功效吗?\\
    Does it really has such a variety of effects?
  \end{tabular} &
  \begin{tabular}{@{}p{7.5cm}@{}} 【柚子的功效与作用】柚子的营养价值\_柚子的功效\_亲子百科 ...:现代医药学研究发现,柚肉中含有非常丰富的 维生素C 以及类胰岛素等成分,故有降血糖、降血脂、减肥、美肤养容等功效。 经常食用,对高 血压 、糖尿病、血管硬化等疾病有辅助治疗作用,对肥胖者有健体养颜功能。 柚子还具有健胃、润肺、补血、清肠、利便等功效,可促进伤口愈合,对败血症等有良好的辅助疗效。 此外,柚子含有生理活性物质皮甙,可降低血液的黏滞度,减少血栓的形成,故而对脑血管疾病,如脑血栓、中风等也有较好的预防作用。 而鲜柚肉由于含有类似胰岛素的成分,更是糖尿病患者的理想食品。 查看详情 柚香雪梨丝 葡萄柚子汁 能清热润肺止咳的多功效蜂蜜柚子茶! 怀孕能吃杏脯吗 孕妇消化不良能吃柚子吗 怀孕初期高血糖吃什么好 多吃柚子对身体有什么好处 红肉柚子的功效与作用 多吃柚子能减肥吗 柚子的适用人群 \\ 【The efficacy and role of grapefruit】 The nutritional value of grapefruit\_Parent-child encyclopedia ... Modern medical research has found that grapefruit is very rich in vitamin C and insulin-like ingredients, so it has the effect of lowering blood sugar, lowering blood fat, weight loss, skin beauty and other effects. Regular consumption, high blood pressure, diabetes, vascular sclerosis and other diseases have auxiliary therapeutic effects, for obese people have a body-building function. Grapefruit also has a stomach, lung, blood, intestinal, laxative and other effects, can promote wound healing, sepsis, etc. has a good adjunctive effect. In addition, grapefruit contains physiologically active substances dermatoside, can reduce blood viscosity, reduce the formation of blood clots, so the cerebrovascular diseases, such as cerebral thrombosis, stroke, etc. also has a good preventive effect. And fresh grapefruit because of the insulin-like ingredients, is the ideal food for diabetic patients. See more details Grapefruit and Snow Pear Grapefruit Juice A multi-purpose honey grapefruit tea that clears heat and moistens the lungs to stop coughing! Can pregnant women eat dried apricots Can pregnant women with indigestion eat grapefruit What is good for high blood sugar in early pregnancy What are the health benefits of eating grapefruit Red meat grapefruit efficacy and effects of eating grapefruit can lose weight Grapefruit for people \end{tabular} &
  \begin{tabular}{@{}p{3.0cm}@{}} 当然了,柚子还含有不少有机酸、维生素还有矿物质元素呢 \\ Of course, grapefruit also contains a lot of organic acids, vitamins and mineral elements \end{tabular} \\

  \cmidrule{1-3}

  \begin{tabular}{@{}p{7cm}@{}}
    怎么就没有人想来我们嘉兴玩呢\\
    How come no one wants to come to Jiaxing?\\ \vspace{-0.3em}
    怎么就没有人,我就想去,我老早就想去你们嘉兴的南湖、烟雨楼、曝书亭\\
    I've always wanted to go to the South Lake, the Smoky Rain Tower and the Exposure Pavilion in Jiaxing.\\ \vspace{-0.3em}
    哟,没想到你也知道我们嘉兴的好玩的地儿,真难得,说到杭州大家只想到杭州\\
    Oh, I didn't expect you to know the fun places in Jiaxing, it's rare, when it comes to Hangzhou, people only think of Hangzhou.\\ \vspace{-0.3em}
    是吗,我还喜欢你们那儿桐乡市的乌镇呢,看地理书上描述的,都好想去呢\\
    Really, I also like your Tongxiang City, Wuzhen, see the geography book descriptions, are eager to go to it!\\ \vspace{-0.3em}
    嘿嘿,乌镇也算有名的了\\
    Lol, Wuzhen is also famous\\ \vspace{-0.3em}
    而且嘉兴不是我国历史文化名城吗,还有“鱼米之乡”、“丝绸之府”的美誉呢,茅盾、徐志摩、金庸、金学曙、陈省身、王国维、丰子恺、张乐平等名家大师不都是嘉兴出来的,所以我真的很想去看看\\
    The city of Jiaxing is a historical and cultural city in China, and also the "home of fish and rice", "the capital of silk" reputation, Mao Dun, Xu Zhimo, Jin Yong, Jin Xueshu, Chen Shishi, Wang Guowei, Feng Zikai, Zhang Leping and other famous masters are not out of Jiaxing, so I I really want to go to see\\ \vspace{-0.3em}
    的确,它在历史上算是个名城啦,是不是还有党政性纪念意义啊?\\
    Indeed, it is a famous city in history, is there also a party political memorial significance?

  \end{tabular} &
  \begin{tabular}{@{}p{7.5cm}@{}} 学党史 | 嘉兴:传承红色精神,名城保护再上新台阶\_澎湃号· ...:嘉兴南湖中共“一大”会址是全国重点文物保护单位,南湖革命纪念馆是其宣传、保护、管理机构,成立于1959年10月,属于近现代纪念性专题纪念馆,1985年邓小平同志亲笔题写馆 ... \\ Learning Party History | Jiaxing: inherit the red spirit, the protection of the famous city to a new level\_The Peng Pai- ... The site of the First Chinese Communist Party Congress at South Lake in Jiaxing is a national key cultural relics protection unit, and the South Lake Revolutionary Memorial Hall is its propaganda, protection and management institution, which was established in October 1959 and is a modern monumental thematic memorial hall. \end{tabular} &
  \begin{tabular}{@{}p{3.0cm}@{}} 对啊,中国共产党第一次全国代表大会就在这里举行的,所以算是个著名的地方了 \\ Yes, the first national congress of the Chinese Communist Party was held here, so it's a famous place
 \end{tabular} \\
\bottomrule

\end{tabular}
}
\end{table*}
\end{CJK*}

\end{document}